\documentclass{article}

\usepackage{microtype}
\usepackage{graphicx}
\usepackage{subcaption}
\usepackage{booktabs} % for professional tables

\usepackage{tocloft}
\usepackage{hyperref}
\usepackage{url}            % simple URL typesetting

\usepackage[accepted]{icml2026}

\usepackage{amsmath}
\usepackage{amssymb}
\usepackage{mathtools}
\usepackage{amsthm}
\usepackage{multirow} 
\usepackage{algorithm}
\usepackage{algorithmic}
\usepackage[labelformat=simple]{subcaption}

\usepackage{pifont}
\usepackage{wrapfig}
\usepackage{enumitem}
\usepackage{xcolor}         % colors
\usepackage{nicefrac}       % compact symbols for 1/2, etc.
\usepackage{amsfonts}       % blackboard math symbols
\usepackage{setspace}

\usepackage[capitalize,noabbrev]{cleveref}

\theoremstyle{plain}

\newtheorem{proposition}{Proposition}

\theoremstyle{definition}

\theoremstyle{remark}

\renewcommand{\cite}{\citep}

\usepackage[textsize=tiny]{todonotes}

\icmltitlerunning{RL-SPH: Learning to Achieve Feasible Solutions for Integer Linear Programs}

\begin{document}

\twocolumn[
  \icmltitle{RL-SPH: Learning to Achieve Feasible Solutions for Integer Linear Programs}



  \icmlsetsymbol{equal}{*}

  \begin{icmlauthorlist}
    \icmlauthor{Tae-Hoon Lee}{kaist}
    \icmlauthor{Min-Soo Kim}{kaist}
    % \icmlauthor{Firstname3 Lastname3}{comp}
  \end{icmlauthorlist}

  \icmlaffiliation{kaist}{School of Computing, KAIST, Daejeon, Republic of Korea}

  \icmlcorrespondingauthor{Tae-Hoon Lee}{th.lee@kaist.ac.kr}
  \icmlcorrespondingauthor{Min-Soo Kim}{minsoo.k@kaist.ac.kr}

  \icmlkeywords{Machine Learning, ICML, Integer Linear Programming, Start Primal Heuristics, Reinforcement Learning, Graph Neural Networks}

  \vskip 0.3in
]



\printAffiliationsAndNotice{}  % no special notice (required even if empty)

\begin{abstract}
Primal heuristics play a crucial role in quickly finding feasible solutions for NP-hard integer linear programming\,(ILP).
Although \textit{end-to-end learning}-based primal heuristics\,(E2EPH) have recently been proposed, they are typically unable to independently generate feasible solutions.
To address this challenge, we propose RL-SPH, a novel reinforcement learning-based start primal heuristic capable of independently generating feasible solutions, even for ILP involving non-binary integers.
Empirically, RL-SPH rapidly obtains high-quality feasible solutions with a 100\% feasibility rate, achieving on average a 28.6$\times$ lower primal gap and a 2.6$\times$ lower primal integral compared to existing start primal heuristics.
% Notably, the LP-free variant of RL-SPH discovers the first feasible solution 97$\times$ more rapidly.
\end{abstract}

\vspace{-0.2cm}
\section{Introduction}\label{Introduction}
\vspace{-0.1cm}
% The traveling salesman problem and the knapsack problem are classic examples of combinatorial optimization\,(CO) problems, extensively studied in operations research and computer science\,\cite{ml4co_comp}.
Combinatorial optimization\,(CO) involves mathematical optimization that aims to minimize or maximize a specific objective function\,\cite{mazyavkina2021reinforcement}.
When both the objective function and the constraints of CO are linear, the problem is referred to as linear programming\,(LP)\,\cite{bengio2021machine}. 
Furthermore, if the variables in LP are required to take integer values, it becomes an integer linear programming\,(ILP)\,\cite{bertsimas1997introduction}.
ILP has been widely applied to real-world scenarios such as logistics\,\cite{kweon2024parcel}, the vehicle routing problem\,\cite{toth2002vehicle}, and path planning\,\cite{zuo2020milp}.

\begin{figure}[t]
     \centering
     \begin{subfigure}[t]{0.95\columnwidth}
         \centering
         \includegraphics[width=\linewidth]{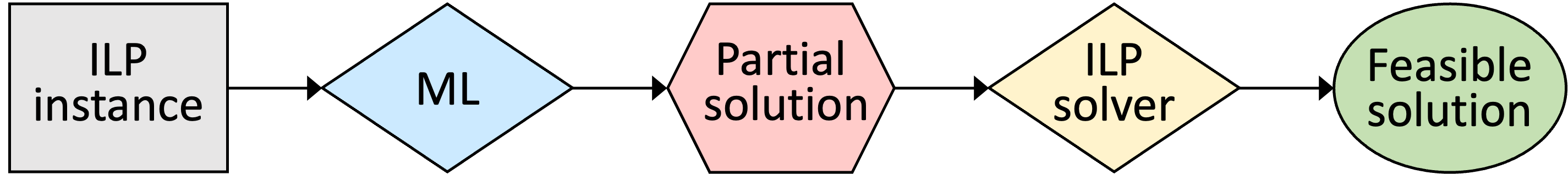}
         \caption{Existing E2EPH integrated with ILP solvers.}
         \label{fig:primal_heuristics}
     \end{subfigure}%
     \vspace{0.1cm}
     \begin{subfigure}[t]{0.485\columnwidth}
         \centering
         \includegraphics[width=\linewidth]{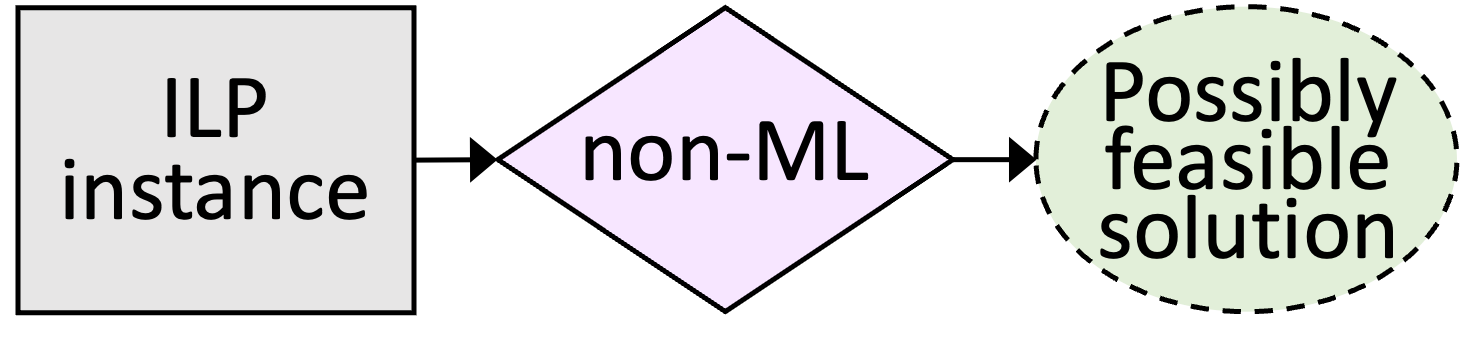}
         \caption{Existing SPH.}
         \label{fig:sph}
     \end{subfigure}
     \hfill
     \begin{subfigure}[t]{0.485\columnwidth}
         \centering
         \includegraphics[width=\linewidth]{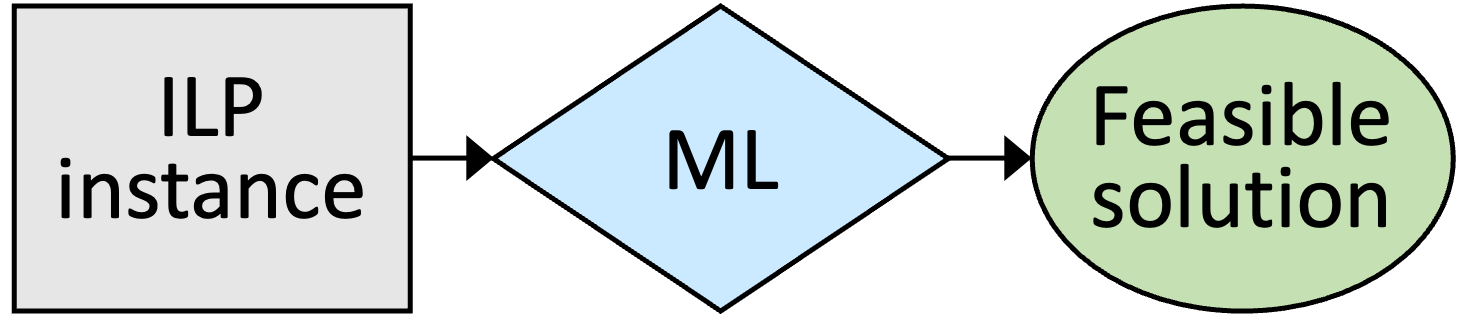}
         \caption{RL-SPH\,(Ours).}
         \label{fig:ours}
     \end{subfigure}
\vspace{-0.1cm}        
        \caption{Comparison among \textit{end-to-end learning}-based primal heuristics\,(E2EPH), start primal heuristics\,(SPH), and ours.}
        % \caption{Comparison among E2E primal heuristics, start primal heuristics\,(SPH), and our approach.}
         \label{fig:method_comparison}
\vspace{-0.4cm}        
\end{figure}

Since ILP is NP-hard, heuristic approaches have attracted significant attention\,\cite{berthold2006primal}.
Primal heuristics aim to quickly find feasible solutions\,\cite{berthold2006primal,shoja2023exact}, in contrast to methods that aim for optimality\,\cite{canturk2024scalable}.
Traditional primal heuristics often rely on expert knowledge, requiring significant manual effort\,\cite{bengio2021machine}.
%Machine learning\,(ML) offers a promising alternative by leveraging shared patterns within a specific distribution\,\cite{ml4co_comp}.
Recently, ML-based primal heuristics have been proposed\,\cite{nair2020solving, shen2021learning, yoon2022confidence, PAS, canturk2024scalable, huang2024contrastive, kdd_complete, liu2025apollo, iclr_complete}, which fall under the category of \textit{end-to-end learning}\,\cite{bengio2021machine, PAS}, as they learn common patterns across ILP instances and directly generate solutions.
Figure\,\ref{fig:primal_heuristics} illustrates how existing \textit{end-to-end learning}-based primal heuristics\,(E2EPH) participate in the process of finding feasible solutions.
A trained ML model generates \textit{a partial solution} over integer variables, which is then passed to an external ILP solver\,(e.g., Gurobi and SCIP) to obtain a feasible solution for the subproblem.
% By a trained ML model

E2EPH methods integrated with ILP solvers efficiently find high-quality solutions by reducing the search space.
Despite these advances, ensuring feasibility remains a major challenge, as inaccurate ML predictions can lead to constraint violations\,\cite{PAS}, posing a significant obstacle to solving ILP.
Recent studies\,\cite{PAS, huang2024contrastive, liu2025apollo} have sought to mitigate this risk by adopting trust regions rather than strictly fixing variables\,\cite{nair2020solving, yoon2022confidence}.
However, they rely on external solvers to attain feasibility.
Several E2EPH have been proposed to eliminate this reliance\,\cite{kdd_complete, iclr_complete}, but they still struggle to obtain feasible solutions independently.
This limitation underscores the need for a new class of E2EPH that can independently produce feasible solutions, known as \textit{start primal heuristics}\,(SPH), which attempt to convert infeasible solutions into feasible ones\,\cite{berthold2006primal}\,(see Appendix\,\ref{Start primal heuristics for ILP}).

The infeasibility caused by inaccurate ML predictions can be more pronounced for non-binary integer\,(hereafter, integer) variables due to their wider value range compared to binary variables.
Many real-world problems involve integer variables\,(e.g., logistics\,\cite{kweon2024parcel} and maritime transportation\,\cite{papageorgiou2014mirplib}).
However, existing E2EPH studies have primarily focused on binary variables\,\cite{PAS, huang2024contrastive, kdd_complete, liu2025apollo, iclr_complete}, highlighting the need for E2EPH capable of effectively handling integer variables.

This work focuses on achieving feasibility in ILP, a critical prerequisite for optimality.
For instance, popular metaheuristics such as local branching\,\cite{liu2022learning} and large neighborhood search (LNS)\,\cite{huang2023searching, liu2024mixed} require an initial feasible solution to initiate their optimization process.
Given that reaching optimality remains a significant challenge in the ML for CO field\,\cite{meta-sage}, ensuring feasibility becomes an essential first step in solving ILP.
However, feasibility is not typically guaranteed by established primal heuristics\,\cite{shoja2023exact}.
While finding feasible solutions for ILP is not trivial due to its theoretically NP-hard nature, it poses an even greater challenge in the context of ML\,\cite{bengio2021machine}.

To tackle this challenge—where ML must generate feasible solutions for ILP independently of external solvers—we propose a novel \textbf{R}einforcement \textbf{L}earning-based \textbf{S}tart \textbf{P}rimal \textbf{H}euristic, called \textbf{RL-SPH}.
Figure\,\ref{fig:ours} depicts how RL-SPH generates feasible solutions for ILP.
We design an ILP-specific RL framework, in which the agent learns variable-constraint relationships to discover feasible solutions.
% Instead of directly predicting variable values, 
Specifically, the agent learns to decide whether to change the value of each variable, guided by reward signals derived from the degree of constraint violations and the quality of the solution.
We also introduce ILP-GT, a Graph Transformer specifically designed for ILP to effectively model long-range dependencies between variables\,(see Appendix\,\ref{GT}).
Our main contributions are as follows, and Appendix\,\ref{Related Work} further contextualizes our research.

\vspace{-0.1cm}
\begin{itemize}[labelindent=0em, topsep=0pt, itemsep=0pt, partopsep=0pt, leftmargin=*]
    \item We propose a novel RL-based start primal heuristic, RL-SPH, which learns to independently generate high-quality feasible solutions for ILP. To the best of our knowledge, RL-SPH is the first E2EPH to establish theoretical reward-feasibility alignment, a conditional feasibility guarantee.
    % \item We propose a novel RL-based start primal heuristic, RL-SPH, which learns to independently generate high-quality feasible solutions for ILP. To the best of our knowledge, RL-SPH is the first E2EPH to establish theoretical reward-feasibility alignment.
    % \item We propose a novel RL-based start primal heuristic, RL-SPH, which learns to independently generate high-quality feasible solutions for ILP. To the best of our knowledge, RL-SPH is the first E2EPH to establish theoretical reward-feasibility alignment, with conditional feasibility guarantees.

    \item We design an RL framework tailored for ILP, wherein the ILP-GT agent interacts to acquire problem-solving capabilities in an end-to-end fashion. By leveraging our feasibility-aware search strategy to narrow the search space, the agent efficiently learns to achieve ILP-feasible solutions via strong reward signals.

    \item We demonstrate that RL-SPH achieves a 100\% feasibility rate\,(FR) across five CO benchmarks—including non-binary ILP—while outperforming seven baselines comprising four established SPH and three E2EPH methods. Our evaluations include a wide array of tests: cross-problem generalization, robustness to distributional shifts, tests on MIPLIB instances, and solver integration, alongside hyperparameter and qualitative analyses.
\end{itemize}

\begin{figure*}[t]
    % \vspace{-0.1cm}
  \centering
  \includegraphics[width=0.95\textwidth]{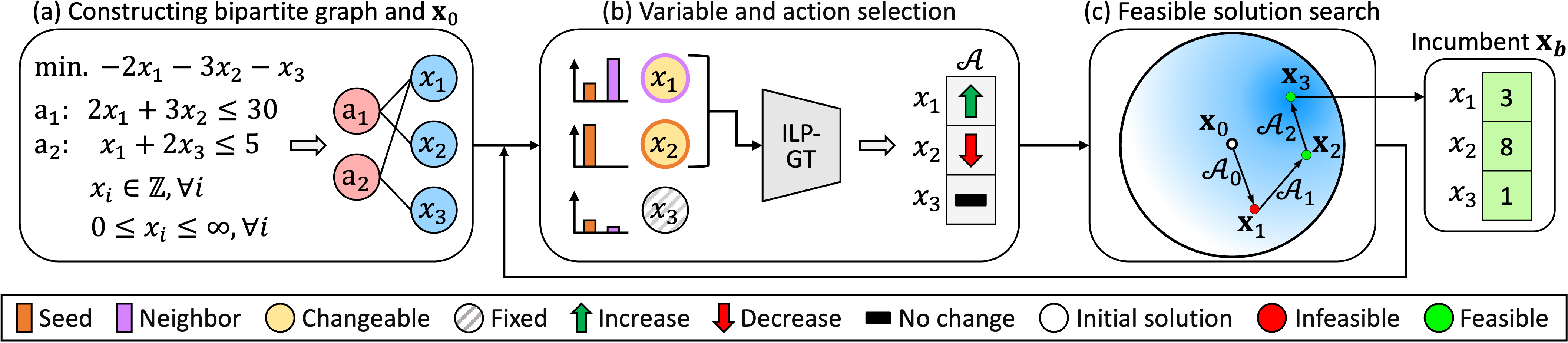}
  \vspace{-0.15cm}
  \caption{The overview of RL-SPH.}
  \vspace{-0.4cm}
  \label{fig:overview}
\end{figure*}

% \vspace{-0.2cm}

\vspace{-0.2cm}
\section{Preliminaries}\label{Preliminaries}

\vspace{-0.1cm}
\subsection{Integer Linear Programming} \label{Mixed Integer Linear Programming}
\vspace{-0.1cm}
% \subsubsection{A standard form}

Integer linear programming\,(ILP) is an optimization problem that minimizes or maximizes a linear objective function, while satisfying linear constraints and integrality constraints on decision variables\,\cite{bertsimas1997introduction}.
A standard form of an ILP instance can be formulated as follows:
\begin{subequations} \label{eq:standard_from_ilp}
\vspace{-0.4cm}
\begin{align}
    \text { minimize } & \mathbf{c}^\top \mathbf{x} \label{eq:obj_func} \\
    \text { subject to } & \mathbf{A} \mathbf{x} \leq \mathbf{b} \label{eq:constraints} \\
    & x_i \in \mathbb{Z},  \quad \forall i  \label{eq:int_requirement} \\
    & l_i \leq x_i \leq u_i,  \quad  \forall i \label{eq:bound} 
\end{align}
% \vspace{-0.3cm}
\end{subequations}
where $\mathbf{x} \in \mathbb{R}^n$ is a column vector of $n$ decision variables,
$\mathbf{c} \in \mathbb{R}^n$ is a column vector of the objective coefficients, 
$\mathbf{A} \in \mathbb{R}^{m \times n}$ is the constraint coefficient matrix,
$\mathbf{b} \in \mathbb{R}^m$ is a column vector of the right-hand side of the constraints,
% $I$ is the index set of integer decision variables,
and $l_i$/$u_i$ denote the lower/upper bounds for each decision variable $x_i$.
A solution $\mathbf{x}$ is said to be \textit{feasible} if it satisfies all constraints\,(Eqs.\,\ref{eq:constraints}--\ref{eq:bound}).
ILP aims to find a feasible solution that minimizes $obj = \mathbf{c}^\top \mathbf{x}$\,(Eq.\,\ref{eq:obj_func}) in the case of a minimization problem, which defines the optimal solution.
% ILP aims to find an optimal solution that minimizes $obj = \mathbf{c}^\top \mathbf{x}$\,(Eq.\,\ref{eq:obj_func}), in the case of a minimization problem.
Additional properties are provided in Appendix\,\ref{Properties of ILP}.

\vspace{-0.1cm}
\subsection{Actor–Critic Reinforcement Learning}
\vspace{-0.1cm}
% We adopt the Actor-Critic\,(AC) algorithm\,\cite{a2c_paper}, which has proven effective for CO problems\,\cite{bello2016neural, chemical, mapdp}.
The Actor–Critic\,(AC) framework\,\cite{a2c_paper} is a foundational RL paradigm that consists of two components: an actor, which selects actions, and a critic, which evaluates them. 
% The actor optimizes parameters $\theta$ to maximize the expected reward by learning a policy $\pi_\theta(\mathcal{A} \mid \mathcal{S})$ that maps the observation $\mathcal{S}$ to a probability distribution over actions $\mathcal{A}$.
In this framework, the actor aims to maximize the expected reward by learning a policy $\pi_\theta(\mathcal{A} \mid \mathcal{S})$ that maps an observation $\mathcal{S}$ to a probability distribution over actions $\mathcal{A}$, while the critic evaluates $\mathcal{S}$ using a value function $V_\theta(\mathcal{S})$.
Motivated by the effectiveness of AC for CO problems\,\cite{chen2019rewrite, chemical, mapdp}, we design an ILP-tailored AC to train our ILP-GT.

\vspace{-0.1cm}
\subsection{Neighborhood Search for CO} \label{Neighborhood search}
\vspace{-0.1cm}

Local search\,(LS) heuristic is widely used for exploring the neighborhood of a current solution for CO problems\,\cite{hillier2015introduction}.
Although LS may encounter local optima, empirical studies have shown it to be effective in quickly identifying high-quality solutions\,\cite{bertsimas1997introduction}.
Classic LS typically perturbs a single variable at a time, whereas exploring larger neighborhoods has been shown to improve effectiveness\,\cite{shaw1998using}.
In contrast, with our newly designed search strategy, RL-SPH allows multiple variables to change simultaneously as shown in Figure\,\ref{fig:overview}, thereby exploring a broader neighborhood.
Our search strategy resembles RENS in that it solves subproblems by fixing a set of variables and also LNS in exploring large neighborhoods. 
However, unlike RENS and RL-SPH, LNS requires an \textit{initial feasible solution}, limiting its use as a start primal heuristic\,\cite{berthold2006primal}.

\section{Methodology}\label{Methodology}
\vspace{-0.1cm}

% This section presents RL-SPH in detail.
Figure\,\ref{fig:overview} illustrates the overall process of RL-SPH.
Given an ILP instance, RL-SPH constructs a bipartite graph\,(see Appendix\,\ref{Graph Representation of ILP}) and an initial solution $\mathbf{x}_0$.
At each timestep, it selects $\tilde{n}$\,$(=2 \left\lceil\log_2n\right\rceil)$ changeable variables as input to the trained agent.
The agent selects actions expected to yield high rewards and generates a new solution. 
As the process repeats, the best feasible solution found so far\,(i.e., the incumbent) $\mathbf{x}_b$ is updated whenever $\mathbf{x}_{t+1}$ is both feasible and improves upon $\mathbf{x}_b$.

\vspace{-0.1cm}
\subsection{Reinforcement Learning for ILP} \label{Reinforcement Learning for ILP}
\vspace{-0.1cm}

Our RL framework for ILP aims to train an agent to make decisions that maximize rewards while interacting with a given ILP instance $M$.
Figure\,\ref{fig:rl} depicts how the RL agent interacts with $M$, where $\mathcal{S}_t$, $\mathcal{A}_t$, and $\mathcal{R}_{t,\text{total}}$ denote the observation, the set of actions, and the total reward at timestep $t$, respectively.
% The instance $M$ serves as the environment.
Using $\mathcal{A}_{t}$, the agent updates the solution $\mathbf{x}_{t+1}$ for $n$ variables. 
This update affects the left-hand side of the constraints $\mathbf{lhs}_{t+1}$, the feasibility state vector $\mathbf{f}_{t+1}$, and the objective value $obj_{t+1}$.
At the next timestep, the agent receives a new observation $\mathcal{S}_{t+1}$ changed by $\mathcal{A}_{t}$ and selects a new action set $\mathcal{A}_{t+1}$ to maximize rewards.
By comparing the estimated reward with the actual reward $\mathcal{R}_{t+1,\text{total}}$ obtained from $\mathcal{A}_{t+1}$, the agent refines its policy $\pi$.

\vspace{-0.1cm}
\subsubsection{Action} \label{Action}
\vspace{-0.1cm}
At timestep $t$, the agent selects a set of actions $\mathcal{A}_{t} =\,(a_{t,1}, \dots, a_{t,n})$ for $n$ variables based on $\mathcal{S}_t$.
For each variable, the agent can take one of three actions: increase, no change, or decrease, as depicted in Figure\,\ref{fig:rl_diagram}.
The magnitude of change for both increases and decreases is set to 1, as discussed in Appendix\,\ref{appendix:movement_unit}.
Non-changeable variables are treated as fixed\,(i.e., no change), as shown in Figure\,\ref{fig:overview}(b).

\vspace{-0.1cm}        
\subsubsection{Observation} \label{Observation}
\vspace{-0.1cm}

We define the observation $\mathcal{S}_t =\,(\mathbf{x}_t, \mathbf{f}_t, obj_t)$.
The solution $\mathbf{x}_t$ is obtained by updating variable values based on the previous actions $\mathcal{A}_{t-1}$.
For example, if $\mathcal{A}_{t-1}=(1,-1,0)$, then $\mathbf{x}_{t-1}$ in Figure\,\ref{fig:rl_example} is updated to $\mathbf{x}_t=(x_{t,1}, \, x_{t,2}, \, x_{t,3})=(4,8,0)$.
Using $\mathbf{x}_t$, the new $\mathbf{lhs}_t = \mathbf{A} \mathbf{x}_t$ and $obj_t = \mathbf{c}^\top \mathbf{x}_t$ are calculated.
Each element of $\mathbf{f}_t$\,($= \mathbf{b} - \mathbf{lhs}_t$) indicates whether the corresponding constraint is satisfied by $\mathbf{x}_t$.
Non-negative elements in $\mathbf{f}_t$ indicate satisfied constraints, while negative ones indicate violations. 
For example, in Figure\,\ref{fig:rl}, $\mathbf{x}_{t+1}$ yields $\mathbf{lhs}_{t+1}=(34,3)$ for constraints $\mathbf{a}_1$ and $\mathbf{a}_2$.
Since $\mathbf{f}_{t+1} = \mathbf{b} - \mathbf{lhs}_{t+1}=(30, 5) -\,(34, 3) =\,(-4, 2)$, $\mathbf{x}_{t+1}$ violates $\mathbf{a}_1$ but satisfies $\mathbf{a}_2$.

\vspace{-0.1cm}
\subsubsection{Reward} \label{Reward}
\vspace{-0.1cm}
We design reward functions to guide the agent in selecting actions that maximize the total reward $\mathcal{R}_{t,\text{total}}$, as follows:
\begin{equation}
    \mathcal{R}_{t,\text{total}} = \mathcal{R}_{t,\text{opt}} + \mathcal{R}_{t,\text{explore}}
    % \mathcal{R}_{t,\text{total}} = \mathcal{R}_{t,\text{const}} + \mathcal{R}_{t,\text{bound}} + \mathcal{R}_{t,\text{opt}} + \mathcal{R}_{t,\text{explore}}
    \label{eq:total_reward}
\end{equation}
\begin{equation}
    \mathcal{R}_{t,\text{opt}} = 
    \begin{cases}
        \mathcal{R}_{t,\text{p1}}, & \text{if agent is in $phase1$}, \\
        \mathcal{R}_{t,\text{p2}}, & \text{otherwise}.
    \end{cases}
    \label{eq:reward_obj}
\end{equation}
To fulfill the goal of start primal heuristics—obtaining high-quality feasible solutions quickly—we decompose the process into two phases: the rapid discovery of an initial feasible solution in $phase1$ and the search for higher-quality feasible solutions in $phase2$.
$phase1$ lasts until the first feasible solution is found, while $phase2$ begins thereafter.
The reward functions for each phase are defined as follows:
% The reward system operates in two phases: 
\begin{equation}
    \mathcal{R}_{t,\text{p1}} = 
    \begin{cases}
       \mathcal{R}_{t,\text{bound}}, & \text{if }  \neg\mathcal{B}_t \land\, \mathcal{C}_t \land \mathcal{O}_t, \\
       \mathcal{R}_{t,\text{bound}} - \Delta obj_t, & \text{if }  \neg\mathcal{B}_t \land\,\mathcal{C}_t \land \neg\mathcal{O}_t, \\
       \mathcal{R}_{t,\text{F}} + \Delta obj_t, & \text{if }  \mathcal{B}_t \land\, \mathcal{C}_t \land \mathcal{O}_t, \\
       \mathcal{R}_{t,\text{F}} - \Delta obj_t, & \text{if }  \neg\mathcal{C}_t \land \neg\mathcal{O}_t, \\
       \mathcal{R}_{t,\text{F}}, & \text{otherwise}.
    \end{cases}
    \label{eq:reward_phase1_obj}
\end{equation}
% \vspace{-0.1cm}
% {\small
\begin{equation}
    \mathcal{R}_{t,\text{p2}} = 
    \begin{cases}
        \Delta obj_t, & \text{if } \mathbf{x}_{t+1} \in \mathcal{F} \land obj_{t+1} < obj_b, \\
        -\Delta obj_t \cdot \alpha, & \text{if }  \mathbf{x}_{t+1} \in \mathcal{F} \land obj_{t+1} \geq  obj_b, \\
         \mathcal{R}_{t,\text{F}}, & \text{if }  \mathbf{x}_{t+1} \notin \mathcal{F} \land obj_{t+1} < obj_b, \\
       \mathcal{R}_{t,\text{F}}\cdot \alpha  & \text{otherwise}.
    \end{cases}
    \label{eq:reward_phase2_obj} 
\end{equation}
% \vspace{-0.1cm}
\begin{equation}
    \mathcal{R}_{t,\text{F}} = \mathcal{R}_{t,\text{bound}} + \frac{1}{\sqrt{\tilde{n}}}\mathcal{R}_{t,\text{const}} \label{eq:reward_feasible} 
\end{equation}
\vspace{-0.15cm}
\begin{equation}
    \mathcal{R}_{t,\text{bound}} = - \sum_{i=1}^n \mathbb{I} \left( x_{t+1,i} \notin [l_i, u_i] \right)
    \label{eq:reward_bound}
\end{equation}
\vspace{-0.15cm}
\begin{equation}
    \mathcal{R}_{t,\text{const}} =\sum_{j=1}^m  \min(f_{t+1,j}, 0) - \min(f_{t,j}, 0)
    \label{eq:reward_const}
\end{equation}
where 
$\mathcal{B}_t$ is the condition $x_{t+1,i} \in [l_i, u_i]$ for all $i$,
$\mathcal{C}_t$ is $\mathcal{R}_{t,\text{const}} > 0$,
$\mathcal{O}_t$ is $obj_{t+1} < obj_t$,
$\mathcal{F}$ is the feasible region, 
$\Delta obj_t = |obj_{t+1}-obj_{t}| / \max(|\textbf{c}|)$,
$\alpha$ is a toward-optimal bias,
$\mathbb{I}$ is the indicator function,
$x_{t,i}$ of $\mathbf{x}_t$ is the value of the $i$-th variable at $t$,
$f_{t,j}$ is the element of $\mathbf{f}_t$ for the $j$-th linear constraint, 
and $\tilde{n}$ is the number of changeable variables.

\begin{figure}[t]
    % \vspace{-0.1cm}
     \centering
     \begin{subfigure}[t]{0.97\columnwidth}
         \centering
         \includegraphics[width=\linewidth]{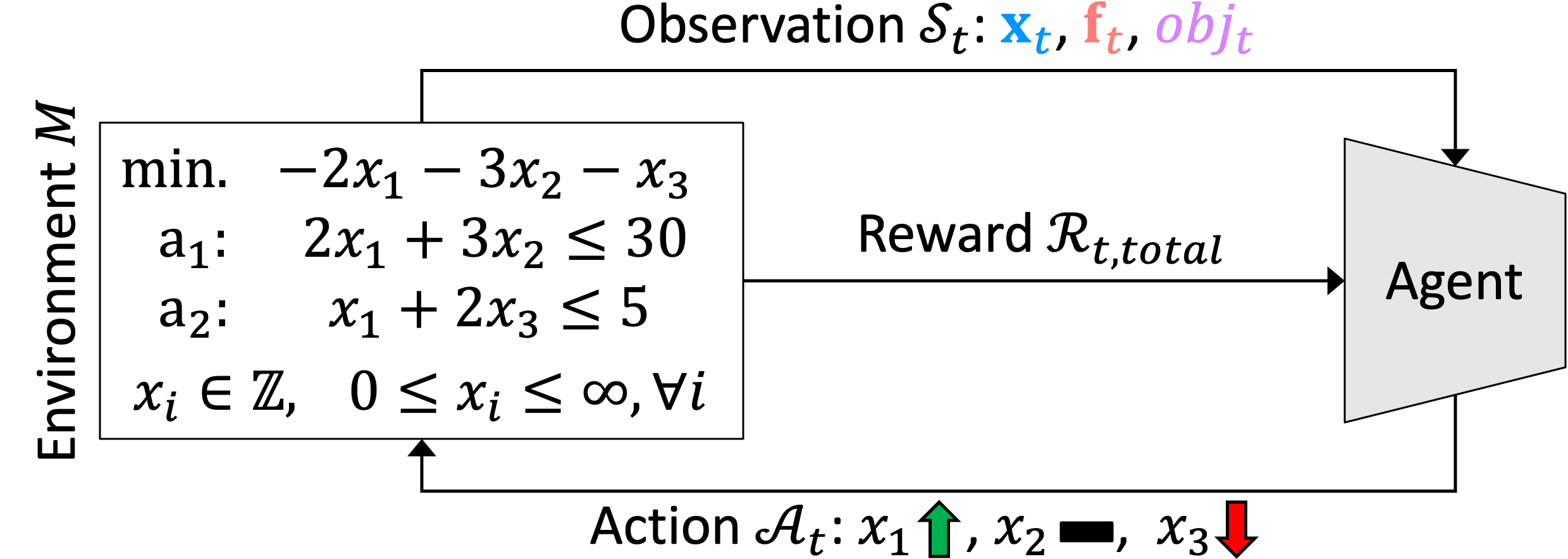}
         \caption{Diagram of RL framework for ILP.}
         \label{fig:rl_diagram}
     \end{subfigure}%
     \vspace{0.15cm}
     \begin{subfigure}[t]{0.97\columnwidth}
         \centering
         \includegraphics[width=\linewidth]{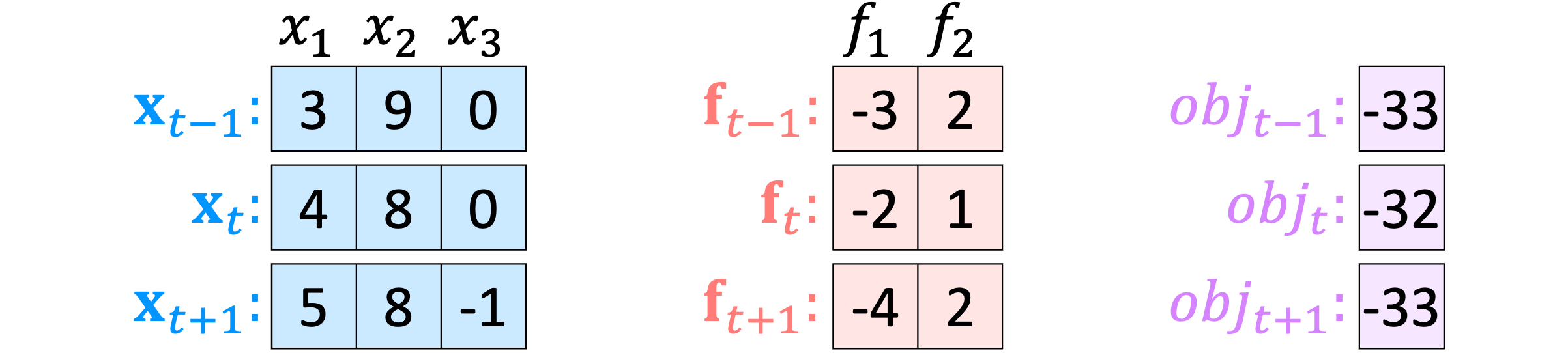}
         \caption{Examples of observation for Figure\,\ref{fig:rl_diagram}.}
         \label{fig:rl_example}
     \end{subfigure}
        \vspace{-0.1cm}        
        \caption{Reinforcement learning for ILP.}
        \vspace{-0.3cm}        
        \label{fig:rl}
\end{figure}

Finding a feasible solution is a prerequisite for improving the incumbent. 
Thus, our primary goal is to find a feasible solution that satisfies all constraints.
The feasibility reward $\mathcal{R}_{t,\text{F}}$, used in Eqs.\,\ref{eq:reward_phase1_obj} and\,\ref{eq:reward_phase2_obj}, is computed based on the variable bounds and linear constraints. 
The bound reward $\mathcal{R}_{t,\text{bound}}$ imposes a penalty proportional to the number of variables that violate their bounds.
For example, in Figure\,\ref{fig:rl_example}, $\mathcal{R}_{t,\text{bound}} = -1$ since $l_i=0$ and $x_{t+1,3}= -1$.
The constraint reward $\mathcal{R}_{t,\text{const}}$ reflects the improvement\,(or deterioration) for each infeasible linear constraint.
For example, in Figure\,\ref{fig:rl_example}, $\mathcal{R}_{t,\text{const}}$ is $[\{-4-(-2)\}+\{0-(0)\}]=-2$.

In $phase1$, the primary goal is to find the first feasible solution from an infeasible initial solution.
Since improvements in constraint violations or objective values are meaningless when variable bounds are violated, satisfying the variable bounds should be prioritized above all else.
Thus, positive values of $\mathcal{R}_{t,\text{const}}$ are ignored to prioritize satisfying the bounds\,(Cases 1, 2 in Eq.\,\ref{eq:reward_phase1_obj}).
In all other cases, the agent receives $\mathcal{R}_{t,\text{F}}$ during $phase1$\,(Cases 3, 4, 5 in Eq.\,\ref{eq:reward_phase1_obj}).
Achieving a better\,(i.e., lower) objective value in $phase1$ leads to a stronger starting point for $phase2$.
Accordingly, the agent receives a reward proportional to the changes in the objective value\,(Cases 2, 3, 4 in Eq.\,\ref{eq:reward_phase1_obj}).
However, a positive $\Delta obj_t$ is given only when feasibility improves\,(Case 3 in Eq.\,\ref{eq:reward_phase1_obj}), preserving the priority.
A well-trained agent selects actions that maximize the rewards and is thus theoretically directed toward a feasible solution, grounded in the reward-feasibility alignment established in Proposition~\ref{prop:feasibility}:
\begin{proposition}\label{prop:feasibility}
     Suppose $\mathbf{x}_t \notin \mathcal{F}$,  $\mathcal{R}_{t,\text{const}} > 0$, and $\mathcal{R}_{t,\text{bound}} = 0$ $\text{ for all } t < \mathcal{T}$. Then $\mathbf{x}_\mathcal{T} \in \mathcal{F}$.
\end{proposition}
\vspace{-0.5cm}
\begin{proof} 
Appendix\,\ref{appendix:proof_feasibility} provides the proof of Proposition~\ref{prop:feasibility}.
\end{proof}
\vspace{-0.1cm}

In $phase2$, the goal is to improve the incumbent solution $\mathbf{x}_{b}$. 
If the agent finds $\mathbf{x}_{t+1} \in \mathcal{F}$, it receives a reward proportional to the changes in the objective value\,(Cases 1, 2 in Eq.\,\ref{eq:reward_phase2_obj}).
If $\mathbf{x}_{t+1} \notin \mathcal{F}$, it is penalized for violations\,(Cases 3, 4 in Eq.\,\ref{eq:reward_phase2_obj}).
A suitable $\alpha$ promotes the agent to explore promising regions where $obj_{t+1} < obj_b$.
Empirically, search performance degrades in the absence of the toward-optimal bias\,(i.e., $\alpha = 1$); thus, we set $\alpha$ to 2.
Appendix\,\ref{appendix:toward-optimal} presents an evaluation of the impact of toward-optimal bias.

Exploration must take precedence even over satisfying variable bounds; otherwise, the agent might remain stationary to avoid potential penalties.
To discourage such behavior, we assign a heavy exploration penalty $\mathcal{R}_{t,\text{explore}}=-100$ only if $\mathbf{x}_{t+1} = \mathbf{x}_{t}$.
The maximum bound penalty occurs when all changeable variables violate their bounds, expressed as $-\tilde{n}=-2 \left\lceil\log_2n\right\rceil$.
Exceeding $-100$ would require a bound penalty corresponding to $n>10^{15}$, which is practically infeasible.
Thus, we impose $-100$ if the agent does not move.
Appendix\,\ref{appendix:vis_expl} further details our reward system.

\vspace{-0.1cm}
\subsection{Graph Transformer for ILP} \label{GNN Architecture of RL Agent}
\vspace{-0.1cm}
\begin{figure}[b]{}
% \begin{wrapfigure}{r}{0.495\columnwidth}
  % \vspace{-2em}
  \vspace{-0.2cm}
  \centering
  \includegraphics[width=0.95\columnwidth]{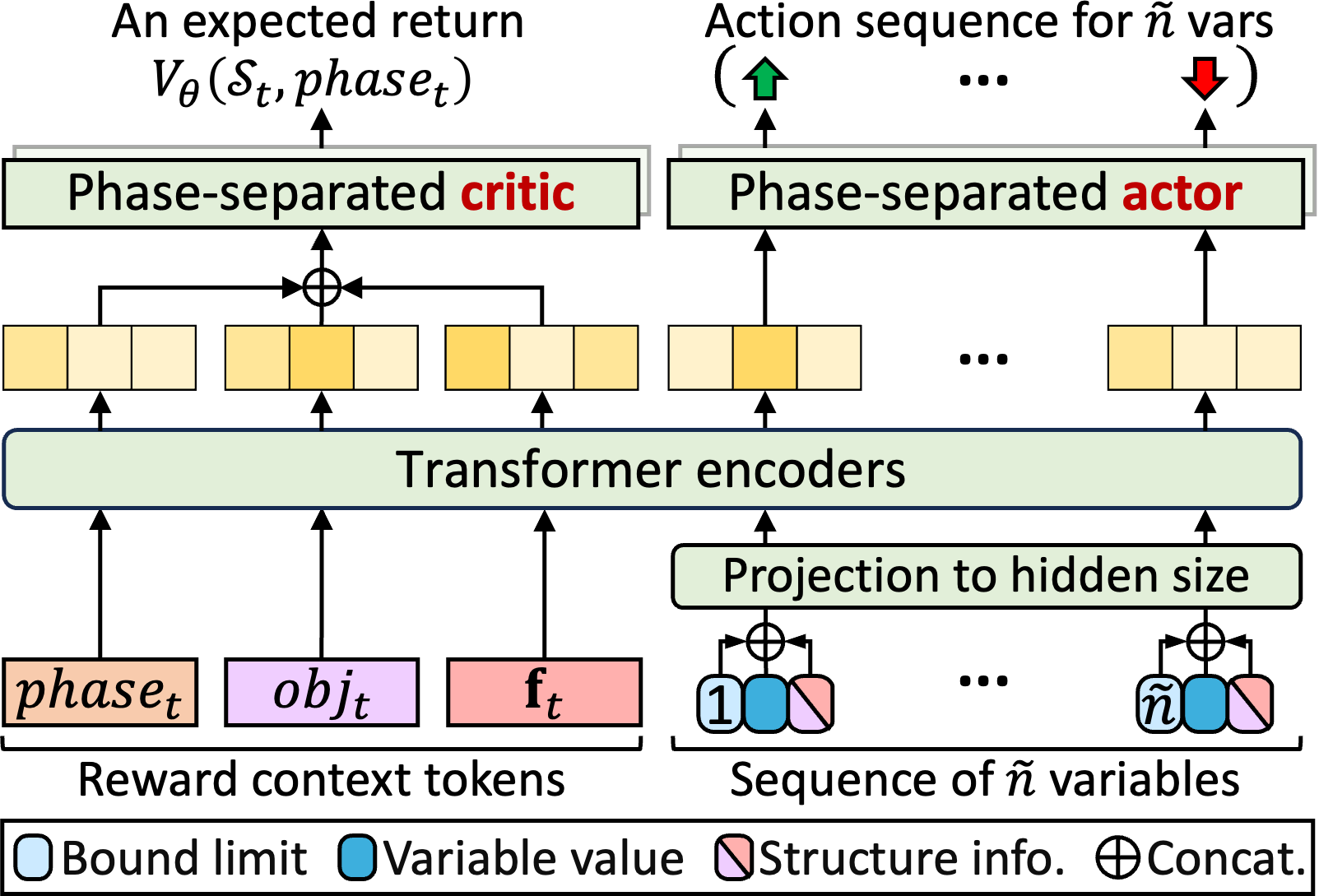}
  \vspace{-0.1cm}
  \caption{The architecture of ILP-GT.}
  \label{fig:gnn}
  % \vspace{-0.3cm}
\end{figure}
% \end{wrapfigure}

Figure\,\ref{fig:gnn} shows the architecture of our Graph Transformer, ILP-GT.
The actor generates a sequence of actions conditioned on the selected $\tilde{n}$ variables and their structural information $(\mathbf{c}^\top|\mathbf{A})$.
To stabilize training, we scale $(\mathbf{c}^\top|\mathbf{A})$ to $[-1, 1]$ using equilibration scaling\,\cite{tomlin1975scaling}, which normalizes each constraint by its largest absolute coefficient while preserving problem equivalence.
The scaled structural information is projected to the hidden size.
% This scaling preserves problem equivalence, since multiplying a constraint by a positive scalar does not alter the solution space.
% Two variable-related features are extracted: a binary feature $bnd\_lim$ and the raw variable values.
A binary feature $bnd\_lim$ is set to 1 if a variable value reaches or exceeds its bound, otherwise 0.
For example in Figure\,\ref{fig:rl}, the $bnd\_lim$ for $x_{t,1}$, $x_{t,2}$, and $x_{t,3}$ would be 0, 0, and 1, respectively.
Since the raw variable values may be unbounded, we embed them using Periodic Embedding\,(PE)\,\cite{gorishniy2022periodicembeddings}, which has proven effective for numerical features in ML tasks\,\cite{income, housing}.
PE is formulated as $\operatorname{PE}(z) = \oplus(\sin(\tilde{z}), \cos(\tilde{z}))$, where $\tilde{z} = \left[2 \pi w_1 z, \ldots, 2 \pi w_k z\right]$ with scalar $z$ and trainable $w_i$.
% $\oplus(\cdot)$ is concatenation, with scalar $z$ and trainable $w_i$.
Each final variable node is obtained by concatenating its structural information, $bnd\_lim$, and variable value embedding, followed by a projection to the hidden size.

The critic approximates an expected return using reward context tokens, which consist of a phase indicator $phase_t$, a PE-encoded $obj_t$, and a scaled $\mathbf{f}_t$.
The reward context is essential because the total reward is determined by changes in solution quality and feasibility specific to each phase\,(see Eq.\,\ref{eq:total_reward}).
We also introduce \textit{phase-separated actor} and \textit{critic layers} to align with the two-phase reward designs.
In both phases, all other layers share parameters, and the input/output configuration is consistent.
$\mathbf{f}_t$ is scaled by $\sqrt{|\mathbf{b}|+|\mathbf{b}-\mathbf{lhs}_t|}$ and projected to the hidden size.
The reward context tokens and variable nodes comprise a sequence of length $\tilde{n}+3$, which is fed into the Transformer encoders.

\vspace{-0.1cm}
\subsection{Feasibility-Aware Search Strategy} \label{Solution Search Strategy}
\vspace{-0.1cm}

% Algorithm \ref{alg:var_selection} shows the procedure of the variable selection.
To explore the solution space efficiently, RL-SPH selects $\tilde{n} (= p + q)$ changeable variables that are more likely to improve feasibility at each step $t$.
% At each timestep $t$, RL-SPH selects $\tilde{n} = p + q$ changeable variables to explore the solution space.
The selection process comprises two steps, varying by phase.
In $phase1$, RL-SPH stochastically selects $p$ seed variables that frequently appear in violated linear constraints.
Then, it selects the top $q$ neighbor variables that appear most often with the seed variables in the same constraints.
For example, in Figure\,\ref{fig:rl}, constraint $\mathbf{a}_1$ is infeasible at step $t$; thus, $x_1$ and $x_2$ are candidates for selection.
If $p = q = 1$ and $x_2$ is selected as the seed, then $x_1$ is selected as a neighbor since it shares the same constraint with the seed.
Consequently, $x_1$ and $x_2$ become changeable, while $x_3$ remains fixed.
In $phase2$, RL-SPH stochastically selects $p$ seed variables with a relatively low risk of abruptly depleting the available slack within the linear constraints.
Neighbor selection in $phase2$ follows the same procedure as in $phase1$.
To limit the growth of input size in ILP-GT, we set $p$ and $q$ to $\left\lceil\log_2n\right\rceil$.
The pseudo-code appears in Algorithm~\ref{alg:var_selection},  Appendix\,\ref{appendix:var_selection}.

Algorithm~\ref{alg:solution_search} outlines the overall procedure of our search strategy, which is used during both training and testing.
% Algorithm~\ref{alg:solution_search} outlines the overall procedure of our search strategy.
RL-SPH first selects the variables to be changed with the feasibility-aware selection\,(Line 1), which are then used by the actor $\pi_\theta$ to predict $\mathcal{A}_t$\,(Line 2).
Based on $\mathcal{A}_t$, the agent obtains the next observation $\mathcal{S}_{t+1}$\,(Lines 3-4).
The agent receives a reward determined by the quality of $\mathbf{x}_{t+1}$ and the degree of feasibility improvement\,(Line~5).
If $\textbf{x}_{t+1}$ is a better feasible solution, both $obj_b$ and $\textbf{x}_b$ are updated\,(Lines 6-8).
To guide exploration, we restrict movement to areas where further search is unnecessary\,(e.g., bound violations)\,(Lines 9–11).
In $phase1$, the agent explores the solution space freely unless variable bounds are violated\,(Lines 9–10).
In $phase2$, movements are rolled back unless a better feasible solution is discovered\,(Lines 9–10).
% In $phase1$, the agent explores freely the solution space unless variable bounds are violated\,(Lines 9-10).
% In $phase2$, movements are rolled back unless a better feasible solution is discovered\,(Lines 11-12).
The algorithm returns $\mathcal{R}_{t,\text{total}}$, $\mathcal{S}_{t+1}$, $\textbf{x}_b$, and $obj_b$ for the next step\,(Line 12).
Appendix\,\ref{appendix:novelty_search} provides a discussion of our search strategy.

\begin{algorithm}[h]
\caption{Feasibility-aware solution search of RL-SPH}
\label{alg:solution_search}
\textbf{Input:} instance $M$, actor layer $\pi_\theta$, $\texttt{phase}_t$, observation $\mathcal{S}_t =\,(\mathbf{x}_{t}, \mathbf{f}_{t}, obj_{t})$, incumbent solution $\textbf{x}_b$, and incumbent value $obj_{b}$  \\
% \textbf{Parameter:} Local region size $\Delta$ \\
\textbf{Output:} reward $\mathcal{R}_{t,\text{total}}$, new observation $\mathcal{S}_{t+1}$, $\mathbf{x}_b$, and $obj_b$

\begin{algorithmic}[1]
    % \STATE $obj_{t+1} \gets \textbf{c}^\top \textbf{x}_{t+1}$
    \STATE $\tilde{\mathcal{S}}_t \gets \operatorname{select\_vars}(M, \mathcal{S}_t, \texttt{phase}_t)$ \hfill \COMMENT{See Alg. \ref{alg:var_selection}} 
    \STATE $\mathcal{A}_t \gets \pi_\theta(\tilde{\mathcal{S}}_t, \texttt{phase}_t)$ 
    \STATE $\mathbf{x}_{t+1} \gets \operatorname{move}(\mathbf{x}_{t}, \mathcal{A}_t)$ 
    \STATE \( \mathcal{S}_{t+1} \gets \operatorname{observe}(M, \mathbf{x}_{t+1}) \)
    \STATE $\mathcal{R}_{t,\text{total}} \gets \operatorname{reward}(M,\mathcal{S}_{t+1}, \mathcal{S}_{t}, obj_{b}, \texttt{phase}_t)  $ 
    \IF{$\textbf{x}_{t+1} \in \mathcal{F}$ \AND $obj_{t+1} < obj_b$}
        \STATE $obj_b \gets obj_{t+1}$
        \STATE $\textbf{x}_b \gets \textbf{x}_{t+1}$
    \ELSIF{($\texttt{phase}_t = 1$ \AND $x_i \notin [l_i, u_i] , \, \exists i$) \\ \OR $\texttt{phase}_t = 2$}
        \STATE $\mathcal{S}_{t+1} \gets \mathcal{S}_{t}$
    % \ELSIF{\texttt{phase = 1} \AND $x_i \notin [l_i, u_i] , \, \exists i$}
    %     \STATE $\mathcal{S}_{t+1} \gets \mathcal{S}_{t}$
    % \ELSIF{\texttt{phase = 2}}
    %     \STATE $\mathcal{S}_{t+1} \gets \mathcal{S}_{t}$
    \ENDIF
\STATE \textbf{return} $\mathcal{R}_{t,\text{total}}$, $\mathcal{S}_{t+1}$, $\textbf{x}_b$, $obj_b$
\vspace{-0.1cm}
\end{algorithmic}
\end{algorithm}

% \vspace{-0.43cm}
\subsection{Learning Algorithm} \label{Learning Algorithm}
\vspace{-0.1cm}

During training, our agent begins with an initial solution $\mathbf{x}_0$ obtained via either \textit{LP-relaxation} or \textit{random assignment}.
If a more sophisticated initialization method is available, it can be readily integrated into RL-SPH.
At timestep $t$, the agent observes $\tilde{\mathcal{S}}_t$ with $\tilde{n}$ variables. 
Afterwards, it selects $\mathcal{A}_t$ using $\pi_\theta(\mathcal{A}_t \mid \tilde{\mathcal{S}}_t, phase_t)$.
Upon executing $\mathcal{A}_t$, the ILP environment returns the $\mathcal{R}_{t,\text{total}}$ and $\mathcal{S}_{t+1}$.
The actor is trained to encourage actions that yield $\mathcal{R}_{t,\text{total}} > V_\theta(\mathcal{S}_t,phase_t)$ and to discourage those that yield lower rewards. %$\mathcal{R}_{t,\text{total}}$.
The critic is trained to minimize the gap between $\mathcal{R}_{t}$ and $V_\theta(\mathcal{S}_t,phase_t)$ to provide accurate feedback to the actor.
The agent stays in $phase1$ for a predefined number of steps to ensure sufficient training, even after finding the first feasible solution.
Once the step limit is reached, it moves to a new instance.
Appendix\,\ref{appendix:training_algo} details the full training procedure.

\vspace{-0.05cm}
\section{Experiments}\label{Experiments}
\vspace{-0.05cm}
We validate the feasibility performance of RL-SPH through comprehensive experiments. 
First, we compare RL-SPH with widely-adopted start primal heuristics to demonstrate that it achieves the primary goal of quickly finding high-quality feasible solutions.
Second, we assess RL-SPH against representative E2EPH methods, alongside the top-performing baselines from the first experiment.
Third, we verify that our reward formulation effectively induces a directionality toward feasibility.
Fourth, we evaluate the generalizability of RL-SPH in cross-problem settings.
Fifth, we conduct ablation studies to assess the effectiveness of our feasibility-aware search strategy and ILP-GT components.
Finally, Appendix\,\ref{appendix:additional-exp} provides supplemental analyses, including tests on MIPLIB benchmarks, integration with ILP solvers, hyperparameter analyses, and qualitative analyses.

\vspace{-0.1cm}
\subsection{Experimental Setup} \label{Experimental Setup}
\vspace{-0.1cm}
\subsubsection{Benchmarks} \label{Benchmarks}
\vspace{-0.1cm}
% \vspace{-0.1cm}

We conduct experiments on five NP-hard ILP benchmarks commonly used in prior works\,\cite{gasse2019exact, qi2021smartpump, PAS, huang2023searching}.
For minimum vertex cover\,(MVC), we generate instances based on the Barabási-Albert random graph models\,\cite{albert2002statistical}, with 3,000 nodes, yielding 3,000 variables and 11,931 constraints on average.
We generate instances for independent set\,(IS) using the same model as MVC with 1,500 nodes.
For set covering\,(SC)\,\cite{balas1980set}, we generate instances with 3,000 variables and 2,000 constraints.
For combinatorial auction\,(CA)\,\cite{leyton2000towards}, instances are generated with 4,000 items and 2,000 bids, resulting in 4,000 variables and 2,677 constraints on average.
We generate instances with 2,000 integer variables and 2,000 constraints, adapted from \citet{qi2021smartpump}, denoted as non-binary integers\,(NBI).
For each dataset, we generated 1,000 training instances and 100 test instances.
Additional details on the datasets are given in Appendix~\ref{appendix:datasets}.

\vspace{-0.1cm}
\subsubsection{Baselines} \label{baselines}
\vspace{-0.1cm}
In the first experiment, we evaluate RL-SPH against baselines from four major SPH classes: FP, RENS, diving, and rounding\,(see Appendix\,\ref{Start primal heuristics for ILP}).
For comprehensive coverage, we group 15 diving and 6 rounding variants into the Diving Heuristic Family\,(DHF) and Rounding Heuristic Family\,(RHF) baselines, respectively\,(see Appendix\,\ref{appendix:baselines}).
We use the built-in implementations from the open-source ILP solver SCIP\,(v8.1.0), with presolving and branching disabled to isolate heuristic performance.
Each baseline exclusively enables its corresponding heuristics with default settings.
All randomization parameters in SCIP are set to 0 for reproducibility.
Each baseline runs until its termination condition is met, with a 1,000-second time limit.
RL-SPH, which has no predefined termination condition, stops when all baselines complete their search.
If RL-SPH terminates without a feasible solution, it is recorded as a failure.
Unless otherwise stated, RL-SPH uses LP initialization, and its termination criteria follow those of the first experiment.

In the second experiment, we compare RL-SPH with three representative E2EPH methods: PAS\,\cite{PAS}, DDIM\,\cite{kdd_complete}, and DiffILO\,\cite{iclr_complete}.
While DDIM and DiffILO are our primary baselines due to their explicit use of feasibility-related metrics, PAS serves as a proxy for the other E2EPH methods\,\cite{huang2024contrastive, liu2025apollo} that lack such metrics.
The E2EPH baselines were trained and evaluated using their respective benchmark-specific configurations.
In contrast, RL-SPH utilized a consistent hyperparameter configuration across all datasets, as detailed in Appendix\,\ref{appendix:Method_cofig}.
RL-SPH took an average of 30 minutes to train, a 14.7$\times$ speedup over the E2EPH baselines\,(see Appendix\,\ref{appendix:E2EPH-exp}).

\vspace{-0.1cm}
\subsubsection{Metrics} \label{Metrics}
% \vspace{-0.1cm}
We use four evaluation metrics for the main experiment: the \textit{primal gap}\,(PG)\,\cite{berthold2013measuring, huang2023searching, canturk2024scalable}, the \textit{primal integral}\,(PI)\,\cite{berthold2013measuring, ml4co_comp}, the \textit{feasibility rate}\,(FR), and the \textit{first feasible solution time} (FT).
PG quantifies how close a method's incumbent value $obj_b$ is to the best-known solution\,($BKS$), computed as $\operatorname{PG}(obj_b)=\frac{|obj_b-BKS|}{\operatorname{max}(|obj_b|,|BKS|,\epsilon)}$.
% PG quantifies how close a method's incumbent value $obj_b$ is to the best-known solution\,($BKS$), computed as $\operatorname{PG}(obj_b)=\frac{|obj_b-BKS|}{\operatorname{max}(|obj_b|,|BKS|,\epsilon)} \times 100$.
PI measures how quickly the incumbent improves toward $BKS$ over time, defined as $\operatorname{PI}(T)=\sum_{t=1}^T \operatorname{PG}(obj_t)$,
where $\operatorname{PG}(obj_t)=1$ if no feasible solution is available at actual wall-clock time\,$t$\,(in seconds).
FR measures the ratio of instances in which a feasible solution is obtained.
FT records the time taken to obtain the first feasible solution, measured in seconds.
PG, PI, and FT are averaged only over instances where at least one feasible solution is found.

% \vspace{-0.1cm}
\subsection{Comparison with Start Primal Heuristics} \label{Results}
\begin{figure*}[h]
    \vspace{-0.1cm}
  \centering
  \includegraphics[width=1\linewidth]{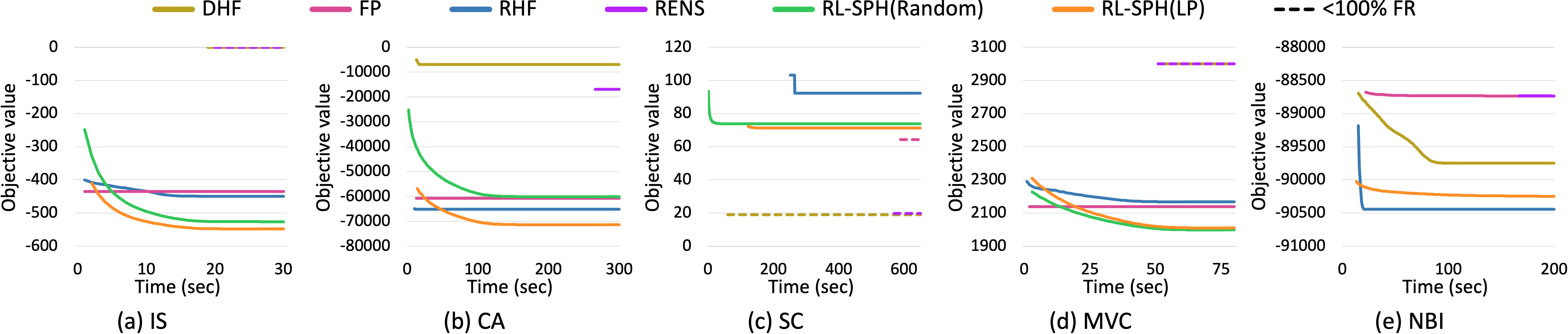}
  \vspace{-0.4cm}
  \caption{Illustration of the objective values of the compared methods over time\,(in seconds). Lines indicate valid averaged incumbent objective values, which are computed only when an incumbent objective value exists for all instances at a given time point. Dashed lines indicate that the FR is not 100\%, and the averaging includes only instances for which at least one feasible solution was found.}
  \vspace{-0.3cm}
  \label{fig:testing_curve}
\end{figure*}

% \vspace{-0.1cm}
Table\,\ref{tab:first_exp} summarizes the evaluation results across FR, PG, PI, and FT, with $BKS$ identified among all methods.
RL-SPH achieved 100\% FR on all benchmarks, demonstrating its effectiveness in learning feasibility.
Only RHF also achieved 100\% FR on all benchmarks.
We denote RL-SPH initialized with LP as RL-SPH(LP), and RL-SPH initialized with a randomly generated solution as RL-SPH(Random).
Compared to the baselines with 100\% FR, RL-SPH(LP) achieved a 28.6$\times$ lower PG, a 2.6$\times$ lower PI, and a 2.5$\times$ faster FT on average.
While RL-SPH(LP) exhibits a higher PG than RHF only on NBI, its comparable PI on NBI indicates that it finds high-quality feasible solutions at an earlier stage\,(see Figure\,\ref{fig:testing_curve}).
These results demonstrate RL-SPH(LP)'s superiority in rapidly discovering high-quality feasible solutions, confirming that it achieves the primary goal.

Notably, RL-SPH(Random) showed comparable performance to RL-SPH(LP), which begins from an LP-feasible solution.
Furthermore, RL-SPH(Random) found the first feasible solution within 2 seconds for all datasets, attaining a 180$\times$ average speedup in FT compared to 100\% FR baselines.
Random initialization enables an early start, accelerating the agent's exploration for feasible solutions without waiting for LP solving.
These results suggest that RL-SPH can find feasible solutions even from lower-quality initializations, motivating the development of initialization strategies that are more refined than random assignment but faster than LP-relaxation.
RENS and DHF yielded similar results on IS and MVC, 
as both start from the same LP-feasible solution due to the same random seed for reproducibility. 
Moreover, the coefficient matrix $\mathbf{A}$ in these datasets is extremely sparse\,(density of 0.13\% and 0.07\%), resulting in LP-feasible solutions with few fractional values.
In such cases, fixing variables has limited impact on the rest of the problem, leading both methods to produce similar solutions.

\setlength{\tabcolsep}{12pt} 
\renewcommand{\arraystretch}{0.7}
\begin{table*}[h]
\centering
\vspace{-0.1cm}
\caption{Performance comparison against SPH baselines on five datasets. \textbf{Bold} and \underline{underline} are used to indicate the best and second-best methods among those with 100\% FR. Improvements are averaged over Dataset-Baseline pairs with 100\% FR.}
\vspace{-0.1cm}
\label{tab:first_exp}
\resizebox{\linewidth}{!}{%
\begin{tabular}{ccccccccc}
\specialrule{1pt}{0pt}{2pt}
\multirow{2}{*}{Dataset} &
  \multirow{2}{*}{Metric} &
  \multirow{2}{*}{DHF} &
  \multirow{2}{*}{FP} &
  \multirow{2}{*}{RHF} &
  \multirow{2}{*}{RENS} &
  \multicolumn{2}{c}{\textbf{RL-SPH\,(Ours)}} \\ 
\cmidrule[0.5pt](l{0.3em}r{0.3em}){7-8}
                     &     &            &            &             &             & w/ Random      & w/ LP          \\ 
\specialrule{1pt}{2pt}{2pt}   
\multirow{4}{*}{IS}  & FR  $(\%) \uparrow$   & 89         & 100        & 100         & 89          & 100         & 100         \\
                     & PG  $(\%) \downarrow$ & 99.98±0.05 & 20.60±2.10 & 18.09±2.74  & 99.98±0.05  & \underline{4.10}±2.25   & \textbf{0.14}±0.59   \\
                     & PI $\downarrow$       & 19.1±2.3   & 4.8±0.7    & 4.9±0.7     & 19.1±2.3    & \underline{3.8}±0.4     & \textbf{2.5}±0.2     \\ 
                     & FT (sec) $\downarrow$ & 11.94±2.27	& \underline{0.22}±0.04	& \textbf{0.10}±0.00	& 11.94±2.25	& 0.27±0.07	& 0.55±0.07     \\ 
\cmidrule[0.5pt](l{0.3em}r{0.3em}){1-8} 
\multirow{4}{*}{CA}  & FR  $(\%) \uparrow$   & 100        & 100        & 100         & 100         & 100         & 100         \\
                     & PG  $(\%) \downarrow$ & 90.61±1.68 & 18.09±9.64 & \underline{12.02}±9.97  & 81.70±30.19 & 19.04±8.20  & \textbf{3.82}±9.77   \\
                     & PI $\downarrow$       & 105.7±27.9 & 31.5±15.6  & \underline{23.0}±15.5   & 112.9±27.7  & 36.7±14.9   & \textbf{21.9}±13.8   \\ 
                     & FT (sec) $\downarrow$ & 10.06±0.90	&10.66±0.70	&\underline{8.03}±0.51	&102.17±28.38	&\textbf{0.41}±0.08	&8.26±0.72    \\ 
\cmidrule[0.5pt](l{0.3em}r{0.3em}){1-8} 
\multirow{4}{*}{SC}  & FR  $(\%) \uparrow$   & 1          & 99         & 100         & 3           & 100         & 100         \\
                     & PG  $(\%) \downarrow$ & 0.00±0.00  & 0.34±3.11  & 29.98±12.38 & 0.00±0.00   & \underline{12.78}±16.25 & \textbf{9.67}±15.41  \\
                     & PI $\downarrow$       & 60.0±0.0   & 347.5±73.5 & 321.5±111.0 & 339.3±212.8 & \textbf{101.3}±136.7 & \underline{168.0}±124.9 \\ 
                     & FT (sec) $\downarrow$ & 59.50±0.00	& 344.75±68.57	& 104.09±16.97	& 338.50±212.62	& \textbf{0.19}±0.07	& \underline{95.43}±7.39     \\ 
\cmidrule[0.5pt](l{0.3em}r{0.3em}){1-8} 
\multirow{4}{*}{MVC} & FR  $(\%) \uparrow$   & 36         & 100       & 100       & 36         & 100       & 100         \\
                     & PG  $(\%) \downarrow$ & 33.67±0.70 & 6.76±1.15 & 8.01±1.20 & 33.67±0.70 & \textbf{0.23}±0.45 & \underline{0.81}±0.86 \\
                     & PI $\downarrow$       & 46.0±5.3   & 5.2±1.2   & 6.1±1.2   & 42.8±5.2   & \textbf{4.2}±0.5   & \underline{5.0}±0.6     \\ 
                     & FT (sec) $\downarrow$ & 40.37±7.05	& 1.17±0.24	& \textbf{0.20}±0.04	& 35.76±6.95	& 1.63±0.07	& \underline{1.55}±0.08    \\ 
\cmidrule[0.5pt](l{0.3em}r{0.3em}){1-8} 
\multirow{4}{*}{NBI} & FR  $(\%) \uparrow$   & 100        & 100        & 100        & 100          & 100        & 100         \\
                     & PG  $(\%) \downarrow$ & 0.77±0.64  & 1.89±0.08  & \textbf{0.00}±0.00  & 1.89±0.08    & 28.32±2.84 & \underline{0.22}±0.06 \\
                     & PI $\downarrow$       & 14.3±1.3   & 20.8±2.1   & \textbf{11.3}±1.1   & 108.5±25.7   & 57.5±16.7  & \textbf{11.3}±0.9     \\ 
                     & FT (sec) $\downarrow$ & 12.15±0.92 & 17.16±1.60 & 10.71±0.99 & 106.35±25.44 & \textbf{0.10}±0.07  & \underline{10.32}±0.71    \\ 
\specialrule{1pt}{2pt}{0pt}
\end{tabular}%
}
% \vspace{-0.1cm}
\end{table*}

% \vspace{-0.1cm}
\subsection{Comparison with SOTA E2EPH} \label{E2EPH-exp}
% \vspace{-0.1cm}

As shown in Table\,\ref{tab:diffilo-exp}, only RL-SPH and RHF consistently achieved a 100\% FR across all datasets, outperforming the other E2EPH baselines.
PAS failed to find a feasible solution on CA and IS, aligning with the findings reported in the DiffILO paper. 
DDIM exhibited higher FT than both RHF and all RL-SPH variants across all datasets. 
This latency stems not only from the LP-solving requirement for input feature construction but also from its slow inference speed.
Specifically, DDIM conducted an average of 165, 149, and 76 trials for solution search on SC, CA, and IS within the time limit, respectively, yielding 0.5 to 2.5 trials per second.
% SC: 34183, CA: 38915, and IS: 3985
Moreover, DDIM attained higher PG than RHF on SC and CA, indicating inferior solution quality.
Coupled with slow inference, this quality gap led to a 1.8$\times$ higher PI for DDIM than for RHF.
Thus, DDIM shows limitations in fulfilling the goal of quickly generating high-quality feasible solutions.

DiffILO exhibited 0\% FR on SC and CA despite making an average of 919,470 and 574,240 trials of a binomial distribution, respectively.
We attribute the FR performance gap between RL-SPH and DiffILO to the prioritization between the objective function and the constraints.
DiffILO is trained using a soft-constrained loss function, where a penalty coefficient determines the relative influence of the objective and constraints, meaning that no explicit priority is enforced. 
Consequently, DiffILO’s 0\% FR on SC and CA can be attributed to the objective function dominating the constraints within its loss formulation.
In contrast, RL-SPH enforces a clear priority between them: a positive reward for the objective value\,(i.e., $+\Delta obj_t$) is granted only when feasibility improves\,(Case 3 in Eq.~\ref{eq:reward_phase1_obj}).

\begin{figure*}[h]
    \vspace{-0.15cm}
  \centering
  \includegraphics[width=1\linewidth]{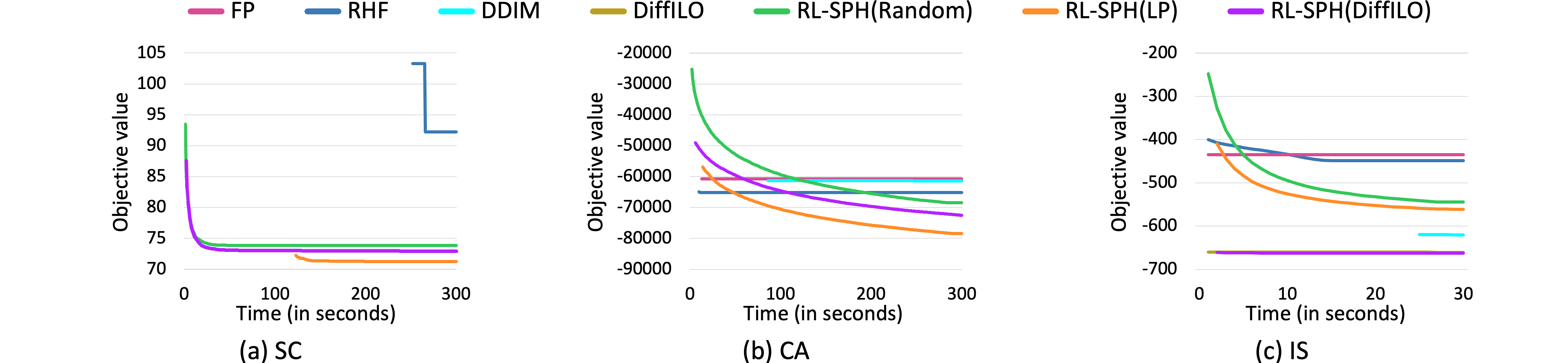}
  \vspace{-0.45cm}
  \caption{Illustration of the objective values of the compared methods over time (in seconds). Only cases with FR $=100$\% are displayed.}
  \vspace{-0.4cm}
  \label{fig:testing_curve_diffilo}
\end{figure*}

Interestingly, RL-SPH was able to convert DiffILO’s initially infeasible solutions into feasible ones on SC and CA.
As shown in Figure\,\ref{fig:testing_curve_diffilo}, the curve for RL-SPH(DiffILO) lies between those of RL-SPH(Random) and RL-SPH(LP) on SC and CA. 
This indicates that the initial solutions provided by DiffILO offer a better starting point than random initialization, but a weaker one compared to LP-relaxation.
Given that learning-based models provide rapid solution approximations, a refined inference approach can outperform LP initialization, as demonstrated by RL-SPH(DiffILO) on IS.

\setlength{\tabcolsep}{8pt} 
\renewcommand{\arraystretch}{0.7}
\begin{table*}[h]
\centering
% \vspace{-0.1cm}
\caption{Performance comparison against representative E2EPH methods on SC, CA, and IS, which are the intersection of the benchmarks used in DDIM, DiffILO, and our study. Since PAS does not provide hyperparameters for the SC benchmark, we exclude it from this experiment. \textit{fail} indicates that the metrics cannot be computed because no feasible solution was obtained\,(i.e., FR $ = 0$\%). The time limit is set to 300, 300, and 30 seconds for SC, CA, and IS, respectively.}
\vspace{-0.1cm}
\label{tab:diffilo-exp}
\resizebox{\linewidth}{!}{%
\begin{tabular}{cccccccccc}
\specialrule{1pt}{0pt}{2pt}
\multirow{2}{*}{Dataset} &
  \multirow{2}{*}{Metric} &
  \multicolumn{2}{c}{Classical SPH baselines} &
  \multicolumn{3}{c}{E2EPH baselines}  &
  \multicolumn{3}{c}{\textbf{RL-SPH\,(Ours)}} \\ 
\cmidrule[0.5pt](l{0.3em}r{0.3em}){3-4}
\cmidrule[0.5pt](l{0.3em}r{0.3em}){5-7}
\cmidrule[0.5pt](l{0.3em}r{0.3em}){8-10}
                     &    &FP   &RHF    &PAS       &DDIM      &DiffILO & w/ Random      & w/ LP    & w/ DiffILO      \\ 
\specialrule{1pt}{2pt}{2pt}   
\multirow{4}{*}{SC}  & FR  $(\%) \uparrow$    & 5            & 100          &   \multirow{4}{*}{--}     & 99 & 0   & 100       & 100        & 100       \\
                     & PG  $(\%) \downarrow$  & 0.00±0.00    & 23.43±4.02   &    &  59.30±4.15 & \multirow{3}{*}{\textit{fail}} & 4.65±7.53 & \textbf{1.17}±1.90  & \underline{3.47}±7.61 \\
                     & PI $\downarrow$        & 243.6±93.8   & 160.2±17.8   &    &  232.3±7.2 &     & \underline{16.0}±22.4 & 100.0±8.2   & \textbf{13.5}±22.6 \\
                     & FT (sec) $\downarrow$  & 242.62±93.77 & 104.09±16.97 &    &  133.00±6.52 &     & \textbf{0.19}±0.07 & 95.43±7.39 & \underline{1.15}±0.25 \\
\cmidrule[0.5pt](l{0.3em}r{0.3em}){1-10} 
\multirow{4}{*}{CA}  & FR  $(\%) \uparrow$    & 100        & 100         &    0     &  100  & 0 & 100        & 100       & 100       \\
                     & PG  $(\%) \downarrow$   & 22.81±4.66 & 17.17±2.74 &    \multirow{3}{*}{\textit{fail}}    &  22.15±2.65  & \multirow{3}{*}{\textit{fail}}  & 13.05±2.07 & \textbf{0.33}±2.33 & \underline{7.78}±2.15 \\
                     & PI $\downarrow$         & 77.1±13.5  & 58.6±8.0   &     &  122.5±8.5 &   & 71.5±5.5   & \textbf{33.3}±4.7  & \underline{52.9}±5.8  \\
                     & FT (sec) $\downarrow$   & 10.66±0.70 & 8.03±0.51  &     &  71.42±6.61 &   & \textbf{0.41}±0.08  & 8.26±0.72 & \underline{4.85}±0.27 \\
\cmidrule[0.5pt](l{0.3em}r{0.3em}){1-10} 
\multirow{4}{*}{IS}  & FR  $(\%) \uparrow$    & 100        & 100        &   0       & 100    & 100       & 100        & 100        & 100       \\
                     & PG  $(\%) \downarrow$  & 34.32±1.29 & 32.24±1.94 &   \multirow{3}{*}{\textit{fail}}      &  6.34±1.07   & \underline{0.35}±0.50 & 17.87±1.86 & 15.35±1.53 & \textbf{0.00}±0.00 \\
                     & PI $\downarrow$        & 11.0±0.4   & 10.8±0.5   &    &  21.2±1.6  & \underline{1.1}±0.2    & 8.5±0.5    & 7.1±0.4    & \textbf{1.0}±0.2   \\
                     & FT (sec) $\downarrow$  & \underline{0.22}±0.04&\textbf{0.10}±0.00&  & 20.11±1.74 & 0.42±0.16 & 0.27±0.07  & 0.55±0.07  & 0.66±0.23 \\

\specialrule{1pt}{2pt}{0pt}
\end{tabular}%
}
% \vspace{-0.3cm}
\end{table*}

Meanwhile, the E2EPH baselines do not support integer variables, making them inapplicable to NBI.
This limitation highlights the need for further research on solution approximation methods tailored for integer variables.
Notably, RL-SPH(DiffILO) was trained with LP initialization rather than with solutions generated by DiffILO, demonstrating its adaptability to an alternative initialization method.
Further comparative analysis is provided in Appendix\,\ref{appendix:E2EPH-exp}.

\vspace{-0.05cm}
\subsection{Analysis of Directionality Toward Feasibility} \label{Directionality}
\vspace{-0.1cm}

To bridge the gap between the theoretical analysis in Proposition\,\ref{prop:feasibility} and practical performance, we provide empirical evidence in Table\,\ref{tab:directionality}.
The trained RL-SPH policy adheres to the prescribed condition (i.e., $\mathcal{R}_{t,\text{const}} > 0$ and $\mathcal{R}_{t,\text{bound}} = 0$ during $phase1$) in 95.2\% of the trajectories, averaged across test datasets. 
The trained policy maintains steady progression toward feasibility, recovering from temporary stagnations in the remaining 4.8\%.
This strong directionality helps explain why RL-SPH consistently achieves 100\% FR across five benchmarks. 
These results emphasize that our reward formulation successfully induces the reward-maximizing behavior that guides toward feasibility.
Appendix\,\ref{appendix:phase-analysis} provides visualizations that illustrate this behavior.

\setlength{\tabcolsep}{7.5pt} 
\renewcommand{\arraystretch}{0.7}
\begin{table}[h]
\centering
\vspace{-0.1cm}
\caption{Empirical adherence rates across five benchmarks.}
\vspace{-0.1cm}
\label{tab:directionality}
\resizebox{\columnwidth}{!}{%

\begin{tabular}{cccccc}
\specialrule{1pt}{0pt}{2pt}
Metric        & IS     & CA     & SC     & MVC    & NBI    \\
\cmidrule[0.3pt](l{0.3em}r{0.3em}){1-6} 
Adherence rate\,(\%) $\uparrow$ & 98.7 & 91.5 & 98.4 & 99.1 & 88.4 \\
\specialrule{1pt}{2pt}{0pt}
\end{tabular}
}
\vspace{-0.1cm}
\end{table}

% \vspace{-0.1cm}
\subsection{Evaluation of Cross-Problem Generalization} \label{main:cross-problem}
% \vspace{-0.05cm}

Table\,\ref{tab:cross-problem} presents the performance of RL-SPH in cross-problem settings using IS and MVC.
In this experiment, we matched the number of variables in MVC to that of IS\,($n=1,500$) to control for factors other than the problem type.
RL-SPH trained solely on IS failed to produce any feasible solution on MVC test instances, and vice versa. 
However, a joint RL-SPH model trained on both problem types achieved a 100\% FR on each test set. 
Moreover, the joint model outperformed RHF—the strongest baseline in terms of FR\,(see Tables\,\ref{tab:first_exp} and \ref{tab:diffilo-exp})—in finding better incumbent values.
Notably, the joint model also obtained feasible solutions for 17 diverse instances from the MIPLIB benchmark\,\cite{MIPLIB}, which differ markedly from the training instances in terms of size\,(up to 67$\times$ more variables), density\,(ranging from 0.1\% to 100\%), and variable types including integers (see Appendix~\ref{appendix:miplib}).

\setlength{\tabcolsep}{2.4pt} 
\renewcommand{\arraystretch}{0.75}
\begin{table}[h]
  \centering
  % \vspace{-0.1cm}
  \caption{Cross-problem performance of RL-SPH.}
  \vspace{-0.1cm}
  \label{tab:cross-problem}
\resizebox{\columnwidth}{!}{%
  \begin{tabular}{cccccc}
\specialrule{1pt}{0pt}{2pt}
\multirow{2}{*}{Test set} & \multirow{2}{*}{Metric} & \multirow{2}{*}{RHF} & \multicolumn{3}{c}{RL-SPH trained on} \\ 
\cmidrule[0.5pt](l{0.3em}r{0.3em}){4-6} 
                              &                         &                           & IS         & MVC        & IS+MVC      \\ 
\specialrule{1pt}{2pt}{2pt}
\multirow{3}{*}{IS}           & FR $(\%) \uparrow$               & 100                       & 100        & 0          & 100         \\
                              & PG $(\%) \downarrow$               & 17.65±2.87                & 0.00±0.00  & \multirow{2}{*}{\textit{fail}}        & 15.55±2.82  \\
                              & PI $\downarrow$                    & 4.6±0.7                   & 2.4±0.2    &         & 5.1±0.6     \\ 
\cmidrule[0.5pt](l{0.3em}r{0.3em}){1-6} 
\multirow{3}{*}{MVC}          & FR $(\%) \uparrow$               & 100                       & 0          & 100        & 100         \\
                              & PG $(\%) \downarrow$               & 9.06±1.58                 & \multirow{2}{*}{\textit{fail}}        & 0.00±0.00  & 7.13±1.56   \\
                              & PI $\downarrow$                    & 2.8±0.4                   &         & 1.8±0.2    & 2.8±0.4     \\ 
\specialrule{1pt}{2pt}{0pt}
\end{tabular}
    }
    % \vspace{-0.1cm}
  % \vspace{-0.8em}  % optional bottom spacing
\end{table}

\setlength{\tabcolsep}{8pt} 
\renewcommand{\arraystretch}{0.7}
\begin{table*}[h]
\centering
\vspace{-0.1cm}
\caption{Ablation study on our solution search strategy. Uniform denotes variable selection uniformly at random, and F-aware refers to the feasibility-aware variable selection. PG$_f$ is a variant of PG in which the incumbent value $obj_b$ is replaced by the first feasible value, measuring the quality of the first feasible solution.
$\mathcal{T}$ represents the number of search steps to find the first feasible solution.}
\vspace{-0.1cm}
\label{tab:ablation_search}
\resizebox{\linewidth}{!}{%
\begin{tabular}{ccccccccccc}
\specialrule{1pt}{0pt}{2pt}
\multirow{2}{*}{Metric} & \multicolumn{2}{c}{IS}           & \multicolumn{2}{c}{CA}        & \multicolumn{2}{c}{SC}           & \multicolumn{2}{c}{MVC}           & \multicolumn{2}{c}{NBI}                \\ 
\cmidrule[0.5pt](l{0.3em}r{0.4em}){2-3} 
\cmidrule[0.5pt](l{0.3em}r{0.4em}){4-5} 
\cmidrule[0.5pt](l{0.3em}r{0.4em}){6-7} 
\cmidrule[0.5pt](l{0.3em}r{0.4em}){8-9} 
\cmidrule[0.5pt](l{0.3em}r{0.4em}){10-11} 
                        & Uniform    & F-aware             & Uniform & F-aware             & Uniform    & F-aware             & Uniform     & F-aware             & Uniform           & F-aware            \\ 
\specialrule{1pt}{2pt}{2pt}
FR $(\%) \uparrow$               & 100        & 100                 & 0       & 100                 & 100        & 100                 & 100         & 100                 & 100               & 100                \\
PG$_f (\%) \downarrow$              & 95.3±1.6 & \textbf{41.4}±1.9 & \multirow{2}{*}{\textit{fail}}     & \textbf{31.4}±4.4 & 52.3±4.8 & \textbf{39.0}±4.3 & 33.8±0.7  & \textbf{16.0}±0.9 & 0.6±0.2         & \textbf{0.3}±0.1 \\ 
$\mathcal{T} \downarrow$                    & 399±98 & \textbf{36}±2   &         & \textbf{59}±7   & 6±1    & \textbf{4}±1    & 613±109 & \textbf{68}±2   & \textbf{27}±6 & 46±11          \\
\specialrule{1pt}{2pt}{0pt}
\end{tabular}
}
\vspace{-0.3cm}
\end{table*}

% \vspace{-0.1cm}
\subsection{Ablation Study} \label{ablation-study}
\vspace{-0.1cm}

Table\,\ref{tab:ablation_search} presents the effectiveness of our variable selection strategy across five datasets.
Across all datasets, RL-SPH consistently obtained a feasible solution with the feasibility-aware selection strategy. 
In contrast, with the uniform selection, it often struggled to find a feasible solution quickly and even failed within the 100-second time limit on CA. 
For NBI, while RL-SPH required more steps with our search strategy, it achieved a higher-quality first feasible solution.
These results demonstrate that our search strategy is effective in consistently finding high-quality feasible solutions.

Table\,\ref{tab:ablation} presents an ablation study on the ILP-GT architecture, where the time limit is set to 50 seconds.
All models were trained under the same configuration.
$\emptyset$ showed the lowest performance on both benchmarks and failed to find any feasible solution on CA.
With TE, the model achieved a 100\% FR on both benchmarks, indicating its effectiveness in learning feasibility.
Combining all components achieved the best results in PG and PI, highlighting the importance of each component.
Appendix\,\ref{appendix:Long-range} provides a qualitative analysis of the long-range variable dependencies in TE.

\setlength{\tabcolsep}{5pt} 
\renewcommand{\arraystretch}{0.7}
\begin{table}[h]
% \begin{wraptable}{r}{0.48\columnwidth}
  \centering
  % \vspace{-0.1cm}
  \caption{Ablation study on the ILP-GT architecture: RC\,(reward context), PSL\,(phase-separated layers), and TE\,(transformer encoder). $\emptyset$ denotes models without all components.}
  % \vspace{-0.1cm}
    \label{tab:ablation}
\resizebox{\columnwidth}{!}{%
    \begin{tabular}{ccccc}
\specialrule{1pt}{0pt}{2pt}
    Dataset             & Model     & FR  $(\%) \uparrow$  & PG  $(\%) \downarrow$         & PI $\downarrow$       \\ 
\specialrule{1pt}{2pt}{2pt}
    \multirow{5}{*}{IS} & $\emptyset$   & 100 & 41.23±1.95 & 21.8±0.9 \\
                        & \{RC, PSL\}  & 100 & 33.98±2.45 & 28.9±2.4 \\
                        & \{TE, PSL\}  & 100 & 1.10±1.40  & 4.4±0.7  \\
                        & \{TE, RC\}  & 100 & 12.93±1.96 & 11.7±0.9 \\
                        & \textbf{\{TE, RC, PSL\}} & \textbf{100} & \textbf{0.62}±1.08  & \textbf{4.1}±0.5  \\ 
\cmidrule[0.5pt](l{0.3em}r{0.3em}){1-5} 
    \multirow{5}{*}{CA} & $\emptyset$ & 0   & \textit{fail}    & \textit{fail}  \\
                        & \{RC, PSL\}  & 0   & \textit{fail}    & \textit{fail}  \\
                        & \{TE, PSL\}  & 100 & 3.00±2.27  & 12.4±1.1 \\
                        & \{TE, RC\}  & 100 & 1.63±1.96  & 12.5±1.0 \\
                        & \textbf{\{TE, RC, PSL\}} & \textbf{100} & \textbf{1.20}±1.53  & \textbf{11.8}±0.8 \\ 
\specialrule{1pt}{2pt}{0pt}
    \end{tabular}
    }
  \vspace{-0.1cm}  % optional bottom spacing
\end{table}

\vspace{-0.2cm}
\section{Discussion}\label{main_discussion}
% \vspace{-0.1cm}

% \subsection{Scalability} \label{main:scalability}
% \vspace{-0.2cm}

As no E2EPH consistently achieves a 100\% FR, we prioritize FR as the primary metric.
In this context, RHF serves as a stronger baseline than the E2EPH baselines\,(see Appendix\,\ref{appendix:core_focus}).
Our empirical evaluations confirm that RL-SPH significantly outperforms both widely-adopted start primal heuristics and SOTA E2EPH methods in generating high-quality feasible solutions\,(see Tables\,\ref{tab:first_exp} and \ref{tab:diffilo-exp}). 
% Our empirical evaluations confirm that RL-SPH not only surpasses the widely-adopted start primal heuristics but also significantly outperforms SOTA E2EPH methods in generating high-quality feasible solutions\,(see Tables\,\ref{tab:first_exp} and \ref{tab:diffilo-exp}). 
Our literature review also provides evidence that existing E2EPH methods struggle to achieve feasibility\,(see Appendix\,\ref{overview:ML4ILP}), thereby highlighting the clear advantages of RL-SPH.
Moreover, unlike most E2EPH methods, RL-SPH is designed to handle integer domains and attains a 100\% FR on NBI.
Notably, RL-SPH generalizes to unseen MIPLIB instances containing integer variables, despite being trained on binary-only instances\,(see Appendix\,\ref{appendix:miplib}).

Beyond strong performance, RL-SPH offers significant advantages in training efficiency.
Most E2EPH methods rely on supervised learning frameworks, necessitating near-optimal solutions as training labels that limit their practicality for NP-hard problems\,\cite{canturk2024scalable, iclr_complete}.
In contrast, RL-SPH does not require such labels. 
Remarkably, it also achieves nearly 34$\times$ faster training time than the most recent unsupervised learning–based E2EPH\,(see Appendix~\ref{appendix:training_times}).
This efficiency stems from our search strategy, which enables focusing on informative regions where reward signals are dense\,(see Appendix~\ref{appendix:training_efficiency}).

Furthermore, RL-SPH delivers robust results across five benchmarks with a consistent hyperparameter configuration, whereas prior E2EPH methods often require benchmark-specific tuning\,(see Appendix\,\ref{appendix:E2EPH-exp}).
This enhances its practicality, facilitating easier training and deployment across diverse ILP settings with minimal adjustments.
Appendix~\ref{NeuRewriter} situates this general-purpose framework within a broader research context.
RL-SPH provides a foundation for ML-based standalone solvers pursuing optimality, with further refinements (e.g., changes in action magnitude) outlined as future directions in Appendices\,\ref{appendix:novelty_search}--\ref{appendix:future_research}.

% \vspace{-0.05cm}
\section{Conclusion} \label{Conclusion}
% \vspace{-0.05cm}

We proposed RL-SPH, which independently produces high-quality feasible solutions for ILP, grounded in a theoretically established reward-feasibility alignment.
Empirically, RL-SPH attains a consistent 100\% feasibility rate across five benchmarks, even for instances involving integer variables.
% It outperforms seven primal heuristic baselines including RHF that exhibits stronger feasibility performance than SOTA E2EPH methods.
It outperforms four SPH baselines—including RHF, which exhibits stronger feasibility performance than SOTA E2EPH methods—by achieving an average 28.6$\times$ lower primal gap and 2.6$\times$ lower primal integral.
% Furthermore, the LP-free variant of RL-SPH finds the first feasible solution 180$\times$ faster than the SPH baselines.
By addressing the feasibility bottleneck, our study resolves a critical gap left open by prior learning-based heuristics for ILP.
This advance in feasibility represents a meaningful step toward addressing the major challenge of optimality in ML for CO.

\newpage

% \vspace{-0.2cm}
% \paragraph{Appendix Summary.}
% The \hyperref[toc]{appendix} includes proofs, algorithmic pseudo-code, additional preliminaries, related work, positioning relative to prior work, further explanation of the reward design, details of experimental setup, and supplementary experimental results and discussion. 

%%%%%%%%%%%%%%%%%%%%%%%%%%%%%%%%%%%%%%%%%%%%%%%%%%%%%%%%%%%%

% \bibliography{example_paper}
% \bibliographystyle{icml2026}
\section*{Acknowledgements}

This work was supported by the National Research Foundation of Korea(NRF) grant funded by the Korea government(MSIT)(No. RS-2024-00347471) and Institute of Information \& communications Technology Planning \& Evaluation (IITP) grant funded by the Korea government(MSIT) (No. 2019-0-01267, GPU-based Ultrafast Multi-type Graph Database Engine SW), and the IITP(Institute of Information \& Communications Technology Planning \& Evaluation)-ITRC(Information Technology Research Center) grant funded by the Korea government(Ministry of Science and ICT)(IITP-2025-RS-2020-II201795).

\section*{Impact Statement}

This paper presents work whose goal is to advance the field of Machine
Learning. There are many potential societal consequences of our work, none which we feel must be specifically highlighted here.

\bibliography{icml2026}
\bibliographystyle{icml2026}

%%%%%%%%%%%%%%%%%%%%%%%%%%%%%%%%%%%%%%%%%%%%%%%%%%%%%%%%%%%%

\newpage
\appendix
\onecolumn

% \addtocontents{toc}{\protect\setcounter{tocdepth}{0}}
% % Add an empty content line to force clearing
% \addtocontents{toc}{}
% \section*{Appendix}
% \setcounter{tocdepth}{2}
% \renewcommand{\contentsname}{Appendix contents}
% \phantomsection
% \label{toc}
% \tableofcontents
% \clearpage

% % --- Restore normal TOC depth ---
% \addtocontents{toc}{\protect\setcounter{tocdepth}{2}}

\section{Proof of Proposition \ref{prop:feasibility}}\label{appendix:proof_feasibility}

Proposition\,\ref{prop:feasibility} establishes the theoretical soundness of the reward formulation in achieving feasibility. 
In this section, we provide the proof of the proposition: \textit{Suppose $\mathbf{x}_t \notin \mathcal{F}$,  $\mathcal{R}_{t,\text{const}} > 0$, and $\mathcal{R}_{t,\text{bound}} = 0$ $\text{ for all } t < \mathcal{T}$. Then $\mathbf{x}_\mathcal{T} \in \mathcal{F}$.}

\subsection{Term Definitions}
\begin{itemize}[nosep, itemsep=0.5ex]
    \item $\mathcal{F}$: The set of feasible solutions (feasible region) for the given original ILP problem.
    \item $\mathbf{x}_t$: The solution of the problem at timestep $t$.
    \item $\mathcal{R}_{t,\text{bound}}$: The bound reward at timestep $t$ (see Eq. \ref{eq:reward_bound}).
    \item $\mathcal{R}_{t,\text{const}}$: The constraint reward at timestep $t$ (see Eq. \ref{eq:reward_const}).
    % \item $\mathbf{f}_t$: The feasibility state at timestep $t$.
    \item $a_{j,i}$: The coefficient of the $i$-th variable in the $j$-th linear constraint.
    \item $x_{t,i}$: The value of the $i$-th variable at timestep $t$.
    \item $f_{t,j}$: The feasibility state for the $j$-th linear constraint at timestep $t$.
    \item $lhs_{t,j}$: The left-hand side of the $j$-th linear constraint at timestep $t$.
    \item $b_{j}$: The right-hand side of the $j$-th linear constraint.
\end{itemize}

\subsection{Preliminaries} \label{proof:background}
In our system, the action policy ensures that the integrality requirements\,(Eq.\,\ref{eq:int_requirement}) are satisfied\,(see Section\,\ref{Action}). 
Therefore, integrality constraints can be omitted from this analysis.
Furthermore, since $phase1$ continues until the first feasible solution is found, we focus on the satisfaction of linear constraints\,(Eq.\,\ref{eq:constraints}) and bound constraints\,(Eq.\,\ref{eq:bound}) during $phase1$.

Since a well-trained agent selects actions that maximize the reward, the rewards $\mathcal{R}_{t,\text{const}}$ and $\mathcal{R}_{t,\text{bound}}$ are expected to be positive and zero, respectively, following the definitions in Eqs.\,\ref{eq:reward_bound} and \ref{eq:reward_const}.
Accordingly, we suppose that $\mathcal{R}_{t,\text{const}}>0$ and $\mathcal{R}_{t,\text{bound}}=0$ during $phase1$\,(i.e., exploration starting from an infeasible solution) in order to analyze the alignment between our reward design and the objective of achieving feasibility in ILP.

\subsection{Proof}

Suppose  \( \mathbf{x}_t \notin \mathcal{F} \), \( \mathcal{R}_{t,\text{const}} > 0 \), and $\mathcal{R}_{t,\text{bound}} = 0$. By the definition of the constraint reward, we have:
\[
\mathcal{R}_{t,\text{const}} = \sum_{j=1}^m \min(f_{t+1,j}, 0) - \min(f_{t,j}, 0) > 0
\]

In the context of ILP, the space of variables $x_i$ is discrete.
Since $\mathbf{A}$ and $\mathbf{b}$ consist of constant coefficients, the value of $lhs_{t,j}(=\sum_{i=1}^n a_{j,i}x_{t,i})$ is restricted to a discrete set.
Consequently, $f_{t,j}(=b_j-lhs_{t,j})$ also changes in discrete increments.
This discreteness implies that $\mathcal{R}_{t,\text{const}}$ must be at least a fixed minimum value $\delta$ whenever $\mathbf{x}_{t} \notin \mathcal{F}$.
Thus, we have:
\[
\mathcal{R}_{t,\text{const}}=\sum_{j=1}^m \min(f_{t+1,j}, 0)-\min(f_{t,j}, 0) \ge \delta>0.
\]

Let $ D_t=\sum_{j=1}^m \min(f_{t,j}, 0)$ at step $t$. From this, we recursively obtain the general form:
\[
D_1\ge D_0+\delta
\]
\[
D_2\ge D_1+\delta\ge D_0+2\delta
\]
\[
\vdots
\]
\[
D_t\ge D_0+t \cdot \delta
\]

As the initial solution is infeasible, it follows that $ D_0<0$.
Given that $\delta>0$ by the definition above, there must exist a finite timestep $t$ such that $ D_0+t\cdot\delta\ge0$.
Solving this inequality for $t$ yields $t\ge- D_0/\delta$.
Let $\mathcal{T} = \lceil- D_0/\delta\rceil$ be the smallest timestep satisfying this condition.
Given that $D_t \le 0$ inherently due to the $\min(\cdot)$ function, $D_{\mathcal{T}} \ge 0$ forces $ D_{\mathcal{T}} = 0$. 
Thus, $D_{\mathcal{T}}=\sum_{j=1}^m \min(f_{\mathcal{T},j}, 0) = 0$.

Given the nature of the $\min(\cdot)$ function, 
$f_{\mathcal{T},j}$ must be greater than or equal to $0$ for all $j$, which implies that the linear constraints\,(Eq.\,\ref{eq:constraints}) are satisfied as follows:
\[
f_{\mathcal{T},j} = b_j-lhs_{\mathcal{T},j} \geq 0, \forall j
\]
\[
lhs_{\mathcal{T},j} \leq b_j, \forall j
\]
\[
\mathbf{A}\mathbf{x}_\mathcal{T} \leq \mathbf{b}
\]
Furthermore, regarding $\mathcal{R}_{t,\text{bound}}$, a value of zero indicates that no decision variables violate their bounds\,(see Eq.\,\ref{eq:reward_bound}).
Combined with the integrality guarantee outlined in Appendix\,\ref{proof:background}, all constraints are fully satisfied. Hence, $\mathbf{x}_\mathcal{T} \in \mathcal{F}$\,(i.e., $\mathbf{x}_\mathcal{T}$ is feasible).

\vspace{1cm}

\section{Pseudo-Code for Feasibility-Aware Variable Selection}\label{appendix:var_selection}
% \vspace{-0.2cm}
\begin{algorithm}[h]
\caption{Feasibility-aware variable selection}
\label{alg:var_selection}
\textbf{Input:} instance $M$, observation $\mathcal{S}_t = (\mathbf{x}_{t}, \mathbf{f}_{t}, obj_{t})$,  current phase \texttt{phase} \\
\textbf{Parameter:} number of seed variables $p$, number of neighbor variables $q$ \\
\textbf{Output:} observation with selected variables $\tilde{\mathcal{S}}_{t} = (\tilde{\mathbf{x}}_{t}, \mathbf{f}_{t}, obj_{t})$
\begin{spacing}{1.15}
\begin{algorithmic}[1]
    \STATE $\tilde{\textbf{A}} \gets \mathbb{I}(\textbf{A}_{j,i} \neq 0)_{j=1,\dots,m;\, i=1,\dots,n}$
    \IF{$\texttt{phase = 1}$}
        \STATE $\tilde{\textbf{f}}_t \gets \mathbb{I}(f_{t,j} < 0)_{j=1,\dots,m}$
        \STATE $\texttt{score\_seed} \gets \tilde{\textbf{f}}^\top_t\tilde{\textbf{A}}$ \hfill \COMMENT{$\tilde{\textbf{f}}_t \in \mathbb{R}^{m \times 1}, \ \tilde{\textbf{A}} \in \mathbb{R}^{m \times n}$}
        \STATE $\texttt{weight} \gets (\operatorname{max}(\operatorname{abs}(\textbf{c}^\top)) -\operatorname{abs}(\textbf{c}^\top)+1) / \operatorname{max}(\operatorname{abs}(\textbf{c}^\top))$
    \ELSIF{\texttt{phase = 2}}
        \STATE $\tilde{\textbf{f}}_t \gets \mathbb{I}(f_{t,j} > 0)_{j=1,\dots,m}$
        \STATE $\texttt{score\_seed} \gets \tilde{\textbf{f}}^\top_t\tilde{\textbf{A}}$ \hfill \COMMENT{$\tilde{\textbf{f}}_t \in \mathbb{R}^{m \times 1}, \ \tilde{\textbf{A}} \in \mathbb{R}^{m \times n}$}
        \STATE $\texttt{score\_seed} \gets \operatorname{max}(\texttt{score\_seed}) - \texttt{score\_seed} + 1 $
        \STATE $\texttt{weight} \gets \operatorname{abs}(\textbf{c}^\top) / \operatorname{max}(\operatorname{abs}(\textbf{c}^\top))$
    \ENDIF
    \STATE $\texttt{score\_seed} \gets \texttt{score\_seed} \odot \texttt{weight}$ \hfill \COMMENT{$\texttt{score\_seed} \in \mathbb{R}^{1 \times n}$}
    \STATE $\texttt{prob} \gets \texttt{score\_seed} / \operatorname{sum}(\texttt{score\_seed}) \hfill$ \COMMENT{$\texttt{prob} \in \mathbb{R}^{1 \times n}$}
    \STATE $\texttt{indices\_seed} \gets \operatorname{sample}(\texttt{prob}, p) \hfill$ \COMMENT{Sample $p$ seed variables according to $\texttt{prob}$}
    \STATE $\textbf{g} \gets \operatorname{rowwise\_sum}(\tilde{\textbf{A}}[:, \, \texttt{indices\_seed}]) \hfill$ \COMMENT{$\textbf{g} \in \mathbb{R}^{m \times 1}$}
    \STATE $\texttt{score\_neighbor} \gets \textbf{g}^\top\tilde{\textbf{A}}$ \hfill \COMMENT{$\texttt{score\_neighbor} \in \mathbb{R}^{1 \times n}$}
    \STATE $\texttt{score\_neighbor}[:, \, \texttt{indices\_seed}] \gets -1$ \hfill \COMMENT{Prevent to select seed variables}
    \STATE $\texttt{indices\_neighbor} \gets \operatorname{top}(\texttt{score\_neighbor}, q) \hfill$ \COMMENT{Obtain top $q$ neighbor variables}
    \STATE $\texttt{changeable} \gets \operatorname{concatenate}(\texttt{indices\_seed}, \texttt{indices\_neighbor}) \hfill$
    \STATE $\tilde{\mathbf{x}}_{t} \gets \mathbf{x}_{t}[\texttt{changeable}] \hfill$ \COMMENT{Select $\tilde{n}(=p+q)$ variables}
    \STATE $\tilde{\mathcal{S}}_{t} \gets (\tilde{\mathbf{x}}_{t}, \mathbf{f}_{t}, obj_{t})$
\STATE \textbf{return} $\tilde{\mathcal{S}}_{t}$
\end{algorithmic}
\end{spacing}
\end{algorithm}

\newpage
\vspace{-0.1cm}
\section{Pseudo-Code for Learning Algorithm}\label{appendix:training_algo}
\vspace{-0.1cm}
% \vspace{-0.1cm}
We obtain the initial solution\,(Line~2) through one of two methods: LP-relaxation or random assignment.
For LP-relaxation, we apply random rounding to the LP-feasible solution to convert the fractional variable values into integers, which may result in an infeasible solution for the original ILP problem.
For random assignment, when training on the first instance, we randomly assign the value 1 to 1\% of the variables.
For subsequent instances, we assign the value 1 to $r$ randomly selected variables, where $r$ is half the number of variables with non-zero values in the previous instance.

\vspace{-0.2cm}
\begin{algorithm}[h]
\caption{Learning a policy for RL-SPH}
\label{alg:training_algo}
\textbf{Input:} agent parameters $\theta$, instance $M$ \\
\textbf{Parameter:} update limit $N$, 
total step limit $T_{max}$, 
step limit for $phase1$ $T_{stay}$ \\
\textbf{Output:} updated parameters $\theta$
\begin{algorithmic}[1]
\FOR{$N$ updates}
    \STATE \( \mathbf{x}_0 \gets \operatorname{get\_initial\_solution}(M) \)
    \STATE \( \mathcal{S}_0 \gets \operatorname{observe}(M, \mathbf{x}_0) \)
    \STATE \( \mathbf{x}_b \gets \varnothing \)
    \STATE $obj_b \gets \infty$
    \STATE $\texttt{phase}_0$ $\gets 1$
    \STATE \texttt{stay} $\gets$ \texttt{True}
    
    \FOR{$t=0,1,2,\dots,T_{\max}$}
        % \STATE $obj_{b}' \gets \textbf{c}^\top \textbf{x}_{b}$
        \STATE $\mathcal{R}_{t,\text{total}}, \mathcal{S}_{t+1}, \textbf{x}_b, obj_b \gets \operatorname{search}(M, \pi_\theta, \mathcal{S}_{t}, \textbf{x}_b, obj_{b}, \texttt{phase}_t) $    \hfill
        \COMMENT{See Algorithm \ref{alg:solution_search}}     
        \IF{$\texttt{stay} = \texttt{True}$ \AND $(\mathbf{x}_{t+1} \in \mathcal{F}$ \OR $t = T_{stay})$}
            \IF{$t = T_{stay}$}
                \STATE $\texttt{stay} \gets \texttt{False}$
            \ENDIF
            \STATE $\mathcal{S}_{t+1} \gets \mathcal{S}_0$
            \STATE \( \mathbf{x}_b \gets \varnothing \)
            \STATE $obj_b \gets \infty$
            % \STATE $\textbf{x}_{t+1} \gets \textbf{x}_I$
        \ELSIF{$\texttt{stay} = \texttt{False}$ \AND \( \mathbf{x}_{t+1} \) $\in \mathcal{F}$}
            \STATE $\texttt{phase}_{t+1} \gets 2$ 
        \ENDIF
        \STATE $\delta_{td} \gets \mathcal{R}_{t,\text{total}} + \gamma \cdot  V_{\theta}(\mathcal{S}_{t+1},\texttt{phase}_{t+1}) - V_{\theta}(\mathcal{S}_{t},\texttt{phase}_t)$
        \STATE $\mathcal{L}_\theta \gets  -\log \pi_{\theta}(\mathcal{A}_t \mid \mathcal{S}_t,\texttt{phase}_t) \cdot \delta_{td} + \delta_{td}^2$
        \STATE $\theta \gets \operatorname{update}(\mathcal{L}_\theta, \theta)$
        
    \ENDFOR 
\ENDFOR
    \STATE \textbf{return} $\theta$
    % \RETURN $\textbf{x}_b$
\end{algorithmic}
\end{algorithm}

\vspace{-0.3cm}
\section{Additional Preliminaries} 
\vspace{-0.13cm}
\subsection{Properties of ILP} \label{Properties of ILP}
\vspace{-0.13cm}

All ILP problems can be transformed into the standard form\,(Eqs.\,\ref{eq:obj_func}--\ref{eq:bound})\,\cite{bertsimas1997introduction}.
Let $\mathbf{a}_i^\top$ denote a row vector of a constraint, 
$\mathbf{A} = \left(\mathbf{a}_1^\top, \ldots, \mathbf{a}_m^\top\right)$, and $\mathbf{b} = \left(b_1, \ldots, b_m\right)$.
An equality constraint $\mathbf{a}_i^\top\mathbf{x} = b_i$ is equivalent to two inequality constraints: $\mathbf{a}_i^\top\mathbf{x} \geq b_i$ and $\mathbf{a}_i^\top\mathbf{x} \leq b_i$.
Moreover, $\mathbf{a}_i^\top\mathbf{x} \geq b_i$ is equivalent to $-\mathbf{a}_i^\top\mathbf{x} \leq -b_i$.
Maximizing $\mathbf{c}^\top \mathbf{x}$ is equivalent to minimizing $-\mathbf{c}^\top \mathbf{x}$.
Thus, we only address the standard form\,(i.e., minimization) in this study.

ILP is known to be NP-hard due to its integrality constraints\,(Eq.\,\ref{eq:int_requirement})\,\cite{berthold2006primal}.
As the number of integer variables increases, the computational cost grows exponentially\,\cite{floudas1995nonlinear}.
LP-relaxation transforms an original ILP problem into an LP one by removing the integrality constraints\,\cite{bertsimas1997introduction}.
Although LPs are computationally cheaper to solve, which can be solved in polynomial time\,\cite{karmarkar1984new, vanderbei1998linear}, an LP-feasible solution may be infeasible for the original ILP due to the integrality constraints\,(Eq.\,\ref{eq:int_requirement})\,\cite{guieu1999analyzing}.

\vspace{-0.13cm}
\subsection{Start Primal Heuristics for ILP} 
\vspace{-0.13cm}
\label{Start primal heuristics for ILP}

Start primal heuristics\,(SPH) do not require an initial ILP-feasible solution.
Instead, they typically begin with an LP-feasible solution and attempt to convert it into an ILP-feasible one\,\cite{berthold2006primal}, shown in Figure~\ref{fig:sph}.
Representative methods include diving heuristics, feasibility pump\,(FP), rounding heuristics, and relaxation enforced neighborhood search\,(RENS)\,\cite{berthold2006primal, shoja2023exact}.
Specifically, diving heuristics fix fractional variables in the LP solution to promising integer values and iteratively resolve the LP.
FP alternates between two sequences, one LP-feasible and the other ILP-feasible, with the goal of convergence to a feasible ILP solution.
Rounding heuristics attempt to obtain an ILP-feasible solution by rounding fractional LP values up or down.
RENS constructs and solves a sub-ILP of the original problem by fixing or tightening bounds of integer variables based on the LP solution.

% \vspace{-0.2cm}
\vspace{-0.1cm}
\subsection{Bipartite Graph Representation of ILP} 
\label{Graph Representation of ILP}
\vspace{-0.1cm}

Prior studies on E2EPH represent ILP instances as bipartite graphs\,\cite{nair2020solving, yoon2022confidence, PAS, canturk2024scalable, huang2024contrastive, liu2025apollo}.
% In a bipartite graph representation of ILP, the relationships between constraints and decision variables are depicted as edges.
In this representation, one set of nodes corresponds to constraints, and the other to decision variables.
An edge connects a variable node to a constraint node if and only if the variable appears in the corresponding constraint.
For example, in Figure~\ref{fig:overview}(a), variable $x_3$ appears in  constraint $\mathbf{a_2}$; thus, the node representing $x_3$ is connected to the node representing $\mathbf{a_2}$ in the bipartite graph.

\vspace{-0.1cm}
\subsection{Graph Neural Networks for ILP} \label{GT}
\vspace{-0.1cm}

Existing E2EPH methods are typically built upon the message passing neural network\,(MPNN) framework\,\cite{MPNN}.
MPNNs aggregate messages from the neighbor nodes, making them well-suited for capturing local structural information.
However, they struggle to capture long-range dependencies between distant nodes\,\cite{zhang2020graph-bert, wu2021representing}.
To propagate messages between nodes that are $K$ hops apart, an MPNN requires at least $K$ layers.

In the context of ILP, capturing relationships among variables that influence each other across multiple constraints may require deep MPNNs.
However, deeper architecture often suffers from the oversmoothing problem\,\cite{li2018deeper, wu2021representing, survey_Graph_transformer}.
For instance, as shown in Figure~\ref{fig:overview}(a), variables $x_2$ and $x_3$ are four hops apart: $x_2$ - $\mathbf{a_1}$ - $x_1$ - $\mathbf{a_2}$ - $x_3$.
Although $x_2$ and $x_3$ do not appear in the same constraint, they are connected via $x_1$, which is shared by both $\mathbf{a_1}$ and $\mathbf{a_2}$.
A change in $x_2$ can affect $x_1$, which in turn may influence $x_3$.
% Thus, capturing the relationship between these variables is essential.
Modeling such interactions would require four layers, but even shallow MPNNs with 2–4 layers are prone to oversmoothing\,\cite{wunon}.
Thus, we design a new Graph Neural Network\,(GNN) for ILP inspired by Graph Transformers\,\cite{rong2020grover, ying2021transformers, lin2022survey, survey_Graph_transformer}, which can effectively learn relationships between distant nodes\,\cite{wu2021representing}.

\vspace{-0.1cm}
\section{Related Work}\label{Related Work}
\vspace{-0.1cm}

\subsection{Machine Learning for ILP}\label{overview:ML4ILP}
\vspace{-0.1cm}

ML techniques for solving ILP can be broadly categorized into three groups\,\cite{bengio2021machine}.
The first group, \textit{learning to configure algorithms}, uses ML to optimize the configuration of specific components within ILP solvers. 
Examples include deciding parameters for promising configurations\,\cite{hutter2011sequential},  applying decomposition\,\cite{kruber2017learning}, and selecting scaling methods\,\cite{berthold2021learning}. 
The second group, \textit{ML alongside optimization algorithms}, integrates ML into ILP solvers to aid key decisions during the optimization, such as cut selection\,\cite{tang2020reinforcement, paulus2022learning}, variable selection\,\cite{khalil2016learning, alvarez2017machine, gasse2019exact, gupta2020hybrid, paulus2023learning}, node selection in the branch-and-bound\,\cite{he2014learning, labassi2022learning}, and neighborhood selection in LNS\,\cite{song2020general, sonnerat2021learning, wu2021learning, huang2023searching, liu2024mixed}.

RL-SPH belongs to the third group, \textit{end-to-end learning}, which uses ML to directly learn and predict solutions. 
This group includes existing E2EPH methods\,\cite{nair2020solving, shen2021learning, yoon2022confidence, PAS, canturk2024scalable, huang2024contrastive, kdd_complete, liu2025apollo, iclr_complete} as well.
PAS\,\cite{PAS} adopts a predefined trust region instead of strictly fixing variables, which can be viewed as the generalization of the fixing strategy\,\cite{huang2024contrastive}.
Recent methods\,\cite{huang2024contrastive, liu2025apollo} follow this approach as well.
Although using trust regions alleviates the risk of infeasibility, these methods still rely on ILP solvers to obtain feasible solutions.
Several efforts have been made to solve ILPs via ML without relying on ILP solvers \cite{kdd_complete, iclr_complete}, but none have attained a 100\% feasibility rate.
Specifically, \citeauthor{kdd_complete} reported that even when their proposed method is combined with SCIP’s heuristic, it still fails to achieve 100\% feasibility on the CA benchmark.
Similarly, \citeauthor{iclr_complete} documented the average feasible ratios of 50.8\%, 97.1\%, and 99.4\% on the SC, IS, and CA benchmarks, respectively; furthermore, they observed that PAS\,\cite{PAS} and ConPaS\,\cite{huang2024contrastive} fail to find feasible solutions for the majority of instances.
Moreover, the aforementioned methods primarily focus on binary variables.
To the best of our knowledge, RL-SPH is the first standalone E2EPH method that learns to generate feasible solutions independently, establishing a theoretical reward-feasibility alignment and empirically achieving a 100\% feasibility rate across the five evaluated benchmarks, even for ILP involving non-binary integers.

% \vspace{-0.1cm}
\subsection{Positioning Relative to Prior Work}\label{positioning}
% \vspace{-0.1cm}
RL-SPH is a \textit{start primal heuristic} designed to quickly achieve feasibility from ILP-infeasible solutions.
As discussed in Appendix\,\ref{overview:ML4ILP}, RL-SPH can also be categorized under \textit{end-to-end learning}, but it fundamentally differs from existing approaches in terms of how feasible solutions are obtained. 
For example, as illustrated in Figure\,\ref{fig:method_comparison}, typical E2EPH methods rapidly approximate a solution and then delegate the remaining optimization—including ensuring feasibility—to an external solver. 
In this workflow, the solver bears responsibility for producing feasible solutions, typically via its built-in start primal heuristics.
Consequently, most prior E2EPH work has evolved under the assumption that feasibility will be handled by the external solver rather than by the learned model itself.

This distinction underscores that RL-SPH addresses a more challenging problem than existing E2EPH work, as it must generate feasible solutions \textit{without} relying on external solvers. 
Many real-world applications—such as real-time risk assessment\,\cite{chen2023end}—require high-quality feasible solutions quickly and repeatedly\,\cite{kdd_complete,canturk2024scalable}. 
In such settings, RL-SPH is advantageous because it can quickly deliver feasible solutions independently of a solver.
Notably, the LP-free variant of RL-SPH discovers the first feasible solution 180$\times$ faster than established SPH baselines on average\,(see Section\,\ref{Results}).

Feasibility is a necessary prerequisite for any subsequent optimization stage. 
For example, widely used ILP metaheuristics such as local branching\,\cite{fischetti2003local, liu2022learning} and large neighborhood search (LNS)\,\cite{shaw1998using, song2020general, wu2021learning, huang2023searching, liu2024mixed} require an initial feasible solution to operate and further improve the incumbent solution. 
As a result, infeasibility constitutes a significant bottleneck in solving ILP problems.
However, finding a feasible solution remains inherently challenging due to the integrality constraints\,(Eq.\,\ref{eq:int_requirement}), which induce an exponential increase in computational cost with respect to the number of integer variables\,\cite{floudas1995nonlinear}.
Moreover, even well-established primal heuristics typically cannot guarantee feasibility\,\cite{shoja2023exact}.
Thus, the ability to reliably generate feasible solutions for general ILPs is crucial.
Despite the importance of feasibility, learning-based approaches for directly addressing this problem remain underexplored.
Advancing learning-based methods for achieving feasibility establishes a crucial foundation for ML-based standalone solvers pursuing optimality, especially given that achieving optimality is recognized as a major challenge in the ML for CO community\,\cite{meta-sage}.

Existing E2EPH methods and RL-SPH possess \textit{orthogonal} advantages, making them ideal candidates for a hybrid approach. 
Specifically, RL-SPH can leverage existing E2EPH as a solution initializer, a synergy demonstrated in Section~\ref{E2EPH-exp}.
RL-SPH resembles LNS in that it explores neighborhoods larger than those used in traditional local search\,(see Section\,\ref{Neighborhood search}).
However, as previously noted, existing learning-based LNS methods require an initially feasible solution and aim to improve the incumbent solution, which positions them in a different stage of the solution process from start primal heuristics. 
Therefore, they are not direct points of comparison for our study.
Nonetheless, insights from the LNS literature could be leveraged to further enhance the behavior of RL-SPH \textit{after} it obtains a feasible solution.

% \vspace{-0.1cm}
\subsection{Evolution of the Neural Local Rewriting Paradigm}\label{NeuRewriter}
% \vspace{-0.1cm}

NeuRewriter \cite{chen2019rewrite} is a pioneering method that employs an actor-critic algorithm for rewriting solutions in CO tasks such as job scheduling and vehicle routing. While RL-SPH shares the conceptual similarities of identifying and ``rewriting'' sub-regions of a solution, it represents a significant progression in two key dimensions:

\begin{enumerate}[nosep, itemsep=0.6ex]
    \item \textbf{From infeasible to feasible}: NeuRewriter is suited for problems where an initial feasible solution is readily available, focusing on improving the incumbent solution (aligning with our $phase2$). \citet{chen2019rewrite} envisioned extending this approach to a broader range of problems that are hard for classic algorithms when starting with an infeasible solution. RL-SPH directly realizes this vision by establishing a concrete framework to navigate from infeasible to feasible solutions ($phase1$), addressing the previously identified challenge.
    \item \textbf{General ILP formulation}: NeuRewriter relies on task-specific architectures\,(e.g., dedicated encoders). In contrast, RL-SPH operates on an ILP formulation, serving as a general-purpose framework. We validated this with a consistent architecture and hyperparameter configuration across five benchmarks. While we do not claim RL-SPH is optimal for every specific problem, we emphasize its versatility—it can be applied to various CO problems expressible as an ILP without requiring significant architectural adjustments.
\end{enumerate}
This comparison contextualizes the paradigm's evolution, highlighting RL-SPH's contribution in bridging the gap between infeasibility and feasibility within a general-purpose framework.

% \newpage
% \vspace{-0.1cm}
\section{Additional Explanation of Reward System} \label{appendix:vis_expl}
% \vspace{-0.1cm}
\subsection{Rationale on Reward Design} \label{appendix:reward-rationale}
% \vspace{-0.1cm}
As defined in Section\,\ref{Mixed Integer Linear Programming}, ILP is an optimization problem that minimizes an objective value while satisfying a set of constraints: linear constraints, integrality constraints, and variable bounds.
When framing ILP within an RL system, the reward mechanism must likewise be designed to reflect this definition.
Accordingly, we derived the following fundamental types of rewards: those related to objective values, linear constraints, and variable bounds.
The reward for integrality constraints is omitted because they are handled by the action policy (see Section\,\ref{Action}).

According to the above definition, satisfying the set of constraints is a prerequisite for performing optimization.
Therefore, rewards related to the set of constraints should take high priority over the objective-value reward.
We believe it is intuitively reasonable that the more a bound or linear constraint is violated, the greater the penalty should be. 
Likewise, a larger reward is appropriate as the objective value improves. 
% As discussed in Section, satisfying the variable bounds should take precedence over satisfying the linear constraints. 
From an optimization perspective, improving the feasibility of linear constraints while violating variable bounds is meaningless.
For example, in an ILP where variables must take values of either $0$ or $1$, assigning a value such as $-1$ is invalid.
Thus, satisfying the variable bounds has the highest priority among the variable bounds, linear constraints, and objective values. 
The reward structure described in Eq.\,\ref{eq:reward_phase1_obj} obeys the mentioned priority. 
Additionally, to avoid the bound penalty being overly dominated by the constraint reward, we scaled it by the square root of the number of changeable variables. 
The maximum value of the objective coefficients scales the reward regarding the objective value. 
This scaling helps stabilize learning by mitigating large reward fluctuations as well.

% \vspace{-0.1cm}
\subsection{Feasibility Reward} \label{appendix:Feasibility_reward}
\begin{figure}[h]
  \centering
  \includegraphics[width=0.3\columnwidth]{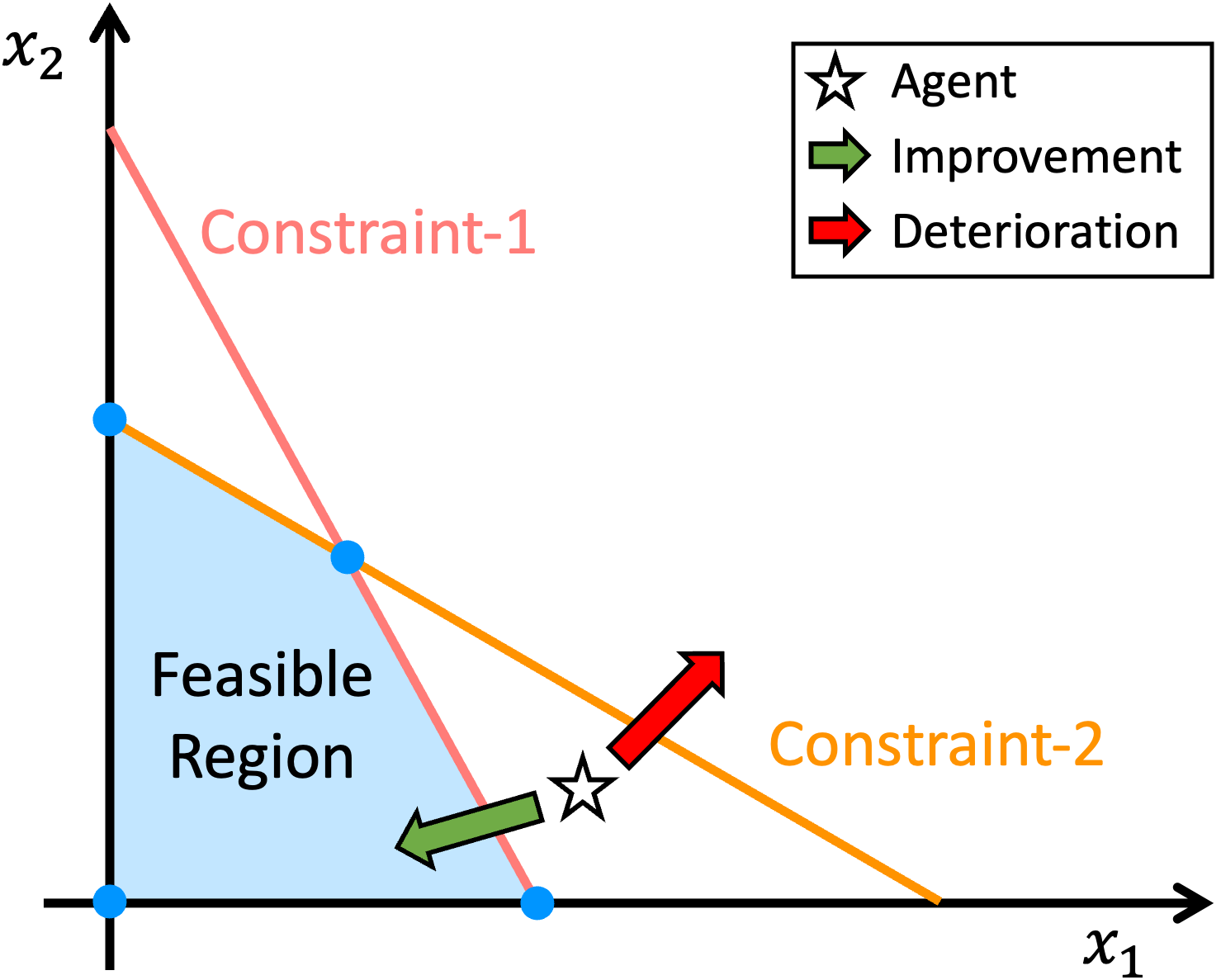}
  \caption{Illustration of feasibility improvement and deterioration.}
  \label{fig:feasibility}
\end{figure}

% \vspace{-0.1cm}
The feasibility reward $\mathcal{R}_{t,\text{F}}$ is calculated in proportion to the degree of feasibility improvement or deterioration.
In Figure\,\ref{fig:feasibility}, the agent violates Constraint-1 while satisfying Constraint-2. 
If the agent moves closer to the feasible region, it is considered an improvement in feasibility. 
Conversely, if it moves further away and ends up violating Constraint-2 as well, it is regarded as a deterioration.
In this way, rewards are assigned based on how much the agent’s actions improve or worsen feasibility. 
As a result, the agent learns to find feasible solutions to maximize its rewards.

\vspace{-0.1cm}
\subsection{Reward in $phase2$} \label{appendix:phase2}
% \vspace{-0.1cm}

\begin{figure}[h]
  \centering
  \includegraphics[width=0.4\columnwidth]{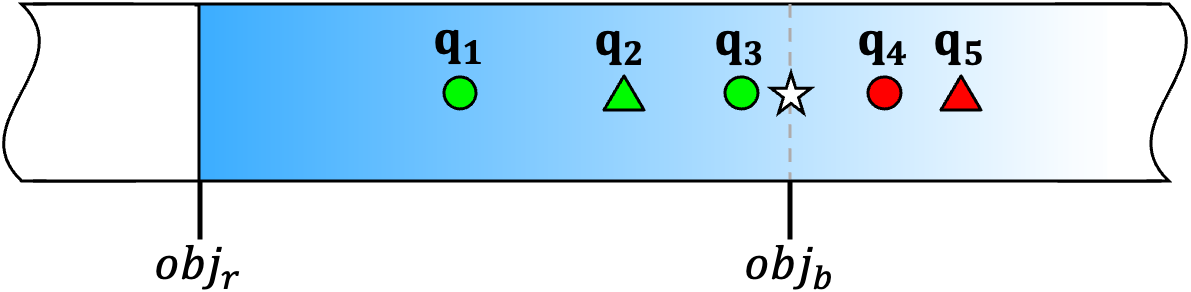}
  \caption{Illustration of reward function in $phase2$. \scalebox{0.8}{$\bigcirc$}: feasible, $\triangle$: infeasible, green: inside, red: outside, \ding{73}: agent, $obj_r$: the objective value of the LP-feasible solution, $obj_b$: incumbent value.}
  \label{fig:reward}
\end{figure}

The green circles in Figure\,\ref{fig:reward} represent feasible solutions whose objective values are $obj_{t+1}<obj_b$\,(Case 1 in Eq.\,\ref{eq:reward_phase2_obj}). 
We regard the agent’s actions that fail to find better feasible solutions than $\textbf{x}_b$ as incorrect\,(Cases 2, 3, 4).
The red circle indicates a feasible solution worse than the incumbent\,(Case 2), while all triangles correspond to infeasible solutions\,(Cases 3, 4).

Let the objective values be $obj_b=-10$, $\mathbf{q_1}=-16$, $\mathbf{q_2}=-13$, $\mathbf{q_3}=-11$, $\mathbf{q_4}=-8$, and $\mathbf{q_5}=-7$, with $\alpha=2$ and $\max(|\textbf{c}|)=1$.
For the green circles, rewards in $phase2$ are calculated by the first case in Eq.\,\ref{eq:reward_phase2_obj}. 
The rewards for $\mathbf{q_1}$ and $\mathbf{q_3}$ are $\Delta obj_t =\frac{|-16 - (-10)|}{1} = 6$ and  $\Delta obj_t =\frac{|-11 - (-10)|}{1} = 1$, respectively.
Since $\mathbf{q_1}$ has the better objective value than $\mathbf{q_3}$, it receives a higher reward.
For the red circle, the reward is $-\Delta obj_t \cdot \alpha = - \frac{|-8 - (-10)|}{1}\cdot2 = -4$. 
For the triangles, the penalties for $\mathbf{q_2}$ and $\mathbf{q_5}$ are $\mathcal{R}_{t,\text{F}}$ and $\mathcal{R}_{t,\text{F}} \cdot \alpha = 2 \cdot \mathcal{R}_{t,\text{F}}$, respectively.
The distances from $obj_b$ to $\mathbf{q_2}$ and $\mathbf{q_5}$ are the same\,(i.e., $|-13 - (-10)|=|-7 - (-10)|=3$), but the penalty for $\mathbf{q_2}$ is smaller than that for $\mathbf{q_5}$ due to amplifying by $\alpha$.

% \vspace{-0.1cm}
\subsection{Toward-Optimal Bias} \label{appendix:bias}
\begin{figure}[h]
     \centering
     \hfill
     \begin{subfigure}[b]{0.47\columnwidth}
         \centering
         \includegraphics[width=0.85\columnwidth]{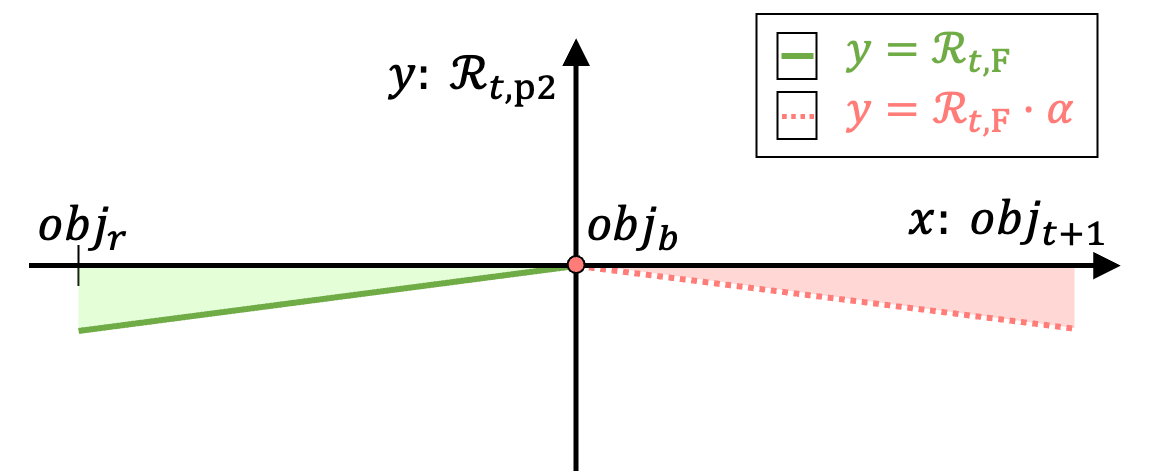}
         \caption{$\alpha=1$}
         \label{fig:alpha_1}
     \end{subfigure}
     \hfill
     \begin{subfigure}[b]{0.47\columnwidth}
         \centering
         \includegraphics[width=0.85\columnwidth]{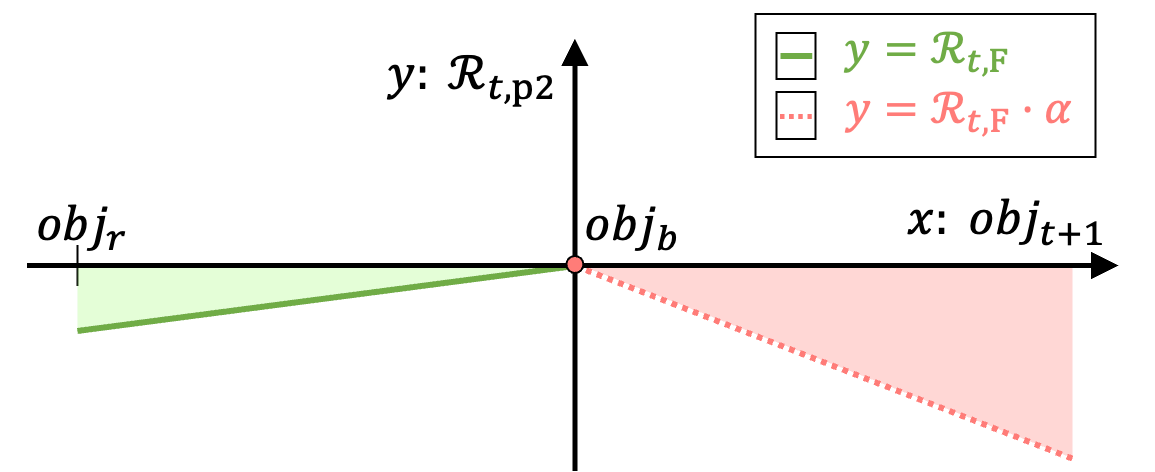}
         \caption{$\alpha=3$}
         \label{fig:alpha=3}
     \end{subfigure}
     \hfill
     \caption{Illustration of the potential penalties of $\mathcal{R}_{t,\text{p2}}$ as a function of the objective value $obj_{t+1}$ in $phase2$. The $x$-axis represents $obj_{t+1}$, and the $y$-axis represents $\mathcal{R}_{t,\text{p2}}$, with the origin set at $obj_b$.}
    \label{fig:alpha}
\end{figure}

% \vspace{-0.1cm}
Figure\,\ref{fig:alpha} visualizes the agent’s potential penalties by $\mathcal{R}_{t,\text{F}}$ in $phase2$.
The third and fourth quadrants illustrate the penalties, assuming that $\mathcal{R}_{t,\text{F}}$ is a linear function of the gap from $obj_{t+1}$ to the incumbent value $obj_b$ for simplicity.
As illustrated in Figures\,\ref{fig:alpha_1} and\,\ref{fig:alpha=3}, a higher $\alpha$ results in higher penalties for solutions with $obj_{t+1} > obj_b$.
By controlling the toward-optimal bias $\alpha$, we can guide the agent to explore promising regions\,(i.e., $obj_{t+1}<obj_b$) for better feasible solutions rather than making ineffective moves\,(i.e., $obj_{t+1}>obj_b$), thus increasing its opportunities for learning.
Appendix~\ref{appendix:toward-optimal} provides experimental results on the toward-optimal bias $\alpha$.

% \newpage
% \vspace{-0.1cm}
\section{Details of Experimental Setup} \label{appendix:experiment_setup}

% \vspace{-0.1cm}
\subsection{Benchmark Datasets} \label{appendix:datasets}
% \vspace{-0.1cm}

Table~\ref{tab:dataset_size} shows the average sizes of each benchmark dataset used in our experiments.
We generated instances for IS, CA, SC, and MVC following the code\footnote{\url{https://github.com/ds4dm/learn2branch}} from\,\cite{gasse2019exact}.
For NBI, instances were generated based on the description in\,\cite{qi2021smartpump}.
Table\,\ref{tab:data_param} summarizes the parameters used for NBI instance generation.
Considering that the ratio of non-zero coefficients $\rho$ in typical LP problems is less than 5\%\,\cite{hillier2015introduction}, we set a higher density of 10\% to promote more interactions between variables in constraints.
According to the default settings of the ILP solver Gurobi\,\cite{gurobi} and SCIP\,\cite{scip}, the lower bound $l_i$ and upper bound $u_i$ for decision variables are set to 0 and $\infty$, respectively.

% \vspace{-0.3cm}
\setlength{\tabcolsep}{10pt} 
\begin{table}[h]
\vspace{-0.2cm}
\caption{Average sizes of each dataset.}
\vspace{-0.2cm}
\label{tab:dataset_size}
\begin{center}
% \begin{small}
\resizebox{0.85\columnwidth}{!}{%
\begin{tabular}{lcccc}
\toprule
Dataset                    & \# binary variables & \# integer variables & \# constraints & Density \\ \midrule
Independent set\,(IS)       & 1,500               & 0                    & 5,941     & 0.13\%     \\
Combinatorial auction\,(CA) & 4,000               & 0                    & 2,677     & 0.21\%     \\
Set covering\,(SC)          & 3,000               & 0                    & 2,000     & 5\%     \\
Minimum vertex cover\,(MVC) & 3,000               & 0                    & 11,931     & 0.07\%     \\
Non-binary integers\,(NBI)  & 0                   & 2,000                & 2,000     & 10\%     \\ \bottomrule
\end{tabular}
}
% \end{small}
\end{center}
\vspace{-0.1cm}
\end{table}

NBI instances follow the structure of an Unbounded Multidimensional Knapsack Problem, with non-zero coefficients in $\mathbf{A}$ sampled from $\{1, \dots, 10\}$, positive RHS values, and $\mathbf{c}$ from $\{-10, \dots, -1\}$.
These instances are formulated in the standard form as in Eq.\,\ref{eq:standard_from_ilp}: $\text{min. }\mathbf{c}^\top\mathbf{x} \quad \text{s.t. }\mathbf{A}\mathbf{x} \leq\mathbf{b}, \quad x_i \in \mathbb{Z}_{\ge 0},\forall i$.
Under this formulation, $\mathbf{A}$ and $\mathbf{b}$ can be interpreted as representing resource consumption and capacities, respectively.
Since all objective coefficients are negative, the minimization objective is equivalent to maximizing total profit (i.e., maximizing $-\mathbf{c}^\top\mathbf{x}$).

\vspace{-0.1cm}
\begin{table}[h]
% \vskip 0.1in
\caption{Parameters for non-binary integer instance generation.}
\vspace{-0.2cm}
\label{tab:data_param}
\begin{center}
\begin{small}
% \begin{sc}
\begin{tabular}{l|l}
    \toprule
    Parameter & Distribution \\
    \midrule
    $\textbf{c}$ & $\text{randint}[-10,-1]$ \\
    $\textbf{A}$ & $\text{randint}[1,10]$ with density $\rho = 0.1$ \\ 
    $\textbf{b}$ & $\textbf{A} \xi + \epsilon$, where \\
                 & $\xi_i \sim \text{randint}[1,10], \forall i=1, \ldots, n$ and \\
                 & $\epsilon_j \sim \text{randint}[1,10], \forall j=1, \ldots, m$ \\
    $l_i$ & $0, \forall i=1, \ldots, n$ \\
    $u_i$ & $\infty, \forall i=1, \ldots, n$ \\
    \bottomrule
\end{tabular}
% \end{sc}
\end{small}
\end{center}
\end{table}

% \vspace{-0.1cm}
\subsection{Implementation Details} \label{appendix:Method_cofig}
% \vspace{-0.1cm}

% They were trained concurrently on 64 different instances with 5,000 parameter updates for the results in Tables\,\ref{tab:main_result} and\,\ref{tab:ablation}, and 50,000 for Table\,\ref{tab:neighbor}.
The following hyperparameter configuration remained fixed across all experiments, except for the sensitivity analyses in Appendix\,\ref{appendix:hyperparameter-analysis}.
RL-SPH was trained concurrently on 64 distinct instances, which serves as the batch size of ILP-GT.
We maintained a consistent configuration for 
Eq.~\ref{eq:reward_phase2_obj},
Algorithm~\ref{alg:var_selection}, and Algorithm~\ref{alg:training_algo}: 
the toward-optimal bias $\alpha = 2$,
the number of seed/neighbor variables $p=q=\left\lceil\log_2n\right\rceil$, 
the update limit $N=5000$, 
the total step limit $T_{max}=2000$, 
and the step limit for $phase1$ $T_{stay}=500$.
Training was conducted using Algorithm\,\ref{alg:training_algo} with the RMSprop optimizer\,(learning rate $= 1 \times 10^{-4}$, epsilon $= 1 \times 10^{-5}$, alpha $= 0.99$, weight decay $= 1 \times 10^{-3}$), where the learning rate was linearly decayed.

We implemented our ILP-GT based on the Graph Transformer code from GitHub\footnote{\url{https://github.com/ucbrise/graphtrans}}\,\cite{wu2021representing}, with the same configuration.
We utilized the official implementation\footnote{\url{https://github.com/yandex-research/rtdl-num-embeddings}} of the positional encoding module\,\cite{gorishniy2022periodicembeddings}.
Our RL algorithm is built upon the Actor-Critic implementation in PyTorch\footnote{\url{https://github.com/ikostrikov/pytorch-a2c-ppo-acktr-gail}}\,\cite{a2c_github}, modified to be tailored for ILP.

As shown in Table\,\ref{tab:train_time}, the training durations for IS, CA, SC, MVC, and NBI were 31, 32, 26, 83, and 23 minutes, respectively.
Notably, the training process is highly efficient, with all tested datasets requiring less than 90 minutes.
Appendix\,\ref{appendix:train-curves} also illustrates the stability and convergence of the training process.
Table\,\ref{tab:train_time} also details the size of ILP-GT trained on each dataset. 
We compare RL-SPH's training costs with existing E2EPH methods in Appendix\,\ref{appendix:training_times}.

\begin{table}[h]
\centering
% \vspace{-0.3cm}
% \vspace{-0.1cm}
\caption{Training time and model size of RL-SPH on each dataset described in Table\,\ref{tab:dataset_size}.}
\vspace{-0.1cm}
\label{tab:train_time}
\resizebox{0.65\columnwidth}{!}{%
\begin{tabular}{cccccc}
\specialrule{1pt}{0pt}{2pt}
Dataset       & IS & CA & SC & MVC & NBI \\ 
\specialrule{1pt}{2pt}{2pt}
Training time~(min) & 31 & 32 & 26 & 83  & 23  \\ 
\# trainable parameters & 3.9M &	2.7M &	2.4M &	6.2M &	2.4M  \\ 
\specialrule{1pt}{2pt}{0pt}
\end{tabular}%
}
\end{table}
% \vspace{-0.3cm}

\vspace{0.1cm}
\subsection{Details of SPH Baselines} \label{appendix:baselines}
% \vspace{-0.1cm}
To ensure comprehensive coverage, DHF encompasses 15 diving heuristics, and RHF covers 6 rounding variants implemented in SCIP\,(v8.1.0), as follows:

\begin{itemize}[labelindent=0em, topsep=0pt, itemsep=0pt, partopsep=0pt]
    \item \textbf{Diving Heuristic Family\,(DHF)}: \textit{adaptivediving, conflictdiving, guideddiving, objpscostdiving, distributiondiving, farkasdiving, fracdiving, linesearchdiving, pscostdiving, veclendiving, rootsoldiving, nlpdiving, coefdiving, intdiving,} and \textit{actconsdiving}.
    \item \textbf{Rounding Heuristic Family\,(RHF)}: \textit{simplerounding, rounding, zirounding, randrounding, shifting,} and \textit{intshifting}.
\end{itemize}

\vspace{0.2cm}
\subsection{Evaluation Environment} \label{appendix:Environment}
% \vspace{-0.1cm}
We conducted all evaluations under identical configurations.
The evaluation machine is equipped with two AMD EPYC 7302 @ 3.0GHz, 2048GB RAM, and four NVIDIA A100 GPUs.
All experiments were performed using a single NVIDIA A100 GPU.
The software environment includes PyTorch 1.12.0, Gymnasium 0.29.1, and SCIP 8.1.0.

\newpage

% \vspace{-0.1cm}
\section{Additional Experiments} \label{appendix:additional-exp}
% \vspace{-0.1cm}

\subsection{Further Comparison with SOTA E2EPH} \label{appendix:E2EPH-exp}

We trained PAS, DDIM, and DiffILO using the default configurations and the authors’ released code\footnote{\url{https://github.com/sribdcn/Predict-and-Search_MILP_method}}$^{,}$ \footnote{\url{https://github.com/MIRALab-USTC/L2O-DiffILO}}$^{,}$ \footnote{\url{https://github.com/agent-lab/diffusion-integer-programming}}.
The best-performing PAS and DiffILO models were selected within 1,200 training epochs, and this epoch setting is among those used by \citet{iclr_complete}.
As summarized in Table\,\ref{tab:comparison-train-time}, PAS, DDIM, and DiffILO required approximately 8.7$\times$, 1.4$\times$, and 33.9$\times$ more training time than RL-SPH, respectively.
This indicates that the baselines were allocated sufficient training time in our experiment.

\setlength{\tabcolsep}{15pt} 
\begin{table}[h]
\caption{Comparison of training duration\,(in minutes) and RL-SPH's relative speedup against baseline methods.}
\label{tab:comparison-train-time}
\begin{tabular}{cccccccc}
\specialrule{1pt}{0pt}{2pt}
\multirow{2}{*}{Dataset} & \multicolumn{4}{c}{Training duration (min)} & \multicolumn{3}{c}{RL-SPH's speedup} 
\\ 
\cmidrule[0.5pt](l{0.3em}r{0.3em}){2-5}
\cmidrule[0.5pt](l{0.3em}r{0.3em}){6-8}
        & RL-SPH & PAS   & DDIM & DiffILO & PAS  & DDIM & DiffILO \\ 
\specialrule{1pt}{2pt}{2pt}   
SC      & 26     & --     & 66   & 850     & --    & 2.5$\times$  & 32.7$\times$    \\ 
CA      & 32     & 343   & 43   & 1177    & 10.7$\times$ & 1.3$\times$  & 36.8$\times$    \\
IS      & 31     & 209   & 14   & 1002    & 6.7$\times$  & 0.5$\times$  & 32.3$\times$    \\
\cmidrule[0.5pt](l{0.3em}r{0.3em}){1-8}
Average & 29.7   & 276.0 & 41.0 & 1009.9  & \textbf{8.7$\times$}  & \textbf{1.4$\times$}  & \textbf{33.9$\times$}   \\
\specialrule{1pt}{2pt}{0pt}
\end{tabular}
\end{table}

DiffILO achieved a 100\% FR only on the IS dataset\,(see Table\,\ref{tab:diffilo-exp}), whose characteristics closely match those used in the DiffILO study, as shown in Table~\ref{tab:diff-dataset}. 
In contrast, SC differs in the number of variables and constraints, and our CA dataset contains approximately 2.7$\times$ more variables and 4.7$\times$ more constraints. 
This indicates that, even within the same problem class, variations in dataset characteristics may necessitate additional tuning for DiffILO to maintain high performance. 
This claim is further supported by Table 4 in the DiffILO paper, which shows that DiffILO performed dataset-specific tuning on 11 important hyperparameters.
In contrast, RL-SPH did not employ any dataset-specific tuning; a consistent hyperparameter configuration was used for all datasets, as detailed in Appendix\,\ref{appendix:Method_cofig}. 
Therefore, the fact that RL-SPH achieved a 100\% FR in all experiments without dataset-specific tuning demonstrates its robust capability to generate feasible solutions.

\begin{table}[h]
\setlength{\tabcolsep}{15pt} 
\centering
\caption{Characteristics of datasets used in the DiffILO study and ours. }
\vspace{-0.2cm}
\label{tab:diff-dataset}
\resizebox{0.68\columnwidth}{!}{%
\begin{tabular}{cccc}
\specialrule{1pt}{0pt}{2pt}
Dataset             & Study   & Number of variables & Number of constraints  \\ 
\specialrule{1pt}{2pt}{2pt}
\multirow{2}{*}{SC} & DiffILO & 2,000        & 3,000     \\
                    & Ours    & 3,000        & 2,000     \\ 
\cmidrule[0.5pt](l{0.3em}r{0.3em}){1-4}
\multirow{2}{*}{CA} & DiffILO & 1,500        & 576       \\
                    & Ours    & 4,000        & 2,677     \\ 
\cmidrule[0.5pt](l{0.3em}r{0.3em}){1-4}
\multirow{2}{*}{IS} & DiffILO & 1,500        & 5,943     \\
                    & Ours    & 1,500        & 5,941     \\ 
\specialrule{1pt}{2pt}{0pt}
\end{tabular}
}
\end{table}

% \vspace{-0.2cm}
\vspace{0.1cm}
\subsection{Evaluation of Generalization} \label{appendix:generalization-exp}

% \vspace{-0.1cm}
\subsubsection{Generalization to Unseen MIPLIB Benchmarks} \label{appendix:miplib}
% \vspace{-0.1cm}

We further evaluated the joint RL-SPH model on diverse instances from the MIPLIB benchmark\,\cite{MIPLIB} to assess its generalization capabilities. 
Table\,\ref{tab:joint-exp} presents summary statistics of the instances used in this experiment, along with incumbent values obtained by RHF and RL-SPH.
Notably, the joint model obtains feasible solutions for these MIPLIB instances despite them differing substantially from the training instances\,(IS and MVC) in the following aspects:

\begin{enumerate}[nosep, itemsep=0.5ex]
    \item \textbf{Variable domains}: \texttt{rail507} and \texttt{gen-ip002} contain continuous or integer variables with infinite upper bounds.
    \item \textbf{Scale}: \texttt{scpk4} contains nearly 67 times more variables than the training instances.
    \item \textbf{Density}: Matrix density spans from highly sparse (0.1\% in \texttt{toll-like}) to fully dense (100\% in \texttt{gen-ip016}).
\end{enumerate}
These results demonstrate that RL-SPH generalizes its problem-solving capabilities to unseen instances, even when encountering different variable types and domains.

\newpage

\setlength{\tabcolsep}{6pt} 
\renewcommand{\arraystretch}{1}
\begin{table}[h]
\centering
\vspace{-0.1cm}
\caption{Statistics of instances from the official MIPLIB website and the incumbent values of RL-SPH(IS+MVC) and RHF. To the best of our knowledge, the website does not provide information on how the $BKS$ was obtained and how much computation time was used. The time limit for RHF and RL-SPH is set to 100 seconds.}
\vspace{-0.1cm}
\label{tab:joint-exp}
\resizebox{1\columnwidth}{!}{%
\begin{tabular}{cccccccccc}
\specialrule{1pt}{0pt}{2pt}
\multirow{2}{*}{Test instance} &
  \multicolumn{4}{c}{Variable} &
  \multirow{2}{*}{\#constraints} &
  \multirow{2}{*}{Density} &
  \multirow{2}{*}{$BKS \downarrow$} &
\multicolumn{2}{c}{Incumbent value}  \\
\cmidrule[0.5pt](l{0.3em}r{0.3em}){2-5}
\cmidrule[0.5pt](l{0.3em}r{0.3em}){9-10}
                  & Total & \#binaries & \#integers & \#continuous &      & &       &   RHF    & RL-SPH  \\
\specialrule{1pt}{2pt}{2pt}
scpk4 & 100,000 & 100,000 & 0 & 0 & 2,000 & 0.5\% & 318 & 765 & 13,672 \\
ex1010-pi         & 25,200 & 25,200      & 0          & 0            & 1,468 & 0.3\%        & 233.0   & 706.0      & 660.0    \\
toll-like         & 2,883  & 2,883       & 0          & 0            & 4,408 & 0.1\%        & 610.0   & 1,314.0     & 1,774.0   \\
f2gap401600       & 1,600  & 1,600       & 0          & 0            & 40   & 5.0\%        & 82,307.0   & 82,307.0    & 82,307.0  \\
seymour           & 1,372  & 1,372       & 0          & 0            & 4,944 & 0.5\%        & 423.0   & 622.0      & 597.0    \\
cod105            & 1,024  & 1,024       & 0          & 0            & 1,024 & 5.5\%        & -12.0   & -2.0       & -6.0     \\
queens-30         & 900   & 900        & 0          & 0            & 960  & 10.8\%       & -40.0   & -27.0      & -24.0    \\
rail507 & 63,019 & 63,009 & 0 & 10 & 509 & 1.5\% & 174 & 310 & 469 \\
core4872-1529     & 24,656 & 24,645      & 0          & 11           & 4,875 & 0.2\%        & 1,530.0  & 4,121.0     & 2,973.0   \\
neos-1171737      & 2,340  & 1,170       & 0          & 1,170         & 4,179 & 0.6\%        & -195.0        & -82.5      & -101.0   \\
neos-1442119      & 728   & 364        & 0          & 364          & 1,524 & 0.6\%        & -181.0        & -122.4     & -156.0   \\
bppc8-02 & 232 & 229 & 1 & 2 & 59 & 32.0\% & 507 & 530 & 1,824 \\
gen-ip002         & 41    & 0          & 41         & 0            & 24   & 93.7\%       & -4,783.7  & -4,624.1    & -4,678.2  \\
gen-ip021         & 35    & 0          & 35         & 0            & 28   & 96.4\%       & 2,361.5  & 2,486.7     & 2,507.7   \\
gen-ip054         & 30    & 0          & 30         & 0            & 27   & 65.7\%       & 6,841.0   & 7,148.7     & 7,061.0   \\
gen-ip036         & 29    & 0          & 29         & 0            & 46   & 97.7\%       & -4,606.7   & -4,504.2    & -4,549.0  \\
gen-ip016         & 28    & 0          & 28         & 0            & 24   & 100.0\%      & -9,476.2 & -9,229.3    & -9,352.3  \\
\specialrule{1pt}{2pt}{0pt}
\end{tabular}
\vspace{-0.4cm}
}
\end{table}

In general, the joint model yielded incumbent values that are competitive with or superior to the strongest baseline, RHF. 
However, it struggled to sustain this performance on \texttt{scpk4}, where the number of variables differs significantly from the training set.
Maintaining high performance on exceptionally large instances may require scaling up the training instances.
Furthermore, the joint model lacks proficiency in handling specific constraints, such as Set Partitioning found in \texttt{bppc8-02}, which were absent during training. 
Future work should focus on specific constraint classes, extending RL-SPH's capabilities beyond general constraints addressed in this work.
Nevertheless, we emphasize that RL-SPH finds feasible solutions within an average of 12 seconds for these diverse MIPLIB instances, despite being jointly trained on IS and MVC which possess limited characteristics\,(e.g., binary-only variables and sparse matrices).
This demonstrates RL-SPH’s potential to function as a unified solver across diverse problem settings, which is highly desirable in real-world CO\,\cite{liu2024multi}.

\vspace{0.3cm}
\subsubsection{Distribution Changes} \label{appendix:distribution-exp}
% \vspace{-0.1cm}
We conducted an experiment to evaluate the robustness of RL-SPH to changes in distribution. 
As described in Section\,\ref{Benchmarks}, the IS dataset was generated using the Barabási–Albert random graph model\,(BA). 
BA generates scale-free networks that follow a power-law degree distribution. 
It captures the preferential attachment mechanism observed in many real-world networks, such as citation networks and social networks. 
In contrast, Erdős–Rényi random graph model\,(ER) constructs a graph by starting with $u$ nodes and connecting each pair of nodes with probability $\mathrm{p}$\,\cite{albert2002statistical}. 
This process yields a graph where the expected number of edges is approximately $\mathrm{p}\cdot u(u-1)/2$, with edges distributed uniformly at random. 
Therefore, graphs generated by ER and BA have different degree distributions.

Table\,\ref{tab:gen-distrubution} shows that RL-SPH, trained on IS instances generated by BA, produced feasible solutions with a 100\% FR on test instances generated by ER.
Furthermore, RL-SPH outperforms RHF in terms of PG and PI, which is the strongest baseline, demonstrating that it robustly finds high-quality feasible solutions under distribution shifts.

\renewcommand{\arraystretch}{0.9}
\setlength{\tabcolsep}{10pt} 
\begin{table}[h]
\centering
\vspace{-0.1cm}
\caption{Generalization performance of RL-SPH under distribution changes.}
\vspace{-0.1cm}
\label{tab:gen-distrubution}
\resizebox{0.56\columnwidth}{!}{%
\begin{tabular}{cccc}
\specialrule{1pt}{0pt}{2pt}
\multirow{2}{*}{Test dataset} & \multirow{2}{*}{Metric} & \multirow{2}{*}{RHF} & \multirow{2}{*}{\begin{tabular}[c]{@{}c@{}}RL-SPH trained on \\ IS\,(BA)\end{tabular}} \\
                              &                         &                           &                                                                                       \\
\specialrule{1pt}{2pt}{2pt}
\multirow{3}{*}{IS\,(ER)}      & FR $(\%) \uparrow$               & 100                       & 100                                                                                   \\
                              & PG $(\%) \downarrow$               & 7.02±2.39                 & \textbf{0.00}±0.00                                                                    \\
                              & PI ↓                    & 2.7±0.6                   & \textbf{2.4}±0.2  \\                                                                  
\specialrule{1pt}{2pt}{0pt}
\end{tabular}
}
% \vspace{-0.2cm}
\end{table}

% \vspace{-0.1cm}
\subsubsection{Size Changes} \label{appendix:size-exp}
% \vspace{-0.1cm}
% CO research on ML for large-scale ILP is crucial. 
The application of ML to large-scale ILP has emerged as an important issue.
Recently, there have been studies on ML-based heuristics for solving large-scale ILP\,\cite{ye2023gnn-gbdt, ye2024light-milp}. 
These works have proposed techniques that utilize a graph partitioning algorithm\,\cite{tsourakakis2014fennel} to divide large-scale ILP instances into smaller subproblems for optimization. 
Their experimental results showed that breaking a large problem into smaller ones can effectively address the challenges of large-scale ILP.
This finding suggests that, even if RL-SPH may not directly handle large-scale ILP, it has the potential to solve them by decomposing the original large-scale problem into smaller subproblems. 
Therefore, we believe the datasets used in our experiments are sufficient for evaluating our method. 
Nevertheless, to more thoroughly assess its robustness, we experimented on scalability, as follows.

Table\,\ref{tab:gen-size} shows that RL-SPH trained on SC\,($n = 3,000$) successfully generated feasible solutions for larger test instances.
Specifically, RL-SPH achieved a 100\% FR even on test instances with more than three times the number of variables as the training instances.
RL-SPH consistently found higher-quality solutions than RHF, indicating that a model trained on smaller instances can generalize effectively to larger ones.

\begin{table}[h]
\centering
% \vspace{-0.1cm}
\caption{Generalization performance of RL-SPH trained on SC\,($n = 3,000$) under size changes. }
\vspace{-0.1cm}
\label{tab:gen-size}
\resizebox{0.62\columnwidth}{!}{%
\begin{tabular}{ccccc}
\specialrule{1pt}{0pt}{2pt}
\multirow{2}{*}{Metric} & \multicolumn{2}{c}{SC\,($n=6,000$)} & \multicolumn{2}{c}{SC\,($n=10,000$)} \\ 
\cmidrule[0.5pt](l{0.4em}r{0.5em}){2-3} 
\cmidrule[0.5pt](l{0.4em}r{0.5em}){4-5} 
                        & RHF   & \textbf{RL-SPH}     & RHF    & \textbf{RL-SPH}     \\ 
\specialrule{1pt}{2pt}{2pt}
FR $(\%) \uparrow$               & 100        & 100                 & 100         & 100                 \\
PG $(\%) \downarrow$               & 21.47±3.04 & \textbf{0.00}±0.00  & 20.74±3.57  & \textbf{0.00}±0.00  \\
PI ↓                    & 415.3±26.4 & \textbf{237.1}±14.8 & 548.2±31.5  & \textbf{372.8}±20.8 \\ 
\specialrule{1pt}{2pt}{0pt}
\end{tabular}
}
% \vspace{-0.2cm}
\end{table}

% \vspace{-0.1cm}
\subsubsection{Density Changes} \label{appendix:density-exp}
% \vspace{-0.1cm}
Table\,\ref{tab:gen-density} shows that RL-SPH, trained on SC with the density $\rho = 5\%$, consistently generated feasible solutions across test instances with varying densities. 
Specifically, RL-SPH achieved a 100\% FR on both the sparser dataset\,($\rho = 1\%$) and the denser one\,($\rho = 10\%$). 
The results demonstrate that RL-SPH quickly obtained higher-quality solutions than RHF. 
This indicates that RL-SPH is robust to variations in instance density.

\begin{table}[h]
\centering
% \vspace{-0.1cm}
\caption{Generalization performance of RL-SPH trained on SC\,($\rho = 5\%$) under density changes.}
\vspace{-0.1cm}
\label{tab:gen-density}
\resizebox{0.6\columnwidth}{!}{%
\begin{tabular}{ccccc}
\specialrule{1pt}{0pt}{2pt}
\multirow{2}{*}{Metric} & \multicolumn{2}{c}{SC\,($\rho = 1\%$)} & \multicolumn{2}{c}{SC\,($\rho = 10\%$)} \\ 
\cmidrule[0.5pt](l{0.4em}r{0.5em}){2-3} 
\cmidrule[0.5pt](l{0.4em}r{0.5em}){4-5} 
                        & RHF   & \textbf{RL-SPH}    & RHF    & \textbf{RL-SPH}    \\
\specialrule{1pt}{2pt}{2pt}
FR $(\%) \uparrow$               & 100        & 100                & 100         & 100                \\
PG $(\%) \downarrow$               & 23.77±2.33 & \textbf{0.00}±0.00 & 20.73±4.47  & \textbf{0.00}±0.00 \\
PI ↓                    & 80.8±11.2  & \textbf{43.6}±7.4  & 233.4±37.7  & \textbf{113.2}±8.7 \\
\specialrule{1pt}{2pt}{0pt}
\end{tabular}
}
% \vspace{-0.2cm}
\end{table}

% \vspace{-0.1cm}
\subsection{Synergies with Solver-Internal Mechanisms}

We examine two ways in which RL-SPH synergizes with standard
optimization routines: integration with ILP solvers for accelerated solution discovery\,(Appendix\,\ref{appendix:integration}), and interaction with the presolving stage\,(Appendix\,\ref{appendix:presolving}).

\subsubsection{Integration with ILP Solvers} \label{appendix:integration}

We evaluate RL-SPH integrated with SCIP\,(RL-SPH+SCIP) against two baselines: PAS\,\cite{PAS} integrated with SCIP\,(PAS+SCIP) and SCIP.
Both SCIP and PAS+SCIP are run with their default settings.
PAS is trained for the same duration as RL-SPH, taking around 31 minutes on IS, as shown in Appendix~\ref{appendix:Method_cofig}.
Table\,\ref{tab:second_exp} presents the optimization performance of the compared methods under a 50-second time limit, where $BKS$ was obtained by SCIP within 1,000 seconds.
RL-SPH+SCIP fixes each variable to a unanimous value based on the feasible solutions generated by RL-SPH(LP) during the first five seconds.
RL-SPH+SCIP outperformed both SCIP and PAS+SCIP in terms of PG and PI on both benchmarks, achieving high-quality solutions with a PG below 1\% relative to the $BKS$.
Unlike PAS, RL-SPH is capable of generating feasible solutions independently, allowing it to safely fix more variables using an ensemble of feasible solutions, which results in a smaller problem size.
Moreover, RL-SPH supports non-binary integer variables with a 2.3$\times$ lower PI than SCIP, whereas PAS does not support such variables\,(NBI in Table\,\ref{tab:second_exp}).
Although reducing the problem size can significantly accelerate the discovery of high-quality solutions\,\cite{nair2020solving}, it may also introduce sub-optimality.
Given that the primary goal of primal heuristics is to quickly find high-quality feasible solutions\,\cite{berthold2006primal, canturk2024scalable}, exact optimality is often considered a secondary concern.
Nevertheless, improving the optimality within RL-SPH could be a promising direction for future research\,(see Appendix\,\ref{appendix:future_research}), as it remains an open challenge of ML for CO\,\cite{meta-sage}.

\begin{table}[h]
  \centering
  \caption{Solving performance of RL-SPH integrated with SCIP. }
  \vspace{-0.1cm}
  \label{tab:second_exp}
  \scalebox{0.9}{
  \begin{tabular}{ccccc}
    \specialrule{1pt}{0pt}{2pt}
    Dataset              & Metric & SCIP      & PAS+SCIP             & \textbf{RL-SPH+SCIP} \\ 
    \specialrule{1pt}{2pt}{2pt}
    \multirow{3}{*}{IS}  & FR  $(\%) \uparrow$      & 100       & 100                  & \textbf{100}         \\
                         & PG  $(\%) \downarrow$    & 1.24±1.01 & 0.30±0.61            & \textbf{0.14}±0.13   \\
                         & PI $\downarrow$          & 2.4±0.3   & 1.7±0.4              & \textbf{0.5}±0.5     \\ 
    \cmidrule[0.5pt](l{0.3em}r{0.3em}){1-5} 
    \multirow{3}{*}{NBI} & FR  $(\%) \uparrow$      & 100       & \multirow{3}{*}{\parbox{1.5cm}{\centering \textit{not\\applicable}}} & \textbf{100}         \\
                         & PG  $(\%) \downarrow$    & 0.24±0.03 &                      &\textbf{0.23}±0.04   \\
                         & PI $\downarrow$          & 11.5±2.3  &                      & \textbf{4.9}±0.9     \\ 
    \specialrule{1pt}{2pt}{0pt}
    \end{tabular}%
    }
    % \vspace{-0.1cm}
\end{table}

We conducted an additional experiment by combining RL-SPH with the commercial solver Gurobi. 
This experiment followed the same protocol as in Table\,\ref{tab:second_exp}. 
To ensure a fair comparison with SCIP, Gurobi was configured to use a single thread as done in the previous work\,\cite{PAS}, with all other parameters kept at their default settings.
Table\,\ref{tab:gurobi-exp} presents the results on NBI, where RL-SPH+Gurobi denotes RL-SPH integrated with Gurobi. 
The experimental results show that RL-SPH converged to $BKS$ faster when integrated with Gurobi, aligning with the findings presented in Table\,\ref{tab:second_exp}.

\begin{table}[h]
\centering
% \vspace{-0.1cm}
\caption{Solving performance of RL-SPH integrated with ILP solvers.}
\vspace{-0.1cm}
\label{tab:gurobi-exp}
\resizebox{0.75\columnwidth}{!}{%
\begin{tabular}{cccccc}
\specialrule{1pt}{0pt}{2pt}
Dataset             &   Metric    & SCIP      & Gurobi    & RL-SPH+SCIP  & \textbf{RL-SPH+Gurobi}         \\
\specialrule{1pt}{2pt}{2pt}
\multirow{3}{*}{NBI}  & FR $(\%) \uparrow$ & 100       & 100       & 100       & 100              \\
& PG $(\%) \downarrow$ & 0.25±0.03 & 0.03±0.02 & 0.24±0.04 & \textbf{0.03}±0.03        \\
 & PI ↓      & 11.5±2.3  & 5.2±1.0   & 4.9±0.9   & \textbf{2.4}±0.5 \\
\specialrule{1pt}{2pt}{0pt}
\end{tabular}
% \vspace{-0.3cm}
}
\end{table}

\subsubsection{Effect of Presolving on RL-SPH} \label{appendix:presolving}

Enabling presolving in RL-SPH simplifies the given instance by eliminating redundant constraints and variables.
For instance, nearly 600 constraints are removed in the IS instances after presolving. 
Table\,\ref{tab:presolve} shows PG of RL-SPH with and without presolving on IS, using a time limit of 30 seconds. 
The results demonstrate that activating the presolver leads to better solution quality within the same time budget.

\begin{table}[h]
\centering
% \vspace{-0.1cm}
\caption{Effect of presolving on RL-SPH performance on IS. The time limit is set to 30 seconds.}
\vspace{-0.1cm}
\label{tab:presolve}
\resizebox{0.45\columnwidth}{!}{%
\begin{tabular}{ccc}
\specialrule{1pt}{0pt}{2pt}
Presolve & RL-SPH(Random) & RL-SPH(LP) \\
\specialrule{1pt}{2pt}{2pt}
\ding{55}       & 4.00±2.04      & 1.05±1.46  \\
\ding{51}      & \textbf{3.84}±2.30      & \textbf{0.83}±1.13  \\
\specialrule{1pt}{2pt}{0pt}
\end{tabular}
}
\end{table}
% \vspace{-0.1cm}

% \vspace{-0.3cm}
\subsection{Hyperparameter Analysis} \label{appendix:hyperparameter-analysis}

\setlength{\tabcolsep}{9pt} 
\renewcommand{\arraystretch}{0.8}

% \vspace{-0.1cm}
\subsubsection{Effect of the Toward-Optimal Bias} \label{appendix:toward-optimal}
% \vspace{-0.1cm}
Table\,\ref{tab:toward-optimal} shows the performance of RL-SPH under different values of the toward-optimal bias $\alpha$.
With a toward-optimal bias\,(i.e., $\alpha>1$), RL-SPH achieves better PG, PI, and \#win than the value obtained with $\alpha = 1$.
In contrast, RL-SPH without such bias\,(i.e., $\alpha = 1$) consistently attains the lowest \#win across all datasets.
These results demonstrate that injecting a toward-optimal bias helps the agent reach higher-quality solutions, highlighting the importance of guiding the agent's search during training.
Increasing the bias from $\alpha=2$ to $\alpha=5$ does not necessarily yield further improvements.
Notably, in SC, the model with $\alpha=5$ achieves a PG of 2.52\%, which is even higher than the value obtained with $\alpha=1$.
Given that the model also exhibits a noticeably high \#viol\_2 of 11.2 in SC, we conjecture that it becomes overly focused on pursuing a lower objective value (i.e., $obj_{t+1} < obj_b$), which in turn leads to more aggressive attempts that may even violate constraints.
Therefore, it is advisable to choose $\alpha$ to be greater than 1 but not excessively large.

\begin{table}[h]
\centering
% \vspace{-0.1cm}
\caption{Performance of RL-SPH under different toward-optimal bias values. \#viol\_2 is the sum of variable bound violations and linear constraint violations in $phase2$. \#win measures the number of test instances in which a method reaches the best value among all compared methods. The time limit is set to 150 seconds for SC and 50 seconds for all other datasets.}
\vspace{-0.1cm}
\label{tab:toward-optimal}
\resizebox{0.83\columnwidth}{!}{%
\begin{tabular}{ccccccc}
\specialrule{1pt}{0pt}{2pt}
Dataset              & Toward-optimal bias & FR  $(\%) \uparrow$ & PG  $(\%) \downarrow$    & PI $\downarrow$     & \#win $\uparrow$    & \#viol\_2  \\ 
\specialrule{1pt}{2pt}{2pt}
\multirow{3}{*}{IS}  & $\alpha=1$      & 100     & 3.41±2.10  & 7.2±1.0   & 9   & 1.0±0.1  \\
                     & $\alpha=2$     & 100     & 0.94±1.29   & 3.4±0.6	  &45   & 0.3±0.1 \\ 
                     & $\alpha=5$     & 100     & \textbf{0.85}±1.20   &	\textbf{3.3}±0.6	  & \textbf{53}  & 0.4±0.1  \\ 
\cmidrule[0.5pt](l{0.3em}r{0.3em}){1-7} 
\multirow{3}{*}{CA}  & $\alpha=1$      & 100     & 3.22±3.00   & 	12.5±1.3  & 	17  & 2.2±0.5  \\
                     & $\alpha=2$     & 100     & \textbf{0.88}±1.64	 & \textbf{11.8}±1.1 & 	\textbf{57}  & 2.1±0.4  \\ 
                     & $\alpha=5$     & 100     & 2.49±2.58 & 	12.4±1.2	 & 26  & 3.1±0.5  \\ 
\cmidrule[0.5pt](l{0.3em}r{0.3em}){1-7} 
\multirow{3}{*}{SC}  & $\alpha=1$      & 100 & 2.39±2.13 &	98.2±6.8  & 	32   & 4.0±4.7 \\
                     & $\alpha=2$     & 100  & \textbf{1.74}±2.22 &	98.4±7.2  & 	\textbf{50} & 4.7±6.3  \\ 
                     & $\alpha=5$     & 100  & 2.52±2.41 &	\textbf{97.8}±7.2   &	33  & 11.2±17.9 \\ 
\cmidrule[0.5pt](l{0.3em}r{0.3em}){1-7} 
\multirow{3}{*}{MVC} & $\alpha=1$      & 100     & 14.18±0.90  & 	8.8±0.4    &	0  &1.7±0.2 \\
                     & $\alpha=2$     & 100     &  \textbf{0.43}±0.69   & 	\textbf{4.6}±0.3    &	\textbf{60}  &0.8±0.1 \\ 
                     & $\alpha=5$     & 100     &  0.67±0.83   & 	4.7±0.4    &	41 &1.1±0.1  \\ 
\cmidrule[0.5pt](l{0.3em}r{0.3em}){1-7} 
\multirow{3}{*}{NBI} & $\alpha=1$      & 100     & 0.24±0.11    &	10.9±0.6  & 	1  & 1.4±0.3\\
                     & $\alpha=2$     & 100     & 0.06±0.07 &	\textbf{10.7}±0.6  & 	29  & 2.7±0.6\\ 
                     & $\alpha=5$     & 100     & \textbf{0.01}±0.03 &	10.8±0.7  & 	\textbf{71} & 4.6±0.7 \\ 
\specialrule{1pt}{2pt}{0pt}
\end{tabular}%
}
% \vspace{-0.2cm}
\end{table}

% \vspace{-0.1cm}
\subsubsection{Effect of the Neighbor Variable Ratio} \label{appendix:seed-neighbor}
% \vspace{-0.1cm}

Table\,\ref{tab:seed-ratio} presents the performance of RL-SPH under different ratios of neighbor variables among the changeable variables\,(i.e., $q/\tilde{n}$).
Regardless of the ratio values, RL-SPH finds a feasible solution in all experimental settings. 
In IS and MVC, higher ratios enable RL-SPH to reach higher-quality feasible solutions more quickly.
These results suggest that IS and MVC exhibit stronger interactions among neighbor variables compared with the other datasets. 
As shown in Table\,\ref{tab:degree}, IS and MVC have relatively high variable degrees compared to their constraint degrees\,(i.e., a higher degree ratio). 
Consequently, when the agent modifies the value of a seed variable to improve feasibility, it must also account for the many constraints involving that variable, along with the other variables that appear in those constraints.
Thus, for datasets with a large degree ratio, a larger neighbor variable ratio is preferable, as it allows the agent to capture interactions among more neighbor variables.

\begin{table}[h!]
\centering
% \vspace{-0.1cm}
\caption{Performance of RL-SPH under varying neighbor variable ratios. The time limits are consistent with those in Table~\ref{tab:toward-optimal}.}
\vspace{-0.1cm}
\label{tab:seed-ratio}
\resizebox{0.71\columnwidth}{!}{%
\begin{tabular}{cccccc}
\specialrule{1pt}{0pt}{2pt}
Dataset              & Neighbor variable ratio & FR  $(\%) \uparrow$ & PG  $(\%) \downarrow$    & PI $\downarrow$     & \#win $\uparrow$ \\ 
\specialrule{1pt}{2pt}{2pt}
\multirow{3}{*}{IS}  & 0.3     & 100    & 8.22±1.70 & 6.9±0.8 & 0 \\ 
                     & 0.5     & 100    & \underline{5.86}±1.94 & \underline{5.7}±0.9 & 0   \\ 
                     & 0.7      & 100   &  \textbf{0.00}±0.00 & \textbf{2.8}±0.2 & \textbf{100} \\ 
\cmidrule[0.5pt](l{0.3em}r{0.3em}){1-6} 
\multirow{3}{*}{CA}  & 0.3     & 100     & \textbf{0.99}±2.08            & \textbf{11.9}±1.2             & \textbf{60}                   \\
                     & 0.5     & 100     & \underline{1.57}±2.11            & \underline{12.1}±1.2             & \underline{40}                   \\
                     & 0.7      & 100     & 8.84±3.38            & 15.0±1.6             & 0                    \\
\cmidrule[0.5pt](l{0.3em}r{0.3em}){1-6} 
\multirow{3}{*}{SC}  & 0.3     & 100  & \underline{2.15}±2.24            & \underline{97.6}±6.7             & \underline{39}                   \\
                     & 0.5     & 100  & \textbf{1.59}±2.10            & 98.2±7.4             & \textbf{52}                   \\
                     & 0.7      & 100 & 2.55±2.37            & \textbf{97.2}±7.4             & 31                   \\
\cmidrule[0.5pt](l{0.3em}r{0.3em}){1-6} 
\multirow{3}{*}{MVC} & 0.3     & 100    & 7.49±0.94            & 7.8±0.4              & 0                    \\
                     & 0.5     & 100    & \underline{6.90}±0.97            & \underline{7.6}±0.4              & 0                    \\
                     & 0.7      & 100   & \textbf{0.00}±0.00            & \textbf{4.8}±0.2              & \textbf{100}                  \\
\cmidrule[0.5pt](l{0.3em}r{0.3em}){1-6} 
\multirow{3}{*}{NBI} & 0.3     & 100     & \textbf{0.03}±0.04            & 10.9±0.7             & \textbf{59}                   \\
                     & 0.5     & 100     & \underline{0.05}±0.06            & \textbf{10.7}±0.6             & \underline{32}                   \\
                     & 0.7      & 100     & 0.09±0.06            & 10.9±0.8             & 10                   \\
\specialrule{1pt}{2pt}{0pt}
\end{tabular}%
}
\end{table}

\begin{table}[h]
\centering
\vspace{-0.1cm}
\caption{Average degree of constraints and variables.}
\vspace{-0.1cm}
\label{tab:degree}
\resizebox{0.62\columnwidth}{!}{%
\begin{tabular}{cccccc}
\specialrule{1pt}{0pt}{2pt}
    Dataset               & IS   & CA   & SC     & MVC  & NBI    \\ 
\specialrule{1pt}{2pt}{2pt}
    Degree of constraints $d_c$ & 2.00 & 8.31 & 150.00 & 2.00 & 198.98 \\
    Degree of variables $d_v$   & 7.94 & 5.56 & 100.00 & 7.96 & 198.98 \\
    Degree ratio $d_v/d_c$         & 3.96  & 0.67  & 0.67    & 3.98  & 1.00    \\ 
\specialrule{1pt}{2pt}{0pt}
% \vspace{-0.6cm}
\end{tabular}
}
\end{table}

% \vspace{-0.3cm}
\subsection{Qualitative Analysis} \label{appendix:qualitative}
% \vspace{-0.1cm}

% \vspace{-0.1cm}
\subsubsection{Behavior of the Two-Phase Reward System} \label{appendix:phase-analysis}
\begin{figure}[h]
    % \vspace{-0.1cm}
  \centering
  \includegraphics[width=1\columnwidth]{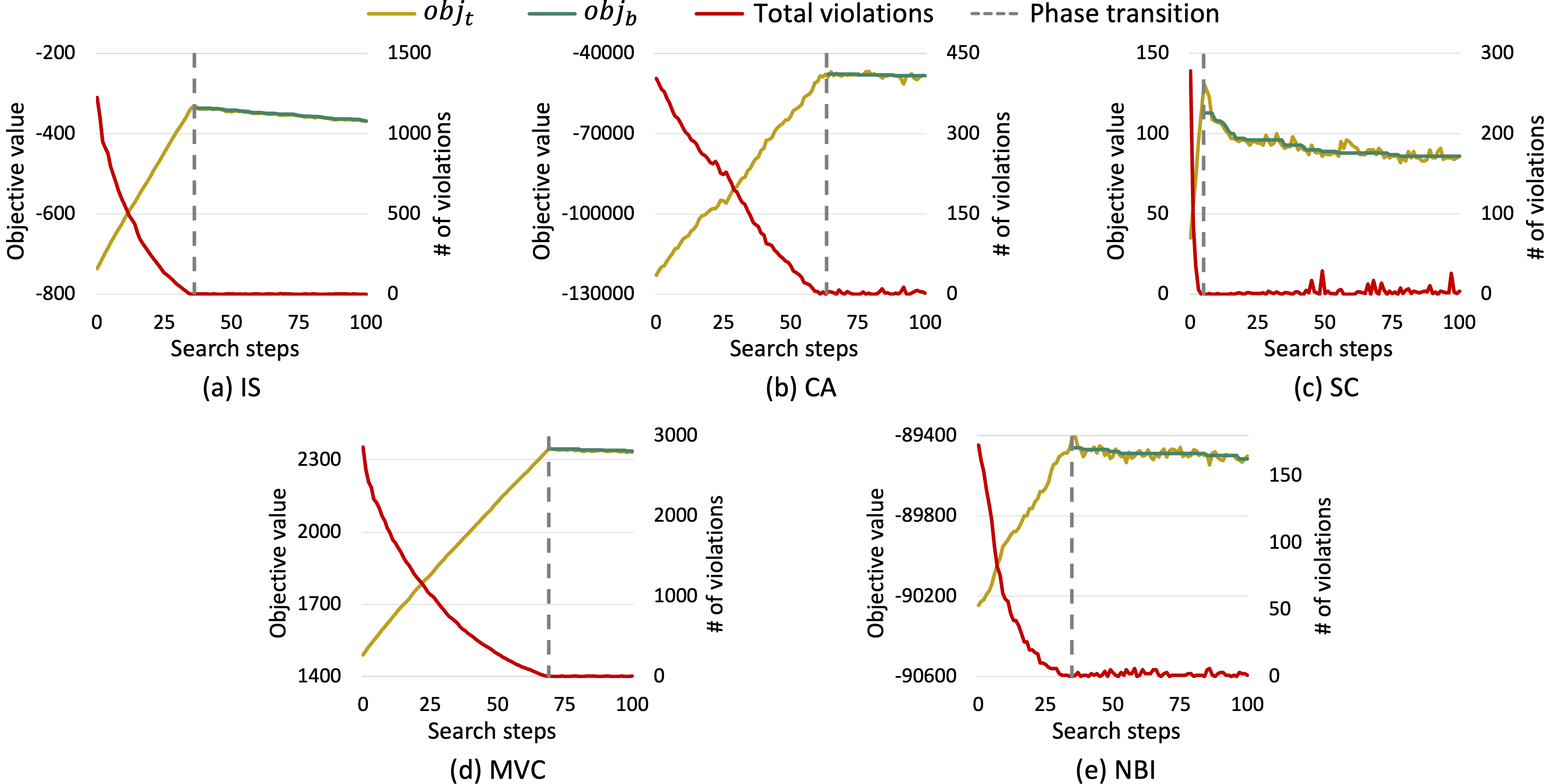}
  \vspace{-0.1cm}
  \caption{Illustration of RL-SPH’s optimization trajectory over search steps, using one randomly sampled instance from each dataset. The total number of violations is computed as the sum of variable bound violations and linear constraint violations. The period before the phase transition corresponds to $phase1$, and the period after the transition corresponds to $phase2$.}
  % \vspace{-0.1cm}
  \label{fig:curve}
\end{figure}

% \vspace{-0.1cm}
To verify that the proposed two-phase reward system operates as intended, we perform an additional analysis.
Figure\,\ref{fig:curve} illustrates the optimization trajectory of RL-SPH over 100 search steps on instances randomly sampled from each dataset. 
As described in Section\,\ref{Reward}, the primary goal of $phase1$ is to repair infeasible solutions and obtain the first feasible one. 
Across all datasets, the initial solutions violate a large number of constraints.
For example, the initial solution for MVC violates approximately 3,000 constraints. 
RL-SPH gradually reduces the number of violated constraints and eventually satisfies all of them, thereby discovering the first feasible solution.

According to the policy defined in Section \ref{Reward}, $phase2$ begins once the first feasible solution is found.
Since the goal of $phase 2$ is to improve the incumbent objective value $obj_b$, the agent searches for higher-quality feasible solutions within its neighborhood\,(see Algorithm\,\ref{alg:solution_search}). 
As shown in Figure\,\ref{fig:curve}, after transitioning into $phase2$, the agent takes moves that preserve feasibility while improving $obj_b$.
Across all datasets, the RL-SPH agent exhibits a consistent behavioral pattern, providing evidence that it operates in accordance with the intended design of our reward system.

% \vspace{-0.1cm}
\subsubsection{Capturing Long-Range Variable Dependencies} \label{appendix:Long-range}
% \vspace{-0.1cm}

\begin{figure}[h]
% \vspace{-0.1cm}
     \centering
     \hfill
     \begin{subfigure}[b]{0.47\columnwidth}
         \centering
         \includegraphics[width=\columnwidth]{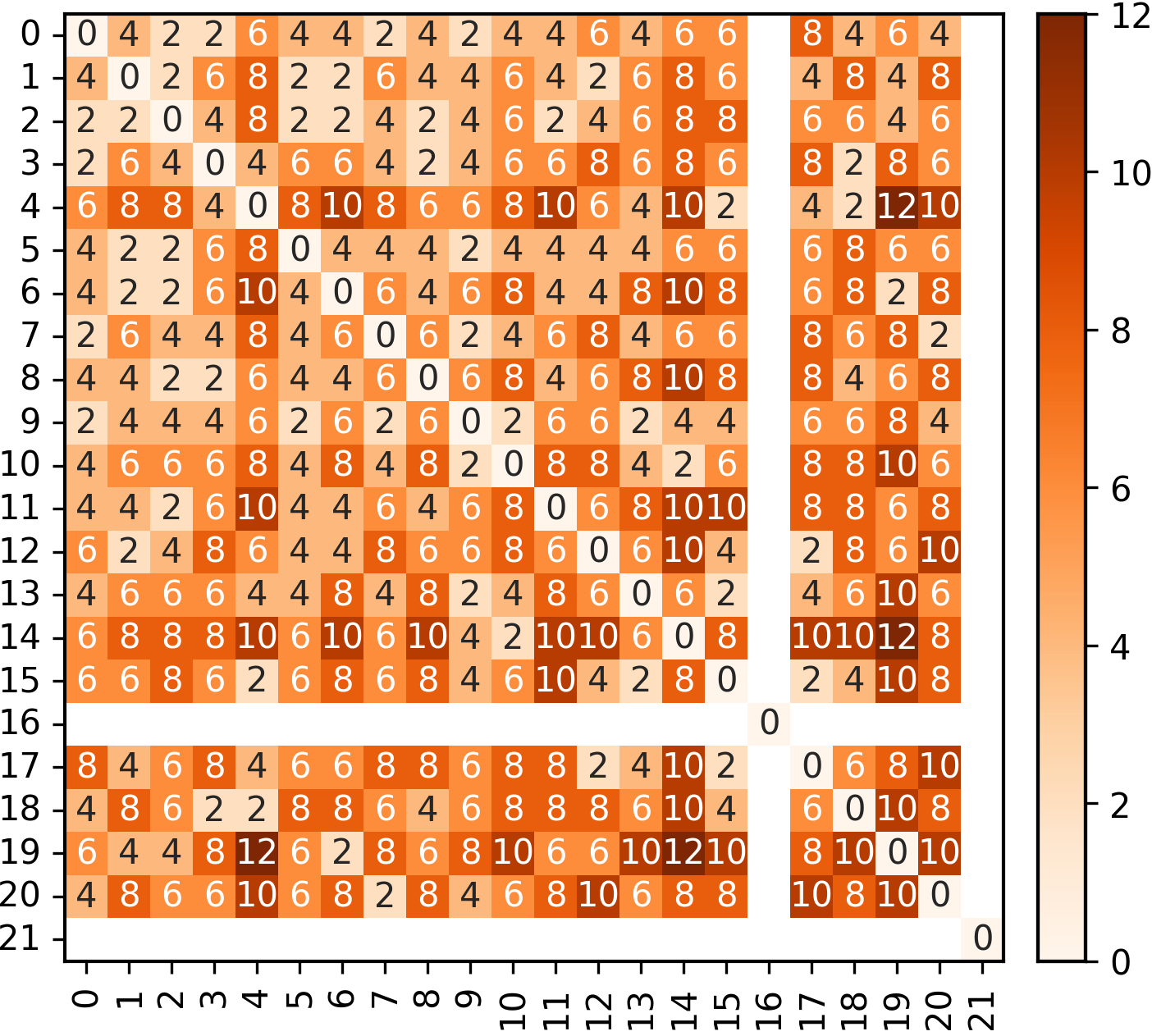}
         \caption{Heatmap of shortest-path hop counts.}
         \label{fig:hop-map}
     \end{subfigure}
     \hfill
     \begin{subfigure}[b]{0.488\columnwidth}
         \centering
         \includegraphics[width=\columnwidth]{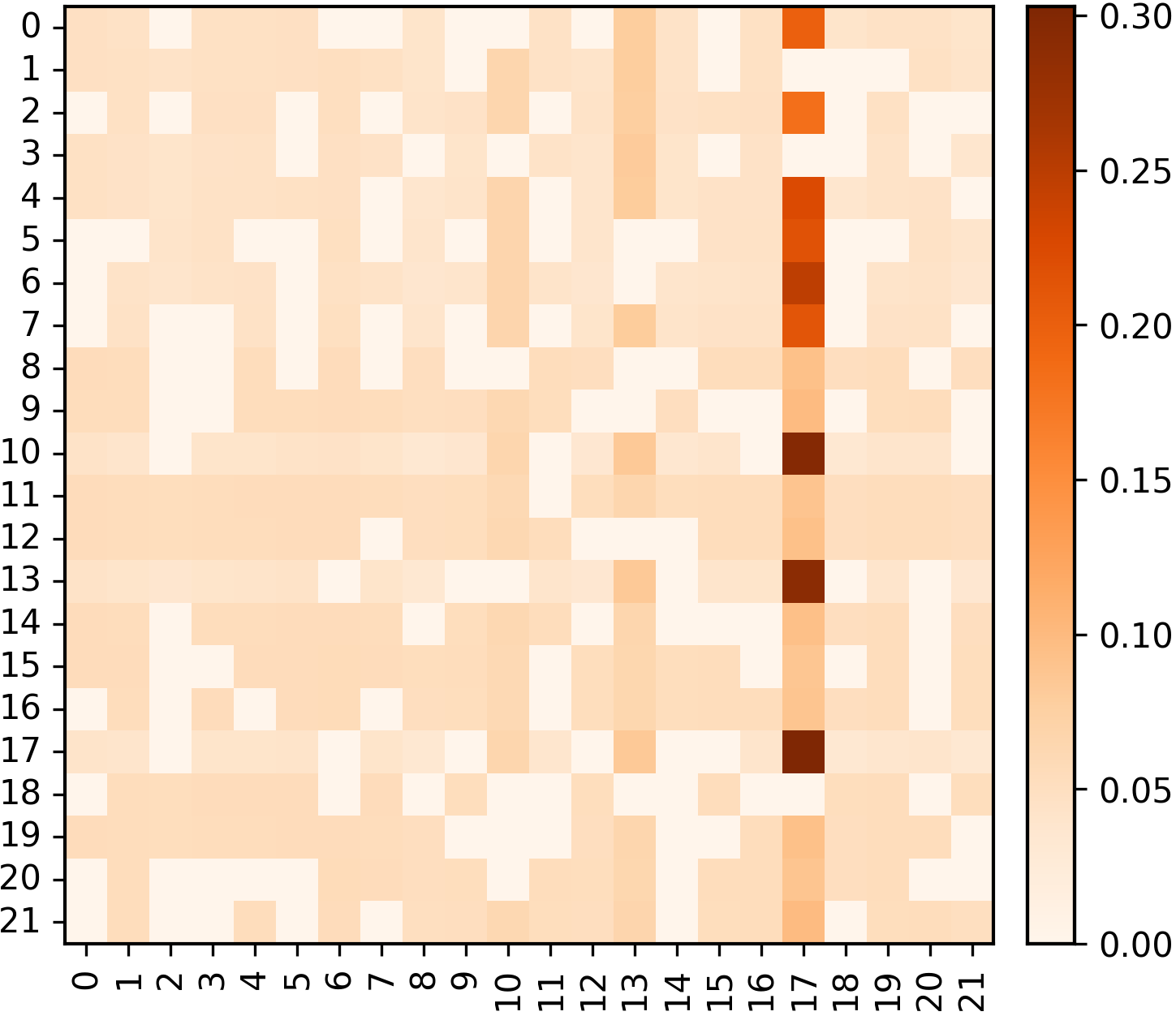}
         \caption{Attention map from our ILP-GT.}
         \label{fig:attn-map}
     \end{subfigure}
     \hfill
     % \vspace{-0.1cm}
     \caption{Heatmap of shortest-path hop counts and attention map from our ILP-GT. This example is randomly sampled from the IS test set. The attention map is taken from the final Transformer layer. The x-axis and y-axis correspond to variable nodes. In (a), blank cells in the heatmap indicate that there is no path between the two nodes.}
     % \vspace{-0.3cm}
    \label{fig:heap-map}
\end{figure}

To examine whether our ILP-GT learns long-range dependencies among variables in ILP problems represented as bipartite graphs, we conduct a case study showing attention weights between distant variables.
Figure \ref{fig:heap-map} shows a snapshot captured during the RL-SPH's optimization of a randomly sampled instance from the IS dataset, where interactions among neighbor variables are important as discussed in Appendix\,\ref{appendix:seed-neighbor}.
Figure \ref{fig:hop-map} reports the shortest-path hop distances between variable nodes that are fed into ILP-GT, while Figure \ref{fig:attn-map} depicts the attention weights computed among these variable nodes.
Although the reward context is also included as input nodes in practice, it is omitted here for consistency with Figure \ref{fig:hop-map}.
As shown in Figure \ref{fig:hop-map}, variable-node 10 is eight hops away from variable-node 17, yet it receives a larger attention weight than several nearer nodes.
This behavior aligns with observations reported in Graph Transformer literature\,\cite{wu2021representing} and suggests that our ILP-GT can effectively capture long-range dependencies among decision variables in ILP.

% \vspace{-0.1cm}
\subsubsection{Training Stability and Convergence} \label{appendix:train-curves}
% \vspace{-0.1cm}

\begin{figure}[h!]
    % \vspace{-0.1cm}
  \centering
  \includegraphics[width=0.97\columnwidth]{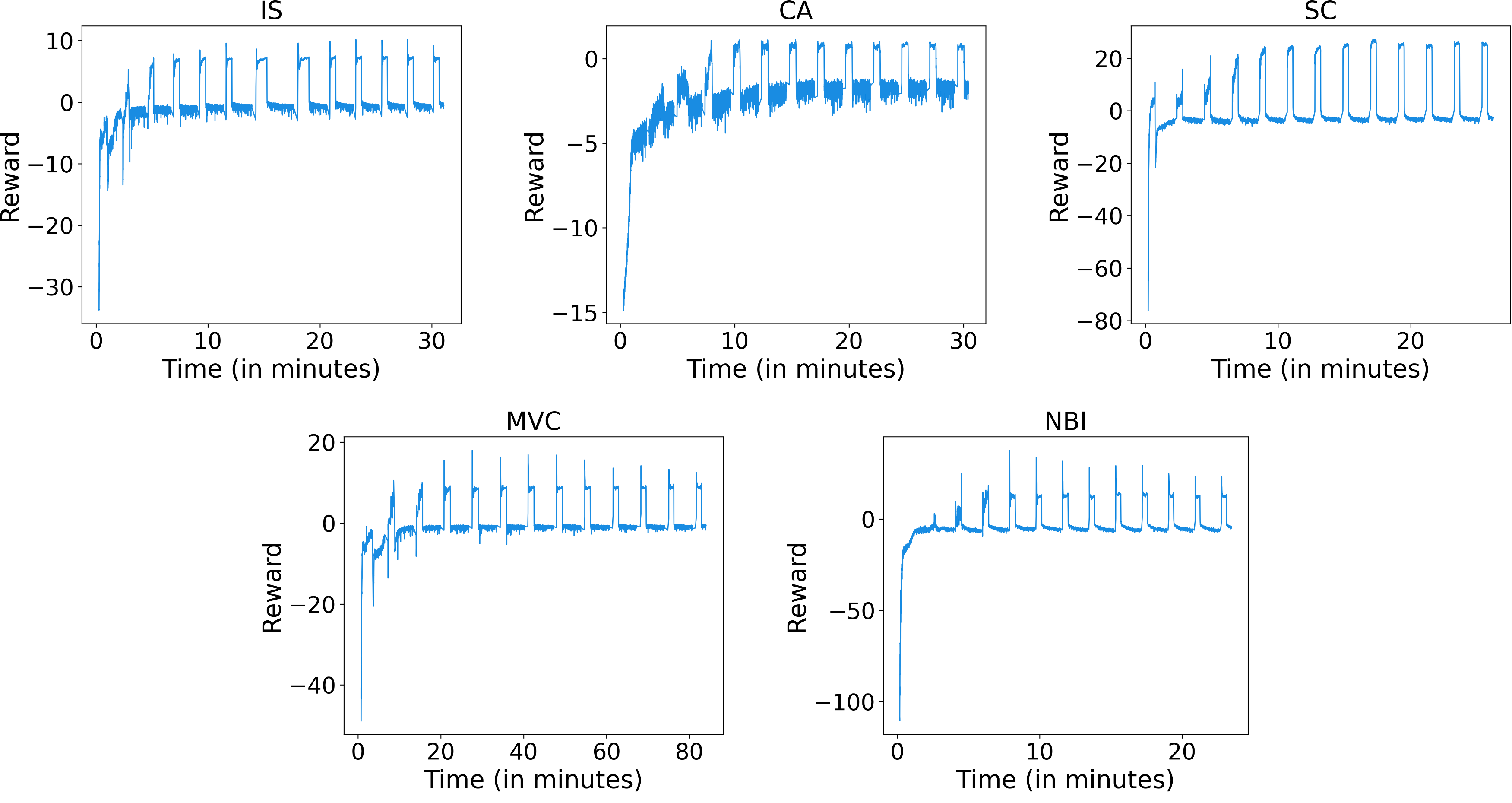}
  \vspace{-0.1cm}
  \caption{The training curves of RL-SPH on each dataset.}
  % \vspace{-0.2cm}
  \label{fig:train-curve}
\end{figure}

Figure\,\ref{fig:train-curve} plots the rewards that RL-SPH receives over time during training on each dataset. 
Since the reward mechanism of RL-SPH consists of $phase1$ and $phase2$, the reward magnitude periodically fluctuates whenever the agent transitions to $phase2$ or begins $phase1$ of the next instances.
As described in Section\,\ref{Learning Algorithm} and Appendix\,\ref{appendix:Method_cofig}, the agent stays in $phase1$ for one quarter of the total step limit allocated to each instance, which indicates that the period with relatively higher rewards in Figure\,\ref{fig:train-curve} corresponds to $phase1$.
Except for MVC, the training process converges within about 10 minutes and consistently exhibits a repeating and stable pattern.

\vspace{-0.2cm}
\section{Additional Discussion} \label{appendix:discussion}
\vspace{-0.1cm}
\subsection{Feasibility as a Core Focus of This Work} \label{appendix:core_focus}
\vspace{-0.1cm}

In this work, our primary focus is on finding feasible solutions from an infeasible start point, rather than optimizing objective values.
By definition in Section\,\ref{Mixed Integer Linear Programming}, an optimal solution is \textit{feasible} and achieves the lowest objective value.
Even if a solution yields the lowest objective value, it is considered invalid if it violates even a single constraint.
Therefore, achieving feasibility is a crucial prerequisite for optimality.
Furthermore, reaching optimality is recognized as a major challenge in the ML for CO community\,\cite{meta-sage}, which underscores the importance of feasibility in ILP.

Despite this importance, feasibility is not typically guaranteed by primal heuristics\,\cite{shoja2023exact}, including E2EPH methods\,\cite{nair2020solving, shen2021learning, yoon2022confidence, PAS, canturk2024scalable, huang2024contrastive, liu2025apollo, kdd_complete, iclr_complete}.
As discussed in Appendix\,\ref{Related Work}, to the best of our knowledge, no existing E2EPH has achieved a 100\% feasibility rate.
These observations suggest that the \textit{feasibility rate}\,(FR) should be regarded as a fundamental evaluation metric in this research area.
Therefore, we argue that feasibility should be prioritized and that methods achieving higher FR should be regarded as stronger baselines.
From this perspective, we consider RHF to be a stronger baseline than other primal heuristics, including the E2EPH methods.

\vspace{-0.1cm}
\subsection{Training Costs} \label{appendix:training_times}
\vspace{-0.1cm}
Most studies on E2EPH are based on supervised learning models\,\cite{nair2020solving, yoon2022confidence, PAS, huang2024contrastive, canturk2024scalable, kdd_complete}. 
Obtaining training labels for ILP\,(e.g., near-optimal solutions) requires a considerable amount of time. 
For example, such labels are obtained by running Gurobi with a time limit of 3,600 seconds for a single instance\,\cite{PAS,huang2024contrastive}.
Considering that training requires on the order of hundreds of instances, this constitutes a substantial computational cost.
This labeling process becomes even more computationally expensive for larger instances, limiting the practicality for NP-hard problems \cite{canturk2024scalable}.
Moreover, training for the supervised learning models usually converges within 5 hours\,\cite{huang2024contrastive}.

In contrast, RL-SPH does not require near-optimal feasible solutions for training. 
In addition, all datasets required less than an hour to train RL-SPH, except MVC with 83 minutes, as shown in Table\,\ref{tab:train_time}.
When compared to a representative supervised learning method\,\cite{PAS} in terms of training time, RL-SPH achieved an average speedup of 8.7$\times$ on the same datasets, as shown in Table\,\ref{tab:comparison-train-time}.
Consequently, RL-SPH incurs lower training costs than supervised learning methods, as it requires no labeling process and less training time.
For an unsupervised method\,\cite{iclr_complete}, training on the datasets took about 16 hours, whereas RL-SPH was nearly 34$\times$ faster, as shown in Table\,\ref{tab:comparison-train-time}.
This substantial reduction in training time underscores the practicality of RL-SPH.

\vspace{-0.1cm}
\subsection{Training Efficiency} \label{appendix:training_efficiency}
\vspace{-0.1cm}
Our feasibility-aware search strategy enables efficient training.
This strategy allows the agent to focus on variables that are more likely to improve feasibility by filtering out uninformative inputs. 
In a real-world analogy, this is akin to preparing for a final exam by focusing on the topics the professor highlighted during lectures, rather than studying an entire 1000-page textbook including bibliography and index. 
The former one enables targeted learning, which could result in improved exam outcomes. 
In this sense, this learning strategy can be viewed as a higher-level attention mechanism—operating at the page-level—compared to one that focuses on important words within a page.
Likewise, our search strategy narrows the broad observation space to more informative regions, enhancing learning efficiency through denser reward signals.
We emphasize this feasibility-aware search strategy as a core contribution, as it provides a systematic way to reduce the search space and enhance the RL agent's learning efficiency.
The insights from this learning strategy may also be useful in other domains, such as RL for routing optimization, which requires several days of training\,\cite{son2024equity}.

\vspace{-0.1cm}
\subsection{Complexity Analysis of Feasibility-Aware Search} \label{appendix:novelty_search}
\vspace{-0.1cm}

The computational complexity of Algorithm\,\ref{alg:solution_search} is determined by three components: variable selection, Transformer processing, and observation computation.
For variable selection, multiplying the feasibility state vector $\tilde{\mathbf{f}}_t \in \mathbb{R}^{m \times 1}$ with the coefficient matrix $\tilde{\mathbf{A}} \in \mathbb{R}^{m \times n}$ requires $O(mn)$ operations.
Processing $\tilde{n}^2 + 3$ input tokens with the Transformer incurs a cost of $O(\tilde{n}^2)$\,\cite{katharopoulos2020transformers}.
For the observation update, only the changeable variables need to be used for computation, which results in $O(m \tilde{n})$.
Therefore, the overall computational complexity of Algorithm\,\ref{alg:solution_search} is $O(mn + \tilde{n}^2)$.
We believe that our strategy represents only one possible way to reduce the search space, and that further optimization is certainly possible.
For instance, variable selection currently takes $O(mn)$, which becomes quadratic when $m \approx n$. 
Addressing this bottleneck remains a promising direction for future research.

% \vspace{-0.1cm}
\subsection{Potential Bottlenecks at Extreme Scales} \label{appendix:bottleneck}
% \vspace{-0.1cm}

While our search strategy alleviates the scalability issue by passing only $\tilde{n}$\,$(=2 \left\lceil\log_2n\right\rceil)$ inputs to our ILP-GT, the quadratic complexity of the Transformer’s attention mechanism could become a bottleneck for extremely large-scale problems. 
For such cases, it would be beneficial to consider attention architectures specifically designed for lower computational complexity, such as RetNet\,\cite{retnet} or other variants.

Even if the quadratic attention in the Transformer is resolved, the variable selection may become a new bottleneck because its complexity becomes quadratic when $m \approx n$\,(see Appendix\,\ref{appendix:novelty_search}).
% Addressing this issue will likewise be an important direction for future work.
Furthermore, if the number of constraints becomes extremely large, projecting the structural information into the hidden size may incur substantial computational cost.
This issue could be addressed by explicitly introducing constraint nodes as part of the ILP-GT input nodes, rather than compressing all structural information into a single projected vector.
We conducted preliminary experiments to explore this direction and observed promising performance, although a more thorough evaluation is certainly required.

% \vspace{-0.1cm}
\subsection{Handling Nonlinear Constraints} \label{appendix:nonlinear_constraints}
% \vspace{-0.1cm}
Since our study is scoped to \textit{Integer Linear Programming}\,(ILP), we did not explicitly consider designing our reward system to handle nonlinear constraints. Nonetheless, we suspect that the reward system might also work in the presence of nonlinearities because the reward is computed based on the change from the previous timestep, regardless of the type of constraints. However, this has not yet been tested and remains to be explored further.

% \vspace{-0.1cm}
\subsection{Changes in Movement Magnitude} \label{appendix:movement_unit}
% \vspace{-0.1cm}
Having a larger neighborhood is effective in enhancing search performance\,\cite{shaw1998using}.
The size of the neighborhood is associated not only with the number of neighbor variables but also with the magnitude of movement. 
We denote $\Delta_x$ as the maximum magnitude of change for any variable.
To allow larger movement, RL-SPH can be easily extended by expanding the action space. 
For example, to set $\Delta_x=2$, the action space can be expanded to \{-2, -1, 0, +1, +2\}. 
Table\,\ref{tab:neighbor} summarizes our preliminary experiments on the effect of $\Delta_x$.
The model with $\Delta_x= 5$ achieved the highest \#win, which leverages the largest action set in this experiment. 
These results suggest that expanding the movement range enhances the agent's capability to reach better objective values.

We set $\Delta_x = 1$ in our main experiments to prioritize feasibility over optimality\,(see Appendix\,\ref{appendix:core_focus}).
Notably, RL-SPH even with $\Delta_x = 1$ outperforms four SPH baselines and the SOTA E2EPH\,\cite{iclr_complete}.
Thus, increasing action magnitudes would not affect our core conclusions.
Nevertheless, this observation points toward a promising avenue for improving solution optimality within RL-SPH, which we discuss further in Appendix\,\ref{appendix:future_research}.

% \vspace{-0.1cm}
\setlength{\tabcolsep}{10pt} 
\renewcommand{\arraystretch}{0.7}
\begin{table}[h]
\centering
\caption{Performance by $\Delta_x$ on the NBI dataset with 100 variables and 50 constraints. A 300-second time limit was applied to all cases.}
\vspace{-0.1cm}
\label{tab:neighbor}
\resizebox{0.5\columnwidth}{!}{%
\begin{tabular}{cccc}
\specialrule{1pt}{0pt}{2pt}
Max Magnitude & FR $(\%) \uparrow$ & PG $(\%) \downarrow$ & \#win $\uparrow$ \\
\specialrule{1pt}{2pt}{2pt}
$\Delta_x=1$          & 100       & 0.36      & 21      \\
$\Delta_x=3$          & 100       & 0.31      & 40      \\
$\Delta_x=5$          & 100       & 0.31      & \textbf{50}     \\
\specialrule{1pt}{4pt}{0pt}
\end{tabular}%
}
\vspace{-0.3cm}
\end{table}

\subsection{Promising Directions to Improve Optimality} \label{appendix:future_research}
% \vspace{-0.1cm}
To further improve the optimality within RL-SPH, several research directions can be explored. 
First, a promising avenue is the dynamic determination of the action space.
As observed in Appendix~\ref{appendix:movement_unit}, a larger movement magnitude could improve the objective value. 
Since the most suitable action space may depend on the agent’s current state, dynamically adapting the action space could be more effective than using a static one.
For instance, when the agent is trapped in a deeper local optimum, a larger movement magnitude may be required to escape.

Second, integrating metaheuristics\,\cite{hussain2019metaheuristic}, high-level strategies that guide the search process, could further enhance RL-SPH. 
For example, Iterated Local Search\,(ILS)\,\cite{lourencco2003iterated} introduces perturbations to escape local optima during the search. 
In this context, RL-SPH could be extended to detect whether it is stuck in a local optimum and learn how to apply perturbations—in both direction and magnitude—to improve search performance.

Third, RL-SPH could be extended to learn how to select promising neighborhoods. 
Prior studies on LNS have demonstrated the effectiveness of learning-based neighborhood selection policies\,\cite{huang2023searching}. 
Building on this line of work, future research could explore how RL-SPH can learn to identify and exploit high-quality neighborhoods.

\end{document}